\documentclass[12pt,oneside, a4paper]{book}
\usepackage{etex}
\usepackage{xcolor}
\usepackage[pdftex]{graphicx}
\usepackage{rotating}
\usepackage{epsfig}
\usepackage{epstopdf}
\usepackage[T1]{fontenc}
\usepackage[utf8]{inputenc}
\usepackage{cmap}
\usepackage{ae}
\usepackage[unicode]{hyperref}
\usepackage{mathptmx}
\usepackage{amscd}
\usepackage{amssymb}
\usepackage{amsmath}
\usepackage{amsfonts}

\usepackage{booktabs}
\usepackage[inline]{enumitem}

\usepackage{placeins}

\usepackage{url}
\usepackage{graphicx}
\usepackage{pifont}
\usepackage{pgfplots}
\usepackage{tikz}
\usepackage{xspace}
\usepackage{comment}
\usepackage{times}
\usepackage{latexsym}
\usepackage[left=2.5cm,right=2.5cm,top=2.5cm,bottom=2.5cm]{geometry}
\usepackage{setspace} 
\usepackage{fancyhdr} % setting up header and position of page numbers
\usepackage{hyperref}
\usepackage{hhline}
\usepackage{delarray}
\usepackage{array}  % package for some table properties
\usepackage{tabularx} % package that allows dynamical changing table cell width
\usepackage{multirow}  % package that enables multiple rows in a table
\usepackage[bf, font=small]{caption}
\usepackage[labelfont=small, font=small]{subcaption}
\usepackage{wasysym}
\usepackage{subeqnarray}
\usepackage{aeguill}
\usepackage{pdflscape} % setting page into landscape view
\usepackage{enumitem} % for itemize lists
\setlist{nolistsep}   % setting for itemize lists

\renewcommand{\arraystretch}{1.5} % stretching row height

\usepackage[square, numbers, sort]{natbib} 
\usepackage{booktabs}

\newcommand{\maybe}[1]{}

\newcommand{\nospacetext}[1]{\makebox[0pt][l]{#1}}

\newcommand{\pandora}{\textsc{Pandora}\xspace}
\let\savenumberline\numberline
\def\numberline#1{\savenumberline{#1.}}

\makeatletter
\renewcommand*\l@chapter[2]{%
  \ifnum \c@tocdepth >\m@ne
  \addpenalty{-\@highpenalty}%
  \vskip 1.0em \@plus\p@
  \setlength\@tempdima{1.5em}%
  \begingroup
  \parindent \z@ \rightskip \@pnumwidth
  \parfillskip -\@pnumwidth
  \leavevmode \bfseries
  \advance\leftskip\@tempdima
  \hskip -\leftskip
  #1\nobreak\normalfont\leaders\hbox{$\m@th
    \mkern \@dotsep mu\hbox{.}\mkern \@dotsep
    mu$}\hfill\nobreak\hb@xt@\@pnumwidth{\hss #2}\par
  \penalty\@highpenalty
  \endgroup
  \fi}
\makeatother

\makeatletter
\renewcommand*\env@matrix[1][\arraystretch]{%
  \edef\arraystretch{#1}%
  \hskip -\arraycolsep
  \let\@ifnextchar\new@ifnextchar
  \array{*\c@MaxMatrixCols c}}
\makeatother

\AtBeginDocument{\addtocontents{toc}{\protect\thispagestyle{empty}}}
\AtBeginDocument{\addtocontents{lof}{\protect\thispagestyle{empty}}}
\AtBeginDocument{\addtocontents{lot}{\protect\thispagestyle{empty}}}

\begin{document}

%%%%%%%%%%%%%%%%%%%%%%%%%%%%%%%%%%%%%%%%%%%%%%%%%%%%%%%%%%%%%%%%%%%%%%%%%%%
\frontmatter

%%%%%%%%%%%%%%%%%%%% NASLOVNICA / FRONT COVER PAGE %%%%%%%%%%%%%%%%%%%%%%%%
\begin{titlepage}
  \fontsize{16pt}{20pt}\selectfont
  \fontfamily{phv}\fontseries{mc}\selectfont
  \newgeometry{left=3cm,right=3cm,top=3cm,bottom=2.5cm}
  \setlength{\intextsep}{0pt plus 0pt minus 0pt}

  \begin{center}
    \begin{figure}[ht!]
      \begin{center}
        \includegraphics[height=4.1184cm, width=5.94cm]{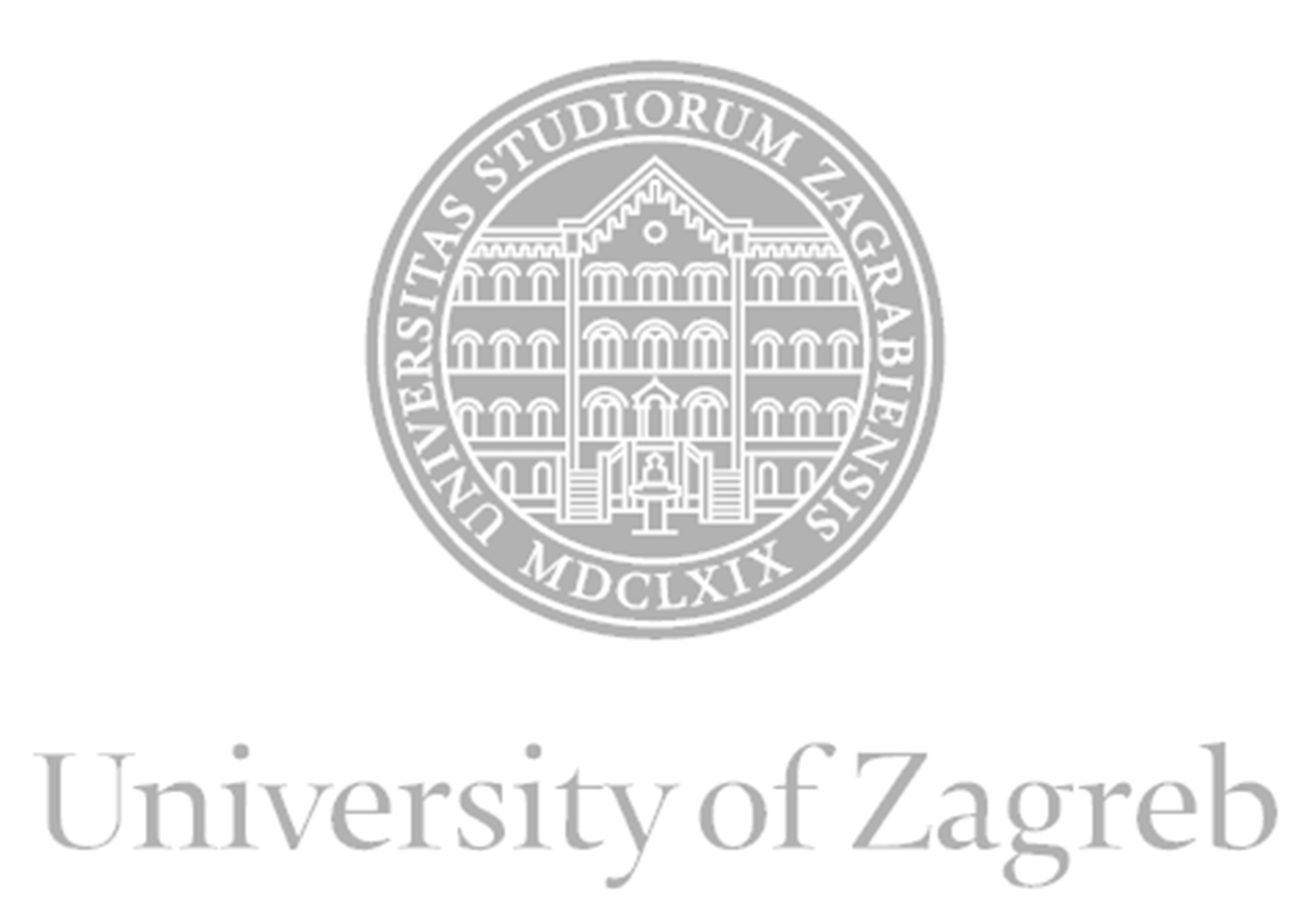}
      \end{center}
    \end{figure}
    \vspace{0cm}
    {FACULTY OF ELECTRICAL ENGINEERING AND COMPUTING} \\
    \vspace{3cm}
    Matej Gjurković \\
    \vspace{2cm}
    {\fontsize{22pt}{22pt}\selectfont\textbf{A COMPUTATIONAL FRAMEWORK FOR INTERPRETABLE TEXT-BASED PERSONALITY ASSESSMENT FROM SOCIAL MEDIA}} \\
    \vspace{2cm}  
    DOCTORAL THESIS \\    
    \vfill{Zagreb, 2025}
  \end{center}
  \restoregeometry
\end{titlepage}

%%%%%%%%%%%%%%% PRVA UNUTARNJA STRANICA / FIRST INNER PAGE %%%%%%%%%%%%%%%%
\begin{titlepage}
  \fontsize{16pt}{20pt}\selectfont
  \fontfamily{phv}\fontseries{mc}\selectfont
  \newgeometry{left=3cm,right=3cm,top=3cm,bottom=2.5cm}
  \setlength{\intextsep}{0pt plus 0pt minus 0pt}

  \begin{center}
    \begin{figure}[ht!]
      \begin{center}
        \includegraphics[height=4.1184cm, width=5.94cm]{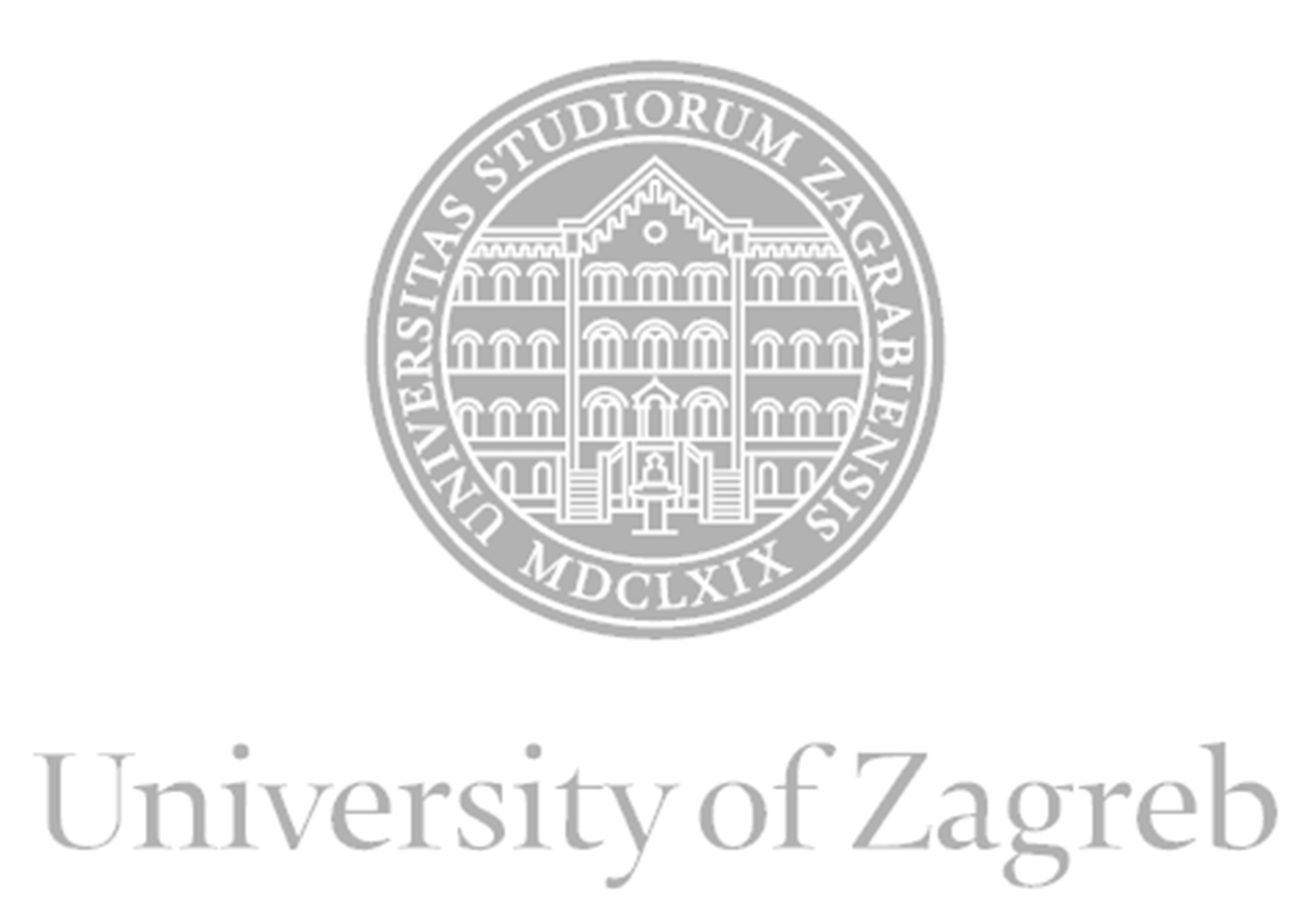}
      \end{center}
    \end{figure}		
    \vspace{0cm}
    {\fontsize{16pt}{16pt}{FACULTY OF ELECTRICAL ENGINEERING AND COMPUTING}} \\
    \vspace{3cm}
    Matej Gjurković \\
    \vspace{2cm}
    {\fontsize{22pt}{22pt}\selectfont\textbf{A COMPUTATIONAL FRAMEWORK FOR INTERPRETABLE TEXT-BASED PERSONALITY ASSESSMENT FROM SOCIAL MEDIA}} \\
    \vspace{2cm}   
    DOCTORAL THESIS \\  
    \vspace{5cm}   % adjust this spacing if necessary
    Supervisor: Professor Jan Šnajder, PhD \\
    \vfill{Zagreb, 2025}
  \end{center}
  \restoregeometry
\end{titlepage}

%%%%%%%%%%%%%% DRUGA UNUTARNJA STRANICA / SECOND INNER PAGE %%%%%%%%%%%%%%%
\begin{titlepage}
  \fontsize{16pt}{20pt}\selectfont
  \fontfamily{phv}\fontseries{mc}\selectfont
  \newgeometry{left=3cm,right=3cm,top=3cm,bottom=2.5cm}
  \setlength{\intextsep}{0pt plus 0pt minus 0pt}

  \begin{center}
    \begin{figure}[ht!]
      \begin{center}
        \includegraphics[height=4.1184cm, width=5.94cm]{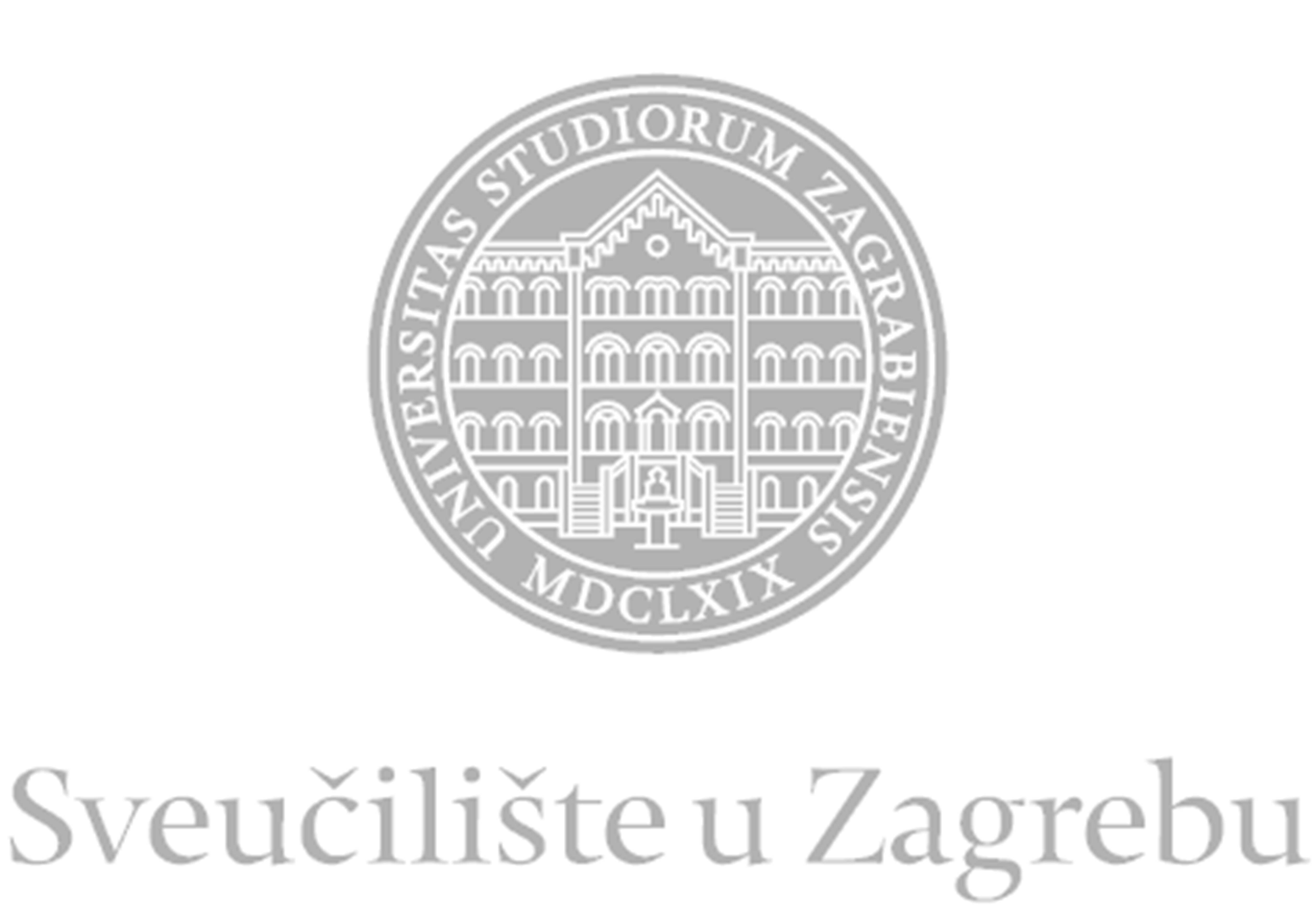}
      \end{center}
    \end{figure}		
    \vspace{0cm}
    {FAKULTET ELEKTROTEHNIKE I RAČUNARSTVA} \\
    \vspace{3cm}
    Matej Gjurković \\
    \vspace{2cm}
    {\fontsize{22pt}{22pt}\selectfont\textbf{RAČUNALNI RADNI OKVIR ZA TUMAČIVU PROCJENU LIČNOSTI NA TEMELJU TEKSTOVA S DRUŠTVENIH MEDIJA}} \\
    \vspace{2cm}    
    DOKTORSKI RAD \\
    \vspace{5cm}    % adjust this spacing if necessary
    Mentor: Prof. dr. sc. Jan Šnajder \\
    \vfill{Zagreb, 2025.}
  \end{center}
  \restoregeometry
\end{titlepage}

%%%%%%%%%%%%%%%%%%%%%%%%%%%%%%%%%%%%%%%%%%%%%%%%%%%%%%%%%%%%%%%%%%%%%%%%%%%
\begin{titlepage}
  \begin{minipage}{\dimexpr\textwidth-1cm}
    \vspace{3cm}
This doctoral dissertation was written at the University of Zagreb, Faculty of Electrical Engineering and Computing,  
at the Department of Electronics, Microelectronics, Computer and Intelligent Systems, in the Text Analysis and Knowledge Engineering Lab (TakeLab).

    \vspace{1cm}
    Supervisor: Prof. Jan Šnajder, PhD

    \vspace{1cm}
    Dissertation length: 145 pages

    \vspace{1cm}
    Dissertation number: \line(1,0){64}
  \end{minipage}
\end{titlepage}

%%%%%%%%%%%%%%%%%%%%%%%%%%%%%%%%%%%%%%%%%%%%%%%%%%%%%%%%%%%%%%%%%%%%%%%%%%%
% insert info page about supervisor which is saved in separate file
\thispagestyle{empty}

\section*{About the Supervisor}

Jan Šnajder has received his BSc, MSc, and PhD degrees in Computer Science from the University of Zagreb, Faculty of Electrical Engineering and Computing (FER), Zagreb, Croatia, in 2002, 2006, and 2010, respectively. From September 2002 he was working as a research assistant, from 2011 as Assistant Professor, from 2016 as Associate Professor, and from 2021 as Full Professor at the Department of Electronics, Microelectronics, Computer and Intelligent Systems at FER. He was a visiting researcher at the Institute for Computational Linguistics at the University of Heidelberg, the Institute for Natural Language Processing at the University of Stuttgart, the National Institute of Information and Communications Technology in Kyoto, and the University of Melbourne. He participated in a number of research and industry projects in the field of natural language processing and machine learning. He is the principal investigator on a HRZZ installation grant project and a HAMAG-BICRO proof-of-concept project, and a researcher on a UKF project. He has (co-) authored more than 100 papers in journals and conferences in natural language processing and information retrieval, and has been reviewing for major journals and conferences in the field. He is the lecturer in charge for six courses at FER and has supervised and co-supervised more than 100 BA and MA theses. He is a member of IEEE, ACM, ACL, the secretary of the Croatian Language Technologies Society, the co-founder and secretary of the Special Interest Group for Slavic NLP of the Association for Computational Linguistics (ACL SIGSLAV). He is a member of the Centre of Research Excellence for Data Science and Advanced Cooperative Systems and the associate editor of the Journal of Computing and Information Technology. He has been awarded the Silver Plaque ``Josip Lončar'' in 2010, the Croatian Science Foundation fellowship in 2012, the fellowship of the Japanese Society for the Promotion of Science in 2014, and the Endeavour Fellowship of the Australian Government in 2015.

\newpage
\section*{O mentoru}

Jan Šnajder diplomirao je, magistrirao i doktorirao u polju računarstva na Sveučilištu u Zagrebu, Fakultetu elektrotehnike i računarstva (FER), 2002., 2006.~odnosno 2010.~godine. Od 2002.~godine radio je kao znanstveni novak, od 2011.~godine kao docent, od 2016.~godine kao izvanredni profesor, a od 2021.~godine kao redoviti profesor na Zavodu za elektroniku, mikroelektroniku, računalne i inteligentne sustave FER-a. Usavršavao se na Institutu za računalnu lingvistiku Sveučilišta u Heidelbergu, Institutu za obradu prirodnog jezika Sveučilišta u Stuttgartu, Nacionalnom institutu za informacijske i komunikacijske tehnologije u Kyotu te Sveučilištu u Melbourneu. Sudjelovao je na nizu znanstvenih i stručnih projekata iz područja obrade prirodnog jezika i strojnog učenja. Voditelj je uspostavnog projekta HRZZ-a i projekta provjere koncepta HAMAG-BICRO-a te je istraživač na projektu UKF-a. Autor je ili suautor više od 100 znanstvenih radova u časopisima i zbornicima međunarodnih konferencija u području obrade prirodnog jezika i pretraživanja informacija te je bio recenzent za veći broj časopisa i konferencija iz tog područja. Nositelj je šest predmeta na FER-u te je bio mentor ili sumentor studentima na više od 100 preddiplomskih i diplomskih radova.

Član je stručnih udruga IEEE, ACM, ACL, tajnik Hrvatskoga društva za jezične tehnologije te suosnivač i tajnik posebne interesne skupine za obradu prirodnog jezika za slavenske jezike pri udruzi za računalnu lingvistiku (ACL SIGSLAV). Član je Znanstvenog centra izvrsnosti za znanost o podacima i kooperativne sustave te je pridruženi urednik časopisa Journal of Computing and Information Technology (CIT). Dobitnik je Srebrne plakete ``Josip Lončar'' 2010. godine, stipendije Hrvatske zaklade za znanost 2012.~godine, stipendije Japanskog društva za promicanje znanosti 2014.~godine te stipendije australske vlade Endeavour 2015.~godine.

%%%%%%%%%%%%%%%%%%%%%%%%%%%%%%%%%%%%%%%%%%%%%%%%%%%%%%%%%%%%%%%%%%%%%%%%%%%
% insert optional page with thanks or dedication
\thispagestyle{empty}

\begin{flushright}
\emph{``He who seeks a goal will remain empty once he reaches it;\
but he who finds the path will always carry the goal within himself.''} --- Nejc Zaplotnik
\end{flushright}

\vspace{1em}

This thesis has been much like my long trail runs, initially bursting with ideas, plans, and the mental equivalent of a runner’s high. Yet, like any real trail, it soon revealed its unique rhythm: steep climbs that felt nearly unclimbable, unexpected turns that led to discovery, and long, quiet stretches where persistence was the only way forward. The reward was a different kind of endorphin, one earned not through motion but through insight, from staying patiently with a question until it revealed its answer.

I am deeply grateful to my supervisor, who did not simply draw the map for me but instead provided the confidence to trust my own navigation and equipped me to make the most of each detour.

My sincere thanks go to my colleagues, who shared the path with me, matching pace when it mattered most and shaping the route through their insights, curiosity, patience, and drive.

Finally, my heartfelt gratitude goes to my family: to my parents and friends for their unwavering support, to my wife who walked every step of this journey by my side, to our daughter who joined us midway and, through this thesis, gently experienced her very first encounter with science, and to our colorful black and white furry friends for their comforting presence.

%%%%%%%%%%%%%%%%%%%%%%%%%%%%%%%%%%%%%%%%%%%%%%%%%%%%%%%%%%%%%%%%%%%%%%%%%%%
% insert page with abstract
\thispagestyle{empty}

\section*{Summary}

Personality refers to individual differences in behavior, thinking, and feeling. With the growing availability of digital footprints, especially from social media, automated methods for personality assessment have become increasingly important. Natural language processing (NLP) enables the analysis of unstructured text data to identify personality indicators. However, two main challenges remain central to this thesis: the scarcity of large, personality-labeled datasets and the disconnect between personality psychology and NLP, which restricts model validity and interpretability.

To address these challenges, this thesis presents two datasets -- MBTI9k and \pandora -- collected from Reddit, a platform known for user anonymity and diverse discussions. The \pandora dataset contains 17 million comments from over 10,000 users and integrates the MBTI and Big Five personality models with demographic information, overcoming limitations in data size, quality, and label coverage.

Experiments on these datasets show that demographic variables influence model validity. In response, the SIMPA (Statement-to-Item Matching Personality Assessment) framework was developed -- a computational framework for interpretable personality assessment that matches user-generated statements with validated questionnaire items. By using machine learning and semantic similarity, SIMPA delivers personality assessments comparable to human evaluations while maintaining high interpretability and efficiency.

Although focused on personality assessment, SIMPA’s versatility extends beyond this domain. Its model-agnostic design, layered cue detection, and scalability make it suitable for various research and practical applications involving complex label taxonomies and variable cue associations with target concepts.

\vspace{1cm}
\noindent\textbf{Keywords}: text-based personality assessment, natural language processing, machine learning, personality psychology
\newpage
\section*{Računalni radni okvir za tumačivu procjenu ličnosti na temelju tekstova s društvenih medija}

Ličnost predstavlja jedan od temeljnih konstrukata u psihologiji koji omogućuje sustavno razumijevanje stabilnih obrazaca mišljenja, emocija i ponašanja pojedinca. U svakodnevnim socijalnim interakcijama, ljudi neprestano i spontano procjenjuju ličnost drugih, pri čemu se oslanjaju na širok spektar signala -- od vizualnih i neverbalnih pokazatelja do govora i jezika. Jezik, pritom, zauzima posebno važno mjesto jer omogućuje verbalizaciju unutarnjih stanja, uvjerenja, stavova i sklonosti, čime posredno i neposredno otkriva osobine ličnosti.

Ova disertacija polazi upravo od te ključne povezanosti jezika i ličnosti te razvija metodološki okvir za automatiziranu procjenu ličnosti iz velikih korpusa tekstnih podataka. Takva procjena poznata je kao automatska procjena ličnosti iz teksta (ATBPA; engl. Automated Text-Based Personality Assessment) i predstavlja rastuće područje istraživanja na sjecištu psihologije ličnosti i obrade prirodnog jezika (NLP; engl. Natural Language Processing), a koja se oslanja na metode strojnog učenja (ML; engl. Machine Learning) i čini jedno od glavnih polja umjetne inteligencije. Ključna prednost ATBPA sustava leži u njihovoj sposobnosti analize velikih količina tekstnih podataka.

Ipak, interdisciplinarna istraživanja u području ATBPA donosi i niz izazova koji proizlaze iz razlika u pristupima i prioritetima različitih disciplina. Dok zajednica obrade prirodnog jezika primarno teži maksimiziranju prediktivne točnosti modela, često koristeći modele temeljene na dubokim neuronskim mrežama, koji uz višu točnost imaju i smanjenu tumačivost, psihologija ličnosti naglasak stavlja na teorijsku utemeljenost, psihometrijsku valjanost i tumačivost procjena. Lingvistika, pak, upozorava na složenost jezičnih fenomena, uključujući semantičke višeznačnosti, očekivane frekvencije pojavljivanja određenih riječi i izraza, diskurzivne kontekste i kulturološke varijacije, koje mogu dovesti do netočnih asocijacija između tragova koji dolaze iz jezika i ličnosti.

Središnji problem sustava ATBPA proizlazi upravo iz složenih međuodnosa između jezičnih značajki, demografskih varijabli i latentnih crta ličnosti. Demografski faktori, poput spola, dobi, kulture i jezične kompetencije, značajno utječu na obrasce jezičnog izražavanja, a istodobno su povezani i s razvojem i distribucijom crta ličnosti. Ovi faktori tako djeluju kao dvosmjerne zbunjujuće varijable u modelima predikcije ličnosti: demografske razlike mogu imitirati razlike u crtama ličnosti i obrnuto -- crte ličnosti mogu utjecati na izražavanje demografskih karakteristika u jeziku. Na primjer, mlađe osobe često koriste moderni sleng i kraće rečenice, što može biti pogrešno tumačeno kao znak impulzivnosti ili niske savjesnosti. Slično tome, žene mogu češće koristiti emocionalno zasićen vokabular, što može biti interpretirano kao viša ugodnost ili neuroticizam, iako to prvenstveno reflektira društveno uvjetovane obrasce komunikacije.
Zbog toga je nužno da modeli za predikciju ličnosti tijekom treniranja imaju pristup demografskim varijablama, što do sada nije bilo moguće budući da postojeći skupovi podataka ne sadrže sve potrebne demografske varijable.

Jedan od dodatnih izazova leži u tehničkoj kompleksnosti modeliranja. Modeli obrade prirodnog jezika karakterizirani su velikim brojem hiperparametara koji imaju znatan utjecaj na performanse modela uključujući izbor značajki, veličinu i vrstu vektorskih reprezentacija, dubinu neuronskih mreža, metode optimizacije, regularizaciju i dr. Zbog toga modeli trenirani na istim podacima mogu dovesti do različitih zaključaka, ovisno o parametrizaciji modela. Ova hiperparametarska osjetljivost dodatno zamagljuje granicu između prediktivne valjanosti i ostalih vrsta valjanosti, što otežava donošenje znanstveno pouzdanih zaključaka o prirodi povezanosti između jezika i ličnosti. Upravo ovi problemi predstavljaju središnju motivaciju za razvoj metodološkog okvira SIMPA unutar ove disertacije.

Kako bi se omogućila temeljitija evaluacija postojećih problema i razvoj novih rješenja, ova disertacija prvo se suočila s ključnim ograničenjem dosadašnjih ATBPA istraživanja -- nedostatkom velikih i demografski obogaćenih korpusa s valjanim oznakama ličnosti autora teksta. Uzrok nedostatka prikladnih skupova dolazi od više povezanih uzroka poput zaštite privatnosti korisnika društvenih mreža, regulative (GDPR) i komercijalnih interesa vlasnika društvenih mreža. Dodatno, svaka platforma s tekstovima korisnika dolazi sa specifičnim karakteristikama koje određuju selekcijske pristranosti uzoraka i korisnika i tekstova. Primjerice, anonimnost autora teksta utječe na to o kojim temama su spremni razgovarati i na koji način, a ograničenja vezana uz duljinu poruka (kao na X-u odnosno Twitteru) utječu na kvalitetu izražaja. Dodatan izazov leži u uparivanju korisnika s rezultatima na testovima ličnosti koji onda služe kao oznake za treniranje modela strojnog učenja. U tu svrhu, izgrađena su dva velika korpusa tekstnih podataka temeljenih na društvenoj mreži Reddit: MBTI9k i PANDORA. Reddit je odabran kao izvorište podataka zbog svojih specifičnih karakteristika: anonimnosti korisnika (koja smanjuje socijalnu poželjnost odgovora), tematske raznolikosti (putem tisuća specijaliziranih subreddita), velikog broja tekstova po korisniku i čestog dobrovoljnog dijeljenja osobnih rezultata testova ličnosti i demografskih informacija.

Korpus MBTI9k temelji se na tekstovima korisnika koji su kroz sustav Reddita deklarirali svoje tipove ličnosti po modelu Myers-Briggs Type Indicator (MBTI). Nakon pažljivog čišćenja, filtriranja i validacije tih oznaka, dobiven je skup od više od 13 000 korisnika i 22 milijuna komentara. Iako MBTI model nije široko prihvaćen u akademskoj psihologiji zbog problematične psihometrijske validnosti, njegova popularnost među korisnicima Reddita omogućila je prikupljanje oznaka za klasifikacijske zadatke na velikoj skali.

Za znanstveno relevantniju analizu razvijen je korpus PANDORA, koji sadrži oznake modela Velikih Pet temeljene na rezultatima različitih upitnika koje su korisnici samostalno objavljivali na Redditu. Iz tih objava, korištenjem regularnih izraza, klasifikacijskih modela i ručne verifikacije, izvučeni su normalizirani rezultati na svih pet crta ličnosti. Korpus sadrži i demografske podatke o spolu, dobi i lokaciji, što omogućuje eksplicitno modeliranje zbunjujućih varijabli u analizama.

Deskriptivne analize ovih skupova otkrile su specifične distribucije osobina u populaciji korisnika Reddita. U korpusu MBTI9k, dominantni su introvertirani, intuitivni, promišljajući i percipirajući tipovi, što odgovara kulturi na Redditu. U korpusu PANDORA, korisnici pokazuju višu otvorenost i nižu ekstraverziju u odnosu na opću populaciju, uz umjerene razlike u savjesnosti, ugodnosti i neuroticizmu. Time su pokazane selekcijske pristranosti na razini korisnika društvene mreže Reddit koje su različite od postojećih skupova podataka koji se temelje na esejima studentske populacije odnosno foruma na kojima se vode razgovori na temu modela ličnosti.

Provedeni prediktivni eksperimenti pokazali su da se crte ličnosti mogu predviđati iz tekstnih značajki uz različite razine uspješnosti. Klasifikacija MBTI dimenzija postizala je makro F1 rezultate između 75\% i 82\% po dimenziji, dok je puna MBTI klasifikacija postizala točnost od 42\%, što znatno nadilazi nasumičnu razinu (25\%). Za crte ličnosti modela Velikih Pet, prediktivne korelacije kretale su se između 0.2 i 0.4, pri čemu je ekstraverzija bila najprediktivnija, a otvorenost, ugodnost i neuroticizam pokazivali su niže korelacije. Jedan od razloga tome jest da se procjena ličnosti uobičajeno radi na razini domena ličnosti, a da su tragovi snažnije povezani s potcrtama domena poput faceta i nijansi. Time je ukazano na potrebu za razvojem novih modela za procjenu na svim razinama taksonomije ličnosti.
Bolje performanse klasifikacijskih modela nasuprot regresijskima motivirale su eksperiment u kojemu se testirala hipoteza da je moguće iskoristiti činjenicu da postoji red veličine više oznaka za model MBTI od modela Velikih Pet. Prvo su trenirani klasifikacijski modeli na skupu korisnika koji su imali samo oznake ličnosti po MBTI, a nakon toga su na izlazima tih modela dotrenirani regresijski modeli za procjenu Velikih Pet crta na skupu korisnika koji su imali oznake oba modela. Konačno, predikcija crta ličnosti na izdvojenom skupu korisnika je pokazala da je moguće napraviti nadzirane modele koji rade predikciju Velikih Pet koji zadovoljavaju konvergentnu valjanost bez tekstova označenih s Velikih Pet.

Daljni eksperimenti pokazali su da su demografske varijable znatno prediktivnije. Klasifikacija spola postizala je točnost veću od 90\%, dok je dob predviđena s korelacijama do 0.7. Ova razlika jasno pokazuje da su demografski signali daleko izraženiji u jeziku od suptilnih psiholoških konstrukata ličnosti, čime dodatno potvrđuju ulogu demografije kao snažnih zbunjujućih varijabli u modelima procjene ličnosti.

Unatoč slaboj prediktivnosti, crte ličnosti pokazale su se kao ključna zbunjujuća varijabla u eksperimentu koji je pokušao ustanoviti postoji li sistemska pristranost klasifikatora spola. Rezultati su ukazali na to da, u slučajevima kada klasifikator donese krivu procjenu, postoji statistički značajni utjecaj crta ličnosti koji je doveo do krive procjene.

Tijekom treniranja modela u predikcijskim eksperimentima utjecaj hiperparametara u treniranju modela pokazao se velikim. Različiti modeli (logistička regresija, SVM, neuronske mreže), različite reprezentacije teksta (n-grami, embeddings, LIWC značajke), kao i razna podešavanja regularizacije i optimizacije dovodila su do varijabilnih rezultata. Takva osjetljivost modela dodatno otežava teorijsku interpretaciju rezultata jer performanse ovise ne samo o kvaliteti podataka već i o tehničkoj parametrizaciji.

Ova ograničenja motivirala su razvoj novog tumačivog računalnog okvira SIMPA (Statement-to-Item Matching Personality Assessment), temeljenog na psihološkoj teoriji Realističkog modela točnosti (RAM). SIMPA dekonstruira procjenu ličnosti u četiri sekvencijalne faze: (1) identifikaciju relevantnih izjava (TRS; engl. Trait Relevant Statement) temeljenih na validiranim upitnicima, (2) procjenu dostupnosti izjava indikativnih za ličnost u tekstovima (TIS; engl. Trait Indicative Statement), (3) detekciju putem semantičkog uparivanja izjava korisnika s upitničkim tvrdnjama koristeći modele semantičke sličnosti te (4) agregaciju bodova za kvantifikaciju crta ličnosti na svim razinama taksonomije ličnosti, nijansi, faceta i domena ličnosti. SIMPA nadograđuje RAM dodavanjem mehanizma povratne sprege u kojem ekspert (model ili ljudski stručnjak) može odabrati pojedine indikativne izjave kao izrazito snažne signale ličnosti te ih promovirati u relevantne izjave.
Realizirana ideja uparivanja relevantnih izjava s izjavama korisnika omogućuje pretvaranje nestrukturiranog teksta u strukturiranu bazu korisničkih izjava označenih vezama prema konstruktima ličnosti na svim razinama taksonomije crta ličnosti, od pozitivne i negativne povezanosti na razini nijansi (pojedinih čestica iz upitnika), koje opisuju specifične obrasce razmišljanja, osjećanja i ponašanja i koje su dalje povezane s facetama i konačno s domenama ličnosti.
Iz ovako strukturiranog skupa označenih izjava moguće je izraditi tumačiv profil ličnosti pojedinih osoba jer je svaka ocjena ličnosti povezana sa specifičnim pronađenim izjavama. SIMPA dodatno omogućuje odbijanje izrade procjene ako nije pronađen dovoljan broj tragova za specifičnu crtu ličnosti. Primjerice, za određenu osobu moguće je dobiti procjenu otvorenosti prema iskustvu na temelju dostupnih dokaza, dok za crtu neuroticizma možda nema dovoljno dokaza. Također, moguće je da ne postoje dokazi za sve facete pojedine domene. Na primjer, moguće je pronaći dokaze (indikativne izjave) za facetu depresivnosti unutar crte neuroticizma, a istovremeno nemati dovoljno tragova za procjenu facete impulzivnosti. Detaljnu analizu dostupnosti tragova moguće je provesti u fazi analize dostupnosti, koja omogućuje analizu ne samo na razini pojedinih autora, nego i na razini čitavog korpusa, primjerice pojedinog podforuma.

Praktična implementacija SIMPA-e provedena je na korpusu PANDORA, koristeći kombinaciju relevantnih izjava preuzetih iz javno objavljenih i validiranih upitnika ličnosti, izjava koje je generirao stručnjak te pomoću izjava koje su generirali veliki jezični modeli. Primijenjeni su modeli semantičke sličnosti trenirani za slične zadatke poput detekcije parafraza, a sustav je evaluiran kroz nadzirane i nenadzirane prediktivne zadatke. Rezultati su pokazali da SIMPA pristup može proizvesti kompetitivne rezultate uz znatno višu razinu tumačivosti i validnosti od metoda tipično korištenih za zadatak predikcije ličnosti na temelju teksta. U aplikaciji nenadzirane procjene ličnosti, eksperimenti su pokazali da značajke temeljene na metodologiji SIMPA, osim poboljšane prediktivne moći, nose dodatne informacije vezane uz tragove ličnosti koje nisu prisutne u značajkama korištenima u prijašnjim istraživanjima.

Daljnja evaluacija uključivala je ljudske i računalne označivače temeljene na velikim jezičnim modelima u zadacima ocjenjivanja crta ličnosti iz minimalnih setova TIS-ova, čime je dodatno demonstrirana primjenjivost SIMPA paradigme u scenarijima u kojima se zahtijeva visoka točnost rezultata. U tom slučaju modeli strojnog učenja predselektiraju skup visoko indikativnih izjava koje ekspertima ubrzavaju procjenu ličnosti. Eksperimenti su dalje pokazali da se veliki jezični modeli mogu koristiti za generiranje i evaluaciju relevantnih izjava, pokazavši potencijal za ubrzavanje razvoja modela za specifične konstrukte. Time je dokazana modularnost i fleksibilnost radnog okvira koji omogućuje zamjenu starije generacije modela u pojedinim dijelovima procesa s novorazvijenim metodama.

Računalni okvir SIMPA adresira ključne slabosti postojećih sustava ATBPA -- uključujući utjecaj demografskih zbunjujućih varijabli, hiperparametarsku osjetljivost modela i nisku tumačivost modela -- kroz modularan i teorijski potkrijepljen pristup koji omogućuje transparentnu procjenu ličnosti na temelju teksta. Premda je fokus disertacije na procjeni ličnosti, radni okvir SIMPA pokazuje svestranost koja nadilazi to područje. Njegova neovisnost o konkretnom modelu, višeslojno prepoznavanje tragova i prilagodljivost velikim skupovima podataka čine ga pogodnim za širok spektar istraživačkih i praktičnih primjena, uključujući složene taksonomije oznaka i varijabilne povezanosti tragova s ciljnim konceptima.

Ova disertacija tako ostvaruje višestruki znanstveni doprinos: (1) izgradnjom bogatih, javno dostupnih i psihološki relevantnih skupova podataka; (2) detaljnom dijagnostikom ograničenja postojećih modela i skupova podataka; (3) razvojem tumačivog i teorijski utemeljenog SIMPA okvira; te (4) demonstracijom njegove primjene kroz eksperimentalne validacije na više razina. Disertacija time postavlja temelje za iduću generaciju automatiziranih sustava procjene ličnosti koji spajaju napredne NLP tehnologije sa znanstvenim standardima psihologije ličnosti.

\vspace{1cm}
\noindent\textbf{Ključne riječi}: tekstna procjena ličnosti, obrada prirodnog jezika, strojno učenje, psihologija ličnosti

%%%%%%%%%%%%%%%%%%%%%%%%%%%%%%%%%%%%%%%%%%%%%%%%%%%%%%%%%%%%%%%%%%%%%%%%%%%
% insert page with extended abstract
% prošireni sažetak na hrvatskom, ako rad nije pisan na tom jeziku
%\include{eg_prosireni_sazetak}
% prošireni sažetak na engleskom, ako rad nije pisan na tom jeziku
% \include{eg_extended_abstract}

%%%%%%%%%%%%%%%%%%%%%%%%%%%%%%%%%%%%%%%%%%%%%%%%%%%%%%%%%%%%%%%%%%%%%%%%%%%
\clearpage
%%%%%%%%%%%%%%%%%%%%%%%%%%%%%%%%% TOC %%%%%%%%%%%%%%%%%%%%%%%%%%%%%%%%%%%%%
\pagestyle{empty} % remove header/footer 
\tableofcontents
\cleardoublepage % start new page

\pagestyle{fancyplain} % puts headers/footers back on

%%%%%%%%%%%%%%%%%%%%%%%%%%%%%%%%%%%%%%%%%%%%%%%%%%%%%%%%%%%%%%%%%%%%%%%%%%%
\mainmatter
%%%%%%%%%%%%%%%%%%%%%%%% POGLAVLJA / CHAPTERS %%%%%%%%%%%%%%%%%%%%%%%%%%%%%
\chapter{Introduction}
Understanding personality is fundamental to human interaction. Whether a teacher is tailoring lessons to students, an employer is selecting new team members, or friends are deepening their bonds, accurately recognizing personality traits is essential in social exchanges. Humans rely on various cues -- visual appearance, speech patterns, and especially language -- to make these assessments. Language serves as a powerful medium through which people express complex thoughts, self-perceptions, and evaluations of others. Words like \emph{kind}, \emph{intelligent}, and \emph{hardworking} illustrate how personality descriptors are inherently embedded in our vocabulary.

This realization that language preserves personality descriptors has become a key to a new scientific discipline -- \textbf{personality psychology}. One early catalyst for this was the lexical hypothesis, credited in part due to Galton \cite{Galton1884} and later advanced by Allport and Odbert \cite{allport1936trait}, Cattell \cite{cattell1943description}, and others. The central idea is that important \textbf{personality traits} become encoded in language over time -- traits that recur across different contexts are referenced by more straightforward, more frequently used words. Subsequent research on the lexical hypothesis found that certain words for describing personality cluster together, paving the way for systematic, scientific approaches to measuring personality. Through statistical analyses of these clusters, researchers identified underlying latent factors that group related descriptors. One of the most prominent findings from this line of research is that consistent patterns of thinking, behaving, and feeling -- what we call personality traits -- reliably emerge across languages and cultures \cite{fiske1949consistency, tupes1958stability}. \textbf{The Big Five} \cite{goldberg1993big} or \textbf{Five-Factor Model} \cite{costa1992neo} represent frameworks for modeling human personality based on these traits. Some researchers prefer a more detailed approach that breaks each core trait into finer facets \cite{costa1991facet}. In contrast, others propose alternative structures that capture additional dimensions like honesty-humility \cite{lee2004psychometric}. Regardless of the framework, the primary concern remains how personality shapes and is reflected in our communication and behavior.

For most laypeople, personality judgment is an everyday function: we see someone smiling and acting kindly and infer they might be open or extraverted. Psychologists, meanwhile, study how these observable cues, such as linguistic style and topic choices, correlate with latent traits. An explicit utterance such as \emph{I’m a highly optimistic person} is a straightforward self-assessment. In contrast, more subtle markers like sentence complexity or word choice can implicitly signal openness, extraversion, or other dimensions. In addition, even small textual changes can drastically alter the perceived meaning. Indeed, there is a significant difference if \emph{someone is psycho-profiling you} or if there is \emph{a psycho profiling you}.

Historically, text was the prime medium for personality-related analysis because it was far easier to store than audio or video. Before widespread digital communication, scholars attempted to identify authors or periods in authorship attribution or stylometry tasks by quantifying word frequencies, sentence lengths, and syntactic roles. As the digital age advanced, \textbf{computational linguistics} and \textbf{natural language processing (NLP)} provided automated techniques, from detecting syntactic structures to directly predicting author traits -- a process we refer to as \textbf{automated text-based personality assessment (ATBPA)}. Increasingly, these approaches rely on \textbf{machine learning (ML)} methods to scale across massive corpora and uncover deeper patterns in textual data.

While large-scale digital data makes personality detection more accessible, new challenges arise. Bridging disciplinary gaps between personality psychology, NLP, and linguistics is one difficulty: NLP researchers often prioritize predictive performance, whereas personality psychologists focus on theoretical validity and psychometric rigor. Linguists, for their part, are keenly aware of laws governing word frequencies and language variation that can confound superficial analyses. Another challenge is that no personality model answers every research question equally well. Broad traits like the Big Five may not capture the fine-grained differences that matter most in specific real-life contexts. Demographics also exert a confounding effect, since characteristics such as gender, age, or cultural background influence language use and personality trait development and distribution. Moreover, data availability poses a significant hurdle. Privacy concerns, business interests, legal frameworks like GDPR, and shifting ethical standards can prevent open access to large-scale labeled corpora. Even where data is available, the quality of personality labels and sociodemographic information may be uneven, leading to biases that cloud research outcomes.

The thesis aims to address these challenges by proposing a computational framework for interpretable personality analysis. One of the main contributions of this thesis is the creation of new datasets built on a novel data source -- the social media platform \textbf{Reddit} -- containing \emph{Big Five} labels alongside demographic information such as gender or cultural background. By explicitly labeling these demographics as potential confounders, the datasets allow a deeper understanding of their effect on personality prediction models and facilitate better bias control. Building on insights from this new data resource, the thesis introduces \textbf{SIMPA}, a computational framework for interpretable personality assessment. SIMPA allows for integrating novel ML techniques while preserving interpretability, enabling researchers to answer different questions about personality expression in text. Combining psychological theory concepts with state-of-the-art ML and a nuanced understanding of language makes it possible to design interpretable and performant models. The following chapters explore these issues in detail, reviewing foundational work in personality and NLP and explaining how SIMPA can be implemented to provide transparent, theoretically grounded personality assessments.

\section{Contribution}

The proposed research aims to address the lack of a personality-labeled dataset of social media texts and develop a method for valid and interpretable automated personality assessment. The dataset labeled with personality labels and demographics allows a deeper analysis of the interplay between language, demographics, and personality. In turn, such a dataset can increase confidence in predictions made by models trained on this data. Moreover, by proposing a novel computational framework, we aim to bridge the two communities of personality psychology and NLP by incorporating the latest developments in NLP while simultaneously providing explicit links on how specific models based on the framework make predictions about personality traits. The original scientific contribution consists of the following:

\begin{enumerate}
    \item A dataset in English containing texts from social media obtained through a computational procedure for enriching the
dataset with the author's personality labels;
    \item A computational framework for interpretable personality assessment based on natural language processing methods for
determining the semantic similarity between statements from social media and personality-relevant statements;
    \item A computational model for personality assessment developed based on the proposed framework and validated on the
developed dataset.
\end{enumerate}

\section{Thesis structure}

This thesis is organized into three parts. Part~\ref{part:one}, \nameref{part:one}, offers the background and context of text-based personality analysis, integrating foundational insights from personality psychology and NLP while surveying prior work in the field. We further establish the research's motivation, methodological underpinnings, and relevance. Specifically, Chapter~\ref{ch:background_nlp} introduces fundamental concepts of NLP and explains the methods commonly employed in previous research on personality assessment from text, along with those used in this thesis. The chapter concludes with an overview of linguistic levels of text analysis and the NLP tasks most pertinent to this study. Chapter~\ref{ch:background_personality} addresses central findings in personality psychology, covering various personality models, demographic influences, and the basic principles of personality assessment. It highlights essential considerations of validity, reliability, and conceptual frameworks for modeling personality judgments. Chapter~\ref{ch:background_text_based_pa} situates text-based personality analysis in its broader research context, describes standard data sources, including the practice of using personality constructs as labels, and surveys seminal contributions in the field.

Part~\ref{part:two}, \nameref{part:two}, explores the development of a large-scale dataset of social media text from Reddit, annotated with personality labels and demographic information. Chapter~\ref{ch:personality_demographics_reddit} details the aims and methods of data collection, including the selection and labeling processes that underpin the dataset. Chapter~\ref{ch:personality_features} examines various feature sets used in subsequent experiments, emphasizing how they replicate and extend approaches previously used in text-based personality assessment. Finally, Chapter~\ref{ch:part_two_experiments} describes experiments designed to examine various issues related to text-based personality analysis from social media, ranging from confounding demographic effects to creating practical-oriented prediction models.

Part~\ref{part:three}, \nameref{part:three}, proposes a theoretical framework for interpretable personality assessment based on social media text. Chapter~\ref{ch:simpa} presents the Statement-to-Item Matching Personality Assessment (SIMPA) framework, which enables personality assessment at multiple levels of personality taxonomies and systematically decomposes the assessment process into four stages. Chapter~\ref{ch:implementation} demonstrates how this framework can be instantiated as multiple working models, underscoring its capacity to remain relevant in a rapidly changing landscape of natural language processing methods while preserving interpretability.

Finally, Chapter~\ref{ch:conclusion} concludes the thesis by synthesizing the key contributions of this research, acknowledging its limitations, and discussing the resulting implications for future research in text-based personality assessment.

\part{Background}
\label{part:one}

This part aims to define essential concepts and findings in natural language processing (NLP) and personality psychology, setting the stage for research conducted in this thesis and its contribution to text-based personality analysis. We will begin by exploring the conceptual and historical progression of NLP methods used in author profiling and personality analysis. 
Next, we will briefly introduce sociolinguistics and provide insights into the relationship between language generation and analysis and the characteristics of text authors. We will discuss the significance of demographic variables that must be considered in such analyses.
Subsequently, we will contextualize personality as one such variable and provide fundamental insights from personality psychology that are essential for the remainder of this thesis. Finally, we will review existing approaches in text-based personality analysis, focusing specifically on the research gaps this thesis aims to address.

\chapter{Natural language processing}
\label{ch:background_nlp}

With the growth of digital communication and the increasing availability of large text data sets, manual text analysis has become impractical. Text data is often complex and ambiguous, making manual analysis time-consuming and prone to error. Computational methods can automate and streamline the analysis process, reducing the potential for human error and increasing the reliability of results. They also provide a consistent and reproducible approach to text analysis, particularly important in scientific research and data-driven decision-making.

Natural language processing (NLP) is a branch of artificial intelligence (AI) and computer science (CS) that deals with the interaction between computers and humans using natural language. NLP algorithms and models often rely on machine learning (ML) methods to help analyze and understand human language, allowing computers to process, analyze, and generate text to mimic human communication. As one of the main areas of research in AI, NLP tackles challenges in a wide array of tasks, ranging from practical ones such as information retrieval or information extraction to more theoretical understanding of the model's inner workings.

The evolution of NLP is reflective of broader trends in ML development. Initially, NLP predominantly relied on rule-based and domain-specific methods, which were labor-intensive and challenging to adapt and maintain. The integration of ML signified a paradigm shift, with systems learning to identify patterns automatically. However, this era still necessitated substantial expert involvement in feature engineering or defining task-specific patterns in text, which were manually crafted and fed into specialized ML algorithms for classification, regression, or generation tasks.

The development of deep learning (DL) methods \cite{GoodBengCour16} ushered in a new era by eliminating the step of manual feature engineering. With these techniques, models could be trained end-to-end, processing raw text as input and directly producing the desired task-specific outputs. However, DL models are more computationally demanding and rely on massive amounts of data, which led to the development of a transfer learning (TL) paradigm \cite{pan2010transferlearning}. The main idea behind TL is to train general-purpose models on large datasets and adjust their weights or fine-tune them for a particular task. The other deficiency of some DL methods, such as recurrent neural networks (RNNs) \cite{elman1990finding}, is that their training cannot be easily parallelized, and they are incapable of modeling long-range dependencies. This motivated the invention of transformer networks \cite{vaswani2017attention}, which could model longer text sequences while enabling the parallelization necessary for efficient high-scale training. Transformers and the transfer learning paradigm have recently revolutionized the field, achieving results comparable to human performance on numerous tasks, even with minimal training data for specific tasks. They form the foundation of two prominent model families: encoder-only and decoder-only models. Encoder-only models, such as BERT \cite{devlin-etal-2019-bert}, utilize a Masked Language Modeling (MLM) strategy, which allows them to capture context bidirectionally. This capability significantly enhances their performance on language understanding tasks. In contrast, decoder-only or autoregressive models employ a Causal Language Modeling (CLM) strategy. They are trained to predict the next token in a sequence of tokens, enabling their generative capabilities, which are essential for tasks such as dialogue systems.
In combination with post-training methods such as supervised instruction fine tuning \cite{ouyang2022training} and alignment (e.g., reinforcement learning with human feedback (RLHF) \cite{christiano2017deep}), transfer learning enabled the development of an increasing number of high-performance generative models, such as ChatGPT \cite{openai2023chatgpt}, which can successfully solve many tasks instructed only by natural language prompts. 

\section{Traditional machine learning}
\label{section:background_trad_ml}

NLP relies on machine learning techniques that automatically learn patterns from data without being explicitly programmed. It relies on the idea that a computer program can learn from experience, much like a human.
In most cases, central to this process is data representation in numerical formats, referred to as \emph{features} within ML terminology, which enables computational systems to process and analyze information effectively. In the domain of NLP, language's discrete and sequential nature presents unique challenges for feature representation.

\subsection{Features}
\label{section:background_features}
Feature engineering is a process of transforming raw text into numerical representations that ML models can effectively process. This process typically begins with tokenization, where the text is divided into individual words or tokens, followed by lowercasing to ensure consistency. Stemming or lemmatization is then often applied to reduce words to their base forms, unifying different morphological variants of the same word. The pipeline can include removing stop words, common words that contribute minimal semantic value, and strategies for handling out-of-vocabulary (OOV) words, such as replacing them with a special <UNK> token.

To capture contextual and morphological information, n-grams can be used, both word-level and character-level. Word n-grams capture sequences of adjacent words, allowing the model to recognize common phrases and contextual dependencies, while character n-grams capture subword patterns, which are particularly useful for handling misspellings, morphological variations, and languages with rich inflection. Although these approaches enrich the feature set by providing a more nuanced understanding of the text, they also contribute to increased dimensionality.

One of the simplest methods for representing words is one-hot encoding. In this method, each word in the vocabulary is represented as a binary vector where only the position corresponding to that word is set to one, and all other positions are zeros. For example, in a vocabulary of three words, ``cat'', ``dog'', and ``mouse'', the word ``dog'' would be represented as [0,1,0]. Although straightforward, one-hot encoding has notable drawbacks: it leads to high-dimensional and sparse vectors and fails to capture meaningful semantic relationships between words.

There are several ways to encode words as features in NLP other than one-hot encoding, as shown in Table~\ref{table:feature_representation}.
One approach is to encode frequency, as in features such as the Bag of Words (BoW) and Term Frequency-Inverse Document Frequency (TF-IDF). BoW counts the occurrences of each word in a document, while TF-IDF weighs these counts by the inverse frequency of the words across the entire corpus, thereby emphasizing more informative terms. Although these methods incorporate term importance and reduce sparsity, they still result in high-dimensional and sparse representations, especially with large vocabularies. This exacerbates the curse of dimensionality, a phenomenon in which the feature space becomes so vast that the model's performance deteriorates due to overfitting and computational inefficiency.

Dimensionality reduction techniques such as Principal Component Analysis (PCA) \cite{jolliffe2002principal} are often employed to mitigate the challenges of high dimensionality and sparsity. PCA transforms the original high-dimensional feature space into a lower-dimensional one by identifying the principal components that capture the most variance in the data. This reduction improves computational efficiency and helps alleviate the curse of dimensionality by eliminating redundant and less informative features, thus enhancing the model's generalizability.

Advancing beyond these methods, the field of NLP has transitioned to continuous word representations known as \emph{word embeddings} \cite{jm3}. Unlike one-hot encoding and frequency-based features, word embeddings map each word to a dense, low-dimensional vector in a Euclidean space. This approach allows words with similar meanings to be positioned close to each other in the vector space, effectively capturing semantic relationships while avoiding high dimensionality and sparsity. To produce embeddings, models such as Word2Vec \cite{mikolov2013distributed} and GloVe (Global Vectors for Word Representation) \cite{pennington2014glove} rely on the main idea of distributional semantics, which is that words with similar meanings are used in similar contexts. Word2Vec employs neural network architectures to learn embeddings by predicting a target word based on its surrounding context (Continuous Bag of Words) or by predicting the context from a target word (Skip-gram), thereby capturing word relationships without the need for labeled data. In contrast, GloVe utilizes global word co-occurrence statistics from large text corpora to generate highly informative vectors by combining global matrix factorization with local context window methods.

An important aspect of feature representation in NLP is the distinction between \emph{closed vocabulary} and \emph{open vocabulary} approaches. 
A closed vocabulary approach relies on a fixed list of words to represent text. This method simplifies processing and often reduces computational demands because the model only handles known words. Additionally, since the vocabulary is predefined, it is typically easier to interpret the resulting features -- particularly when domain experts curate the list of words or phrases relevant to a specific task. Expert-driven features can leverage in-depth specialist knowledge, making them closely aligned with task requirements and transparent to users.
However, closed-vocabulary systems face notable drawbacks. Any word outside the predefined set -- often referred to as an out-of-vocabulary (OOV) word -- will be replaced with a generic placeholder (e.g., <UNK>), causing the loss of potentially significant semantic information. Another limitation appears with frequency-based features (e.g., term counts), which can be misleading if the domain or data source changes. This issue is influenced by Zipf’s law -- which describes how a few words are used extremely often while most are rare -- and by Brevity law, relating to the length of expressions \cite{zipf1949human, zipf1945psychobiology}. Preprocessing steps such as stemming, lemmatization, and tokenization further affect the reliability of these features. Hence, although closed-vocabulary approaches can be efficient and interpretable, they often lack robustness when facing new words, varied language styles, or domain shifts.

By contrast, an open vocabulary approach uses more flexible units -- such as characters, word n-grams, or subword segments (e.g., Byte Pair Encoding \cite{sennrich-etal-2016-neural} or WordPiece \cite{schuster-nakajima-2012-japanese}) -- to represent text. With these methods, the model is not restricted to a fixed set of terms. Instead, it can adapt to previously unseen words by breaking them down into smaller subunits. This strategy preserves critical semantic cues, even for rare, domain-specific, or newly coined expressions. Consequently, open-vocabulary approaches are more resilient to linguistic variation, though they can introduce higher computational and model complexity.

\begin{table}[h]
\centering
{\small
\renewcommand{\arraystretch}{1.2} % Adjusts row height for better readability
\caption{Typical feature representation methods in NLP}
\begin{tabular}{@{}l|p{5cm}@{}}  % @{} removes extra spacing on the left
\hline
\textbf{Feature group} & \textbf{Description} \\
\hline
Bag of Words (BoW) & Represents text by counting the frequency of each word, disregarding grammar and word order but keeping multiplicity. \\
\hline
Bag of N-grams & Extends BoW by capturing sequences of $n$ words, providing local context missing in the BoW model. \\
\hline
Term Frequency (TF) & Counts the frequency of each word in a document, quantifying its importance based on occurrence. \\
\hline
Term Frequency-Inverse Document Frequency (TF-IDF) & Combines term frequency with inverse document frequency to highlight words that are unique to specific documents. \\
\hline
Part-of-Speech Tagging & Utilizes grammatical categories (e.g., noun, verb) to represent morphosyntactic properties. \\
\hline
Categorical Word Lists & Groups words into predefined semantic or syntactic categories, useful in sentiment analysis and text classification. \\
\hline
Continuous Word Representations (Word2Vec, GloVe) & Represents words as vectors in a high-dimensional space, capturing semantic relationships. \\
\hline
\end{tabular}
\label{table:feature_representation}
}
\end{table}

\subsection{Prediction models}
\label{section:background_pred_models}
Once the features have been extracted from the text, they are fed into a ML model. This is typically a classifier that predicts class labels or a regression model that forecasts numerical outputs. The goal of a classifier is to learn how to map the input features to the class labels. The most common type of classifier is a binary classifier, which predicts one of two class labels (e.g., positive or negative). In contrast, a multiclass classifier can predict one of several class labels.
Logistic regression is a popular traditional ML model for binary classification problems, which models the relationship between input features and the binary class label as a logistic function. The logistic function maps the input features to a probability value between 0 and 1, representing the likelihood that the data sample belongs to a class.
%P(Y=1 | X) = \frac{1}{1 + e^{-(\beta_0 + \beta_1 x_1 + \beta_2 x_2 + x\cdots + \beta_n x_n)}}
Another commonly used model is the Support Vector Machine (SVM)  \cite{cortes1995support}, a classical machine learning model for binary and multiclass classification problems. It works by finding a hyperplane that separates the data samples into different classes. The hyperplane is chosen to maximize the distance between the closest examples of different classes, making the SVM model robust to outliers and noise in the data. Moreover, due to its dual form and kernel tricker, the SVM can deal with non-linear problems and high-dimensional dack. The kernel trick implicitly maps the input space into a higher-dimensional feature space without explicitly computing the dimensions in this space. The way the dual form is defined allows only a subset of the training data (the support vectors) to determine the decision boundary, making the model more efficient and easier to interpret.

A supervised learning ML algorithm is given labels for data points and then learns the mapping between the two. In contrast, in unsupervised learning, a machine learning algorithm is used to draw inferences from datasets consisting of unlabeled input data. Essentially, unsupervised learning focuses on identifying patterns in the data without the guidance of a known outcome variable or reward function. The most common use of unsupervised learning is clustering, which involves grouping objects so that objects in the same group (or cluster) are more similar than those in other groups. K-means clustering is one of the most popular and simple clustering algorithms. It partitions the data into K clusters, where each observation belongs to the cluster with the nearest mean. The algorithm iteratively assigns data points to one of the K groups based on the provided features.

%\todo{add PCA and dimensionality reduction}

%\todo{Eval measures for classification and regression, confusion matric, accuracy f1, macro/micro, person, spearman}

\section{Deep learning}
\label{section:background_dl}
%\todo{shorten, just explain lstms and cnns in short and then go to transformer based models without gnns}

Deep learning techniques aim to overcome the limits of traditional machine learning approaches. Unlike their predecessors, these models can automatically learn features directly from the data through an end-to-end training process using backpropagation \cite{rumelhart1986backprop}. Early deep learning models, such as Long Short-Term Memory (LSTM) networks \cite{hochreiter1997long} and Convolutional Neural Networks (CNNs) \cite{lecun1998gradient}, gained prominence in Natural Language Processing (NLP) tasks, including sentiment analysis and machine translation. LSTMs excelled at capturing sequential dependencies, making them particularly suitable for text processing, while CNNs were effective at extracting local features and patterns, contributing to their adoption in various NLP applications. These models excel at detecting complex patterns by utilizing multiple layers of non-linear transformations, which enables them to capture hierarchical and abstract representations of data. This capability has led to significant performance improvements in NLP tasks compared to traditional models that often rely on manually crafted features or shallow relationships. They become particularly effective when trained on large labeled datasets and when leveraging unlabeled data through techniques such as semi-supervised and unsupervised learning, making them versatile for a wide range of NLP applications.

However, DL models require substantial computational resources to train due to their typically large number of parameters and the vast amounts of data involved. This has led to the development of techniques that reduce the need to build a new model for each task from scratch. One such technique is multitask learning \cite{caruana1997multitask}, which is based on the concept that similar tasks might share underlying features. By training models on multiple tasks simultaneously -- especially when ample labeled data is available -- these techniques can improve performance on tasks with limited resources. Multitask learning shares the same goal as transfer learning, which focuses on the sequential reuse of learned features from a source task to boost performance on a target task.

Recurrent Neural Network (RNN) models are a type of DL model that is particularly well-suited for NLP tasks that involve sequential data, such as language. RNN models maintain an updated internal state with each new input, allowing them to capture dependencies between previous and current inputs. This allows models to encode longer sequences of text into a fixed vector. Long Short-Term Memory (LSTM) models are a type of RNN specifically designed to allow for more extended context by mitigating the vanishing gradient problem in standard RNN models. LSTMs achieve this by introducing memory cells and gating mechanisms that control the flow of information through the network.

Bidirectional Long-Short-Term Memory (biLSTM) \cite{graves2005framewise} models are a variant of LSTM models that process the input sequence in forward and backward directions. This allows biLSTMs to better capture dependencies between future and past inputs, making them well-suited for tasks such as named entity recognition and text classification.

\subsection{Transformers and attention mechanism}
\label{section:background_transformers}
The Transformer model \cite{vaswani2017attention} revolutionized NLP by placing the attention mechanism at its core. This mechanism allows models to dynamically focus on different parts of the input sequence when producing the output for a specific task. Unlike traditional sequential architectures, such as RNNs, Transformers process entire sequences in parallel, enabling them to capture contextual relationships more effectively.

The attention mechanism operates by comparing \textit{queries}, \textit{keys}, and \textit{values}. In this framework, \textit{queries} represent the information sought, \textit{keys} serve as markers that identify parts of the sequence, and \textit{values} contain the actual content associated with the \textit{keys}. By calculating similarity scores between \textit{queries} and \textit{keys}, the mechanism determines the significance of different elements, allowing the model to weigh their contributions accordingly.

Importantly, the attention mechanism enables models to perform semantic composition, combining the meanings of individual tokens or words to construct the overall meaning of the entire sequence. An extension of this is multi-head attention, which allows multiple attention layers to function in parallel, focusing on various aspects of the input, such as local relationships, long-range dependencies, and semantic or syntactic nuances.

Through iterative application across layers, Transformers develop contextualized representations of input sequences. It has been shown \cite{jawahar-etal-2019-bert, tenney-etal-2019-bert} that earlier layers capture local relationships, while deeper layers extract more abstract patterns. Although this design facilitates parallelization and efficient scaling, the quadratic complexity of attention with respect to sequence length presents computational challenges, particularly with longer input. To address these issues, various adaptations, such as local and sparse attention, have been proposed to balance complexity and performance.

\subsection{Transfer learning}
\label{section:background_tl}
Transfer learning (TL) \cite{pan2010survey} in ML involves using knowledge acquired from one domain to solve problems in a related domain. Although the concept has been known for decades \cite{caruana1997multitask}, it has gained significant importance with the advancements in DL models, as well as the increasing availability of data and computational resources. TL typically consists of two main stages: pre-training and fine-tuning.

In NLP, deep neural networks often undergo a pre-training phase where they are trained on a large general-purpose language corpus.  This stage allows the model to acquire a broad understanding of linguistic patterns and features that are applicable across a wide range of tasks. More importantly, pre-training enables the model to acquire general linguistic competence and world knowledge, providing a strong foundation for downstream tasks. After this, the model is fine-tuned or adjusted using smaller, task-specific datasets. Fine-tuning refines the model to address the particular nuances and requirements of the new task while retaining the general knowledge acquired during pre-training.

This approach is highly effective for utilizing large datasets and generally outperforms traditional ML algorithms. TL is particularly beneficial in situations where the dataset for a specific task is too small to train a DL model from scratch, as it allows the model to leverage previously learned patterns and representations. By combining general pre-training with task-specific fine-tuning, TL has become a foundational method in NLP, enabling models to achieve state-of-the-art performance across a variety of applications.
%Additionally, the advent of semantic search technologies, which use NLP to understand the intent and contextual meaning behind queries, has further enhanced the applicability of transfer learning in various domains. 

\subsection{Language models}
\label{section:background_lm}
Language models have been around in NLP since its early days \cite{manning1999foundations}. These statistical models assign probabilities to sentences and predict the most probable next word given a sequence of words. Most pre-training procedures focus on solving the most general task of language modeling: predicting the next token in a sequence of text. To solve this task, models must learn many language structures, ranging from morphology and syntax to semantics and pragmatics. There are several approaches to pre-training models, with three predominant strategies: Masked Language Models (MLMs), Causal Language Models (CLMs), and Sequence-to-Sequence (Seq2Seq). MLMs, such as BERT (Bidirectional Encoder Representations from Transformers), are designed to improve language understanding by predicting masked input segments. This approach enables them to learn contextual relationships between words in a text. During training, random tokens in the input sequences are masked, and the model is then tasked with reconstructing the original tokens, thereby improving its understanding of language context.

On the other hand, CLMs -- as exemplified by models like GPT (Generative Pretrained Transformer) \cite{radford2018improving} -- are oriented toward predicting the next word in a sequence. Although the term ``causal'' is widely used in the literature, it is important to note that these models, like MLMs, are fundamentally statistical in nature, relying on patterns in the data rather than capturing any intrinsic causal structure of language. The CLM pretraining strategy is particularly suited for tasks involving text generation and sequential token prediction.

Following the encoder-based MLMs and decoder-based CLMs, Seq2Seq models represent another approach to pretraining language models combining encoders and decoders. In Seq2Seq pretraining, the encoder reads the input data and encodes it into a context vector, capturing the input information's essence. The decoder then uses this context to produce the output sequence. This approach is beneficial for tasks that involve converting one form of data into another, like translation, summarization, or question-answering.
The models based on Seq2Seq architecture include T5 (Text-to-Text Transfer Transformer) \cite{raffel2020exploring} and BART (Bidirectional and Auto-Regressive Transformers) \cite{lewis-etal-2020-bart}.

Although a definitive threshold does not exist, a language model with a large number of parameters, typically more than 1 billion, trained using the aforementioned procedures is often referred to as a Large Language Model (LLM).

%The first widely adopted large language model is called BERT (short for Bidirectional Encoder Representations from Transformers; Devlin et al., 2019). Released for open use by Google in 2018, BERT has been followed by a family of large language models, including RoBERTa (Liu et al., 2019), GPT3 (T. B. Brown et al., 2020), and XLNet (Yang et al., 2019).

%\todo{add encoder vs decoder architecture}

\subsection{Fine-tuning}
\label{section:background_ft}

Fine-tuning is an optimization technique employed to adapt a pre-trained model for specific tasks, allowing it to leverage its extensive knowledge base while enhancing performance on specialized applications. In the case of LLMs, fine-tuning is typically performed during post-training, often in conjunction with alignment procedures to ensure that the model's outputs conform to desired behaviors. Recent advancements in parameter-efficient fine-tuning (PEFT) \cite{houlsby2019parameter} offer a more resource-conscious method for customizing these large-scale models for particular tasks without requiring comprehensive retraining. While this post-training fine-tuning and alignment is a hallmark of LLMs, other pre-trained language models (PLMs), such as BERT \cite{devlin-etal-2019-bert} and its variants, also undergo fine-tuning for downstream tasks, albeit generally without the explicit focus on alignment seen in LLMs.

At the core of PEFT is the insertion of a modest set of trainable parameters at critical points within the model's architecture. These parameters are then fine-tuned through backpropagation while most of the model's weights remain unchanged. This approach allows targeted adjustments that significantly improve task-specific performance without compromising the integrity or the general-purpose knowledge inherent in the original model.

Recently, two additional fine-tuning techniques were introduced to create instruction-based and chat-based LLMs: supervised instruction fine tuning \cite{ouyang2022training} and alignments methods such as RLHF \cite{christiano2017deep}. 
Former uses pairs of instructions and desired completions in the fine-tuning process, which helps models learn how to follow specific instructions. It is often followed by an RLHF algorithm that aligns models with human-preferred answers. 
Reinforcement learning is usually implemented using Proximal Policy Optimization (PPO) \cite{schulman2017proximal}, designed to optimize an agent's policy function to maximize its expected cumulative reward in a given environment. 
The more novel and computationally more efficient approach is based on Direct Preference Optimization (DPO) \cite{rafailov2024direct}, which uses a dataset of human preference pairs containing a prompt and two possible completions -- one preferred and one dispreferred. The LLM is then fine-tuned to maximize the likelihood of generating preferred completions and minimize the likelihood of generating dispreferred ones.

\subsection{Prompt engineering}
\label{section:background_prompt}
Prompt engineering has become an integral component in the effective use of LLMs due to its straightforward and intuitive nature. This technique uses natural language inputs to condition generative models, guiding them to produce specific outputs. It is a versatile approach applicable to various tasks like classification, translation, and information extraction. The most basic scenario is a \textit{zero-shot} setup, where the model is prompted to solve a task without examples to guide them to the best answer. A related approach \textit{few-shot learning} is a technique in which the model is presented with examples of inputs and desired outputs, which helps the model solve the task more efficiently and generate responses in the desired form. There are also more advanced techniques like \textit{learning to prompt}, where models are trained to determine the most effective prompt structures. The latter methods rely on \textit{in-context learning} (ICL) \cite{dong2024survey}, which involves creating prompts that provide relevant context to steer the model's output in a particular direction without updating the model's weights.
The \textit{chain-of-thought} (CoT) \cite{wei2022chain} prompting technique demonstrated that it is possible to enhance the reasoning capabilities of models by incorporating intermediate steps in the reasoning process. This technique can be compelling when combined with few-shot prompting, as it significantly improves performance on complex tasks that require a layer of reasoning before generating a response. Following this line of research, more advanced techniques like the \textit{Tree of Thought} (ToT) \cite{yao2024tree} have been developed.

\section{The linguistic levels of text analysis}
\label{section:background_linguistic_levels}
LLMs demonstrate their linguistic competence not only by producing high-quality text but also by accurately analyzing input text. This section highlights that language is more than a simple concatenation of words and can be analyzed in-depth at different linguistic levels. These linguistic levels provide a structured framework that allows focus on distinct aspects of language's structure, interpretation, and usage. Table~\ref{table:linguistic_analysis} summarized the descriptions and choice of features at all levels of linguistic analysis.

Given our emphasis on text, we will set aside the sound-centric levels of phonetics and phonology, instead concentrating on the higher levels of linguistic analysis. These include morphology, syntax, semantics, and pragmatics, which are integral to spoken and written language. We will briefly mention sociolinguistic aspects of the analysis for each level, as they provide context for the experiments in this thesis. 

\begin{table}[h]
\small
\centering
\caption{Description of linguistic analysis levels with standard features. Contextual embeddings are used in both semantic and pragmatic tasks, as they assist with polysemy disambiguation (semantics) and capturing meaning in context (pragmatics).}
\begin{tabular}{l|p{5cm}|p{5cm}}
\hline
\textbf{Linguistic level} & \textbf{Description} & \textbf{Features} \\
\hline
Phonetics & The study of speech sounds and their production and acoustic properties. & Acoustic models, phonetic algorithms, speech recognition features. \\
\hline
Phonology & The study of phonemes and the systematic organization of sounds in languages. & Phoneme recognition, prosodic analysis, stress pattern detection. \\
\hline
Morphology & The study of the internal structure of words and the rules for word formation. & Morpheme segmentation, part-of-speech tagging, morphological parsing. \\
\hline
Syntax & The study of the rules and principles for constructing sentences in languages. & Parse trees, grammatical tagging, syntactic dependency analysis. \\
\hline
Semantics & The study of meaning in language, concerning words, phrases, and sentences. & Semantic parsing, word sense disambiguation, entity linking, contextual embeddings. \\
\hline
Pragmatics & The study of language use in context and the interpretation of meaning beyond the literal content. & Contextual embeddings, discourse analysis, intention and implicature modeling. \\
\hline
\end{tabular}
\label{table:linguistic_analysis}
\end{table}

\subsection{Morphology}
Morphology entails a study of word formation and structural variation. This domain focuses on how words undergo inflection, derivation, and compounding to articulate grammatical categories such as tense and number. 
%It also encompasses the investigation of irregular word formations, which are often aspects of discerning linguistic patterns and norms.

In NLP, the challenge lies in accurately modeling these morphological characteristics, as they are preconditions for tasks on higher linguistic levels of analysis, such as syntax and semantics. NLP algorithms must effectively recognize and interpret the semantic implications of various word forms, a task complicated by the linguistic diversity and distinct morphological rules inherent to different languages.

Morphological variations often mirror regional dialects and identities of social groups, and their analysis can reveal the underlying sociodemographic variables \cite{nguyen2016computational}. In morphologically rich languages such as Croatian, these can directly disclose important personal characteristics like the gender of the author of the text.

\subsection{Syntax}
Syntax refers to the study of word arrangement to form coherent sentences. It involves understanding grammatical rules, sentence structures, and the relationships among various sentence components. The complexity of syntax is attributed to the wide array of sentence structures across languages and frequent deviations from standard grammatical conventions.

Syntactic parsing is a fundamental task in NLP and involves the analysis of a sentence's structure according to the language's grammar rules. The parse, or syntax tree, is the primary output of syntactic parsing, which represents the syntactic structure of a sentence. The tree shows how words in a sentence are grouped and ordered in a hierarchy and how they relate.
Parsing helps in understanding the grammatical structure of a sentence. This includes identifying parts of speech (POS, e.g., nouns, verbs, adjectives), phrase structures (e.g., noun phrases, verb phrases), and clause structures.
For some tasks in personality research, POS tagging is important as a means to increase precision. For instance, the word \textit{kind} can function as either an adjective or a noun: in the phrase \textit{human kind}, it serves as a noun, which alters its meaning compared to when it is used to describe someone as an adjective.
Furthermore, syntactic parsing identifies the constituents of a sentence or groups of words that function as single units. For example, in the sentence ``The quick brown fox jumps over the lazy dog'', a syntactic parser can identify ``The quick brown fox'' as a noun phrase and ``jumps over the lazy dog'' as a verb phrase.
Parsing also helps to understand the relationships between words in a sentence, such as subject-verb-object relationships. It can serve to disambiguate sentences that may be ambiguous when only words are considered without their syntactic roles.
Dependency parsing is a type of syntactic parsing that identifies dependency relationships between words, showing which words syntactically depend on others in the sentence.

In the context of ATBPA, the ability to analyze text syntax is important. Syntactic choices often reflect the educational background or social status of the text's author. For example, more educated authors tend to use more complex syntactic patterns \cite{lu2010automatic}. Moreover, the syntax can help reveal whether the author is writing in their native language or even identify the author's native language group \cite{koppel2002automatically}.

\subsection{Semantics}
Semantics is the branch of linguistics concerned with the meaning of words and phrases. Semantics confronts challenges such as polysemy, where words possess multiple meanings, and the distinction between context-dependent connotative and denotative meanings.

Among the many frameworks developed in linguistics and computational linguistics, distributional semantics \cite{lenci2024distributional} has become one of the most important approaches in deep learning-based NLP. Distributional semantics is a theory and method that analyzes and models the meanings of words based on their distributional properties in large language corpora. The central idea of distributional semantics is that the meaning of a word can be inferred from the contexts in which it appears. Words that occur in similar contexts are considered to have similar meanings. For example, ``apple'' and ``orange''  might appear in similar contexts (like ``ate an apple'' or ``juicy orange'') and hence are considered semantically similar to some extent.

In practice, distributional semantics often involves representing words as vectors in a high-dimensional space, as described in Section~\ref{section:background_features}. Traditionally, this has been accomplished using static embeddings, such as Word2Vec and GloVe, which assign a single, fixed vector to each word regardless of its context. However, this approach struggles with challenges like polysemy -- where a word has multiple meanings -- and homonymy -- where words share the same spelling but have different meanings.
To address these limitations, models utilize contextualized embeddings \cite{peters2018deep, devlin-etal-2019-bert}, which are generated by sequence-processing models like RNNs and Transformers. These models produce representations that vary depending on the surrounding context, enabling them to disambiguate word senses effectively. 

This evolution is particularly important in applications such as ATBPA, where accurately capturing subtle distinctions in meaning can increase the variability of expressions of differences at the individual and group levels.

\subsection{Pragmatics}
Pragmatics addresses how context impacts language interpretation, focusing on aspects such as indirect meanings, sarcasm, and cultural references. It involves the literal content of communication and its interpretation within specific situational contexts.

%Pragmatics in NLP relates computational modeling of how context and situational factors influence language interpretation. Unlike semantics, which focuses on the literal meaning of words and sentences, pragmatics is concerned with the inferred meanings and intentions behind language use, often shaped by the social and communicative context.

In NLP, pragmatics is related to all subjective tasks or tasks involving multiple participants, where understanding context, intent, and dynamics of human communication is needed to interpret the meaning of what is being written. Sentiment analysis, emotion recognition, or opinion mining are inherently subjective. These tasks are often related to subjective language, such as sarcasm, irony, politeness, and humor, which are often context-dependent and can change the literal meaning of words.
In conversational AI, understanding the pragmatics is vital for developing systems that engage in natural and coherent dialogue. This involves interpreting and generating contextually appropriate responses, understanding indirect speech acts, and maintaining the coherence of the conversation over multiple turns. In power analysis research, understanding the roles of participants and their interactions can reveal power dynamics \cite{fairclough1989language, vandijk1997discourse}. Pragmatics enriches the analysis by interpreting cues that indicate agreement, disagreement, or varying levels of participant engagement \cite{levinson1983pragmatics}.
Pragmatic analysis often considers cultural and social norms, as they heavily influence linguistic devices such as sarcasm or politeness \cite{brown1987politeness}.

\section{Relevant NLP tasks}
\label{section:background_relevant_tasks}
NLP research is often classified by task or the specific goal for which new methods are developed. Although there are hundreds of tasks, some are particularly important for analyzing personality cues in text. Specifically, NLP methods for the semantic comparison of two statements must account for stylistic variations in textual expression, as seen in tasks such as semantic similarity and paraphrase detection.

One of the key challenges in determining semantic similarity between two natural language statements is the lexical gap -- the variation in word choice and phrasing that can obscure underlying similarity. However, this is not the only challenge; factors such as syntactic variation, contextual nuances, and world knowledge also play a crucial role in accurately measuring semantic similarity.

This task has been approached through various tasks such as paraphrase detection \cite{androutsopoulos2010survey}, semantic textual similarity \cite{agirre2012semeval}, and recognizing textual entailment
 \cite{dagan2010recognizing}. To solve these tasks successfully, language understanding capabilities covering various linguistic phenomena, including negation, synonymy, antonymy, semantic relations, lexical entailment relations, logical, and common-sense inference, are necessary  \cite{lobue2011types, cabria2014decomposing}. However, recent advances in deep representation learning  \citep{bengio2013representation, goldberg2017neural} have led to the development of highly accurate models for these tasks (e.g.,  \cite{reimers-2019-sentence-bert}).

\subsection{Semantic similarity}
\label{subsection:background_semanticsim}
Semantic similarity \cite{agirre2012semeval}, a core task in NLP, is concerned with quantifying the similarity between textual units, ranging from sentences and paragraphs to entire documents. Semantic similarity is used in many applications, including clustering similar documents, improving the efficiency of document retrieval systems, and detecting duplicate or redundant content. Central to this concept is the idea that text segments sharing similar meanings will manifest a certain level of semantic proximity, regardless of the differences in their specific word choices.

A range of methodologies are used to measure semantic similarity, each with its unique approach. Jaccard similarity, for instance, assesses the extent of overlap between the sets of words in the texts, while cosine similarity calculates the cosine of the angle between vector representations of the textual entities. These vector representations might be based on feature engineering techniques discussed in Section \ref{section:background_features} or derived from the contextual representations obtained by training LLMs. The latter offers a more nuanced understanding of semantic relationships. %In practice, for large datasets, methods for assessing semantic similarity are frequently utilized in conjunction with vector indexes or stores, especially during the retrieval phases of Retrieval-Augmented Generation (RAG) models.

\subsection{Paraphrase detection}
\label{subsection:background_paraphrasedet}
Paraphrase detection \cite{dolan2004unsupervised}, closely related to semantic similarity, aims to estimate whether two sentences convey the same meaning despite potential variances in wording and structure. Paraphrases are used in different tasks, including information retrieval, text classification, and machine translation. The primary challenge lies in accurately capturing the essence of meaning while accounting for linguistic variability.

For instance, consider the sentences ``She is highly conscientious and always follows through on her commitments'' and ``She is very responsible and consistently meets her obligations.'' These are paraphrases because they convey the same meaning using different words and structures. In contrast, ``He is outgoing and enjoys social gatherings'' versus ``He is outgoing but dislikes attending parties'' show high textual similarity due to the shared word ``outgoing'', yet they are not paraphrases because they express conflicting sentiments.

Various strategies are employed in paraphrase detection. Rule-based approaches use predefined linguistic rules to identify paraphrases \cite{barzilay2003learning}, while ML-based methods apply supervised learning to discern patterns in sentence mappings \cite{socher2011dynamic}. In contrast, DL-based methods, such as Siamese networks \cite{neculoiu2016learning} or Transformer models \cite{reimers2019sentence}, leverage neural networks to encode sentences into vectors. These vectors are then compared to determine if the sentences are paraphrased. Such approaches have gained prominence for handling complex linguistic variations more effectively.

\chapter{Personality}
\label{ch:background_personality}
Personality is characterized by distinct patterns of thoughts, emotions, and behaviors and fundamentally shapes an individual's identity \cite{matthews2009personality}. This concept encompasses individual differences, including traits, values, beliefs, and consistent behaviors. Understanding personality helps to explain an individual's motivations, actions, and social interactions, significantly impacting life choices, educational directions, career paths, and relationship dynamics.
Personality psychology is a research area that focuses on understanding personality, primarily through empirical research methods. It relies on psychometric assessment tools that are concerned with creating objective, reliable, and valid methods to quantify personality traits.

Historically, personality assessments are related to military applications, where they were used for role assignments \cite{weiner2017handbook}. Later, these practices were adopted in corporate settings for effective employee management, including job allocation, training, and development. In therapeutic contexts, personality assessments help individuals identify their strengths and areas of self-improvement.

With the onset of the digital age, personality assessment has undergone a significant transformation. The rise of social media and the advances in ML have broadened the scope of personality assessments within digital environments. These tools have become vital in various applications, including personalizing online user experiences \cite{arazy2015personalityzation}, shaping digital marketing tactics \cite{winter2021effects}, and improving dialogue systems \cite{shumanov2021making}, among many others.

%\todo{improve intro}

\section{Personality models}
\label{section:background_personality_models}
As a broad and diverse field, personality research utilizes different approaches and techniques to understand human behavior. It typically examines personality through several domains: dispositional, biological, intrapsychic, cognitive-experiential, adjustment, and social and cultural \cite{larsen2020personality}. Each domain provides a unique perspective for studying personality.

The dispositional domain concerns the intrinsic traits and characteristics that define individuals, highlighting patterns consistent across diverse situations and throughout one's life. Central to this domain is the concept of \emph{traits}, enduring patterns that capture an individual's unique way of thinking, feeling, and behaving. Although immediate circumstances might momentarily influence an individual's behavior, the underlying traits ensure a consistent behavioral pattern across varied situations and life phases. Finally, traits distinguish one person from another and provide a structured framework for understanding the complexities of human personality.

\subsection{Type- and trait-based models of personality}
\label{subsection:background_typetraitmodels}
%\todo{add graphics depicting trait and type-based personality models}

The study of personality has been approached through both type- and trait-based models \cite{matthews2009personality}, each offering a unique perspective on human behavior and temperament. The type-based approach, rooted in ancient Greek thought, categorizes individuals into distinct and broad categories or \textit{types} \cite{matthews2009personality}. Historically, Hippocrates proposed four temperaments based on bodily fluids: sanguine, choleric, melancholic, and phlegmatic. Carl Jung later introduced psychological types, emphasizing introversion and extraversion. In contrast, the trait-based approach, which gained prominence in the 20th century, posits that personality combines specific and measurable traits or dimensions. The evolution from type to trait models in psychology reflects a progression towards a more detailed and empirically grounded understanding of personality. Instead of broad categories, individuals are seen to have varying degrees of specific traits, such as openness or conscientiousness. The Five-Factor Model (FFM) \cite{mccrae1987validation} or the Big Five \cite{goldberg1993big} is a notable example of this approach, identifying five core traits that describe human personality. Over time, the trait-based approach, emphasizing empirical research and quantifiable measures, has become more dominant in modern psychology, although both models continue to inform our understanding of personality.
Although conceptually distinct, they overlap in many areas. Types can be considered clusters of traits, suggesting that individuals categorized as a particular ``type'' might exhibit a specific combination of traits in a trait-based model.

\textbf{Type-based personality models} classify individuals into a finite number of distinct categories, operating on the premise that individuals with similar personalities share a common set of traits, behaviors, and characteristics. The most well-known type-based personality model is the \textbf{Myers-Briggs Type Indicator (MBTI)} \cite{myers1998mbti}, which divides individuals into 16 unique personality types grounded in four dichotomies: extraversion versus introversion, sensing versus intuition, thinking versus feeling, and judging versus perceiving.

The MBTI is utilized in business organizations and by the general public, as it helps in analyzing individual strengths, weaknesses, and tendencies. In a business setting, the MBTI supports leadership training and employee development. Additionally, type-based models, including the MBTI, are applied in situations where individuals are eager to understand their personalities, such as in career counseling or personal development.

The easy-to-understand categorizations provided by the MBTI, which outline a set of expected characteristics for each type, have facilitated its popularity by allowing individuals to identify with these descriptions more readily. Consequently, it led to the development of numerous online communities and forums centered on MBTI personality types, where individuals frequently find a sense of belonging and understanding.

Although the simplicity of the MBTI has helped it gain considerable popularity, it has not been immune to criticism, especially from the psychological community that cites a lack of empirical support. Studies have underscored its relatively lower reliability and validity than other personality assessment tools \cite{boyle1995myers,capraro2002myers,furnham1996faking,barbuto1997critique,thyer2015}. Much of this criticism stems from the central flaw that, similarly to most naturally occurring phenomena, personality traits generally adhere to a normal distribution, meaning that most individuals possess traits clustering around average values. This central tendency potentially underpins the inaccuracies in personality assessments used by type-based models.

In contrast, \textbf{trait-based personality models} define the personality as a set of continuous dimensions, each measured on a scale. These models are widely used in various fields of psychology in addition to personality, including organizational \cite{hough2008personality} and social psychology \cite{schinka2003handbook}, as they offer a quantitative framework for understanding personality. They facilitate the measurement and comparison of individual personality differences, allowing research on the impact of personality on various outcomes, such as job satisfaction, job performance, mental health, and well-being. Trait-based personality models, such as the Big Five \cite{goldberg1993big} and the Five-Factor Model \cite{costa1992neo}, encompass five broad domains: Openness (or Intellect), Conscientiousness, Extraversion, Agreeableness, and Neuroticism (or Emotional Stability). Table~\ref{table:big5_btf_examples} outlines common patterns of behavior, thought, and emotion associated with each of these traits in the Big Five model.
The conceptual foundation of the theory of the Big Five personality traits originates from the lexical hypothesis \cite{allport1936trait}. This hypothesis posits that the most significant and socially relevant personality differences among people are encoded in language. Essentially, the most critical traits for distinguishing between individuals become integrated into language over time because of their role in interpersonal relations and communication. Moreover, the hypothesis suggests that the importance of a trait is reflected in the prevalence and simplicity of the terms used to describe it. Consequently, personality researchers can identify the most relevant traits by analyzing language, particularly adjectives used to describe individuals.

However, while the lexical hypothesis is conceptually straightforward, applying it to develop personality models proved to be a complex challenge. The primary difficulty was determining how to analyze latent constructs such as personality traits based on lexical cues. Addressing this required the development of statistical tools, including factor analysis, which was instrumental in early personality research \cite{spearman1904general}.

A major line of research that emerged from the lexical hypothesis is the lexical approach to personality modeling. This approach involves compiling lists of descriptive words -- such as adjectives -- that characterize individuals’ traits. These lists, often developed in multiple languages, are presented to individuals who select the words that best describe themselves. Researchers then analyze the co-occurrence patterns of these self-descriptive terms, typically employing dimensionality-reduction techniques like factor analysis. Over decades of research, this method has consistently identified five broad personality domains, collectively known as the Big Five: openness to experience, conscientiousness, extraversion, agreeableness, and neuroticism. While extraversion and agreeableness appear to be relatively universal across cultures and languages, traits such as openness to experience exhibit greater cultural variability \cite{mccrae2002five}.

Substantial evidence indicates that the typical correlations of the Big Five domains with other constructs are relatively low. This suggests a need to consider more specific personality subtraits, such as aspects \cite{deyoung2007aspects}, facets \cite{costa1995facets}, or nuances \cite{mottus2016nuances, mottus2017nuances}, depending on the underlying personality models. Each of these sub-traits provides a more detailed description of the associated traits. For instance, the aspect scale divides each domain into two aspects, the NEO scale \cite{costa1992neo} further dissects each domain into six facets, and nuance refers to all the individual items associated with a particular domain or the smallest units used in measuring a personality model. Such increased granularity affords more precise descriptions of related patterns of thinking, behaving, or feeling.

Today, most research uses facet-level models to examine specific traits within each domain, such as \textit{excitement-seeking} or \textit{cheerfulness} within the \textit{Extraversion} domain or \textit{achievement striving} and \textit{dutifulness} in \textit{Conscientiousness}. This becomes important when recognizing that two individuals can have identical domain trait scores yet display significant differences in individual facets. For instance, a person with average scores across all facets of extraversion will present a markedly different profile from someone whose average score in extraversion stems from an extremely high \textit{excitement-seeking} score balanced by an extremely low score in \textit{cheerfulness}. Table~\ref{table:big5_facets} shows the facets of the NEO personality inventory.

\begin{table}[h]
\centering
{\small
\caption{Characteristics of the Big Five personality traits \cite{john:1999, matthews2009personality}}
\label{table:big5_btf_examples}
\begin{tabular}{l|l|l}
\hline
\textbf{Trait} & \textbf{Aspect} & \textbf{Description} \\
\hline
Openness & Behavior & Often curious, imaginative, and seek new experiences. \\
\hline
& Thinking& Tend to be more open-minded and flexible in their thinking. \\
\hline
& Feeling & May experience a wide range of emotions. \\
\hline
Conscientiousness &  Behavior & Organized, responsible, and reliable. \\
\hline
 & Thinking & Often focus on details and contemplate action consequences. \\
\hline
& Feeling & May feel guilty or ashamed when failing to meet their own standards. \\
\hline
Extraversion & Behavior & Outgoing, sociable, and enjoy being around others. \\
\hline
 & Thinking & Tending to think out loud and enjoy bouncing ideas off of others. \\
\hline
& Feeling & May experience strong emotions, especially in social situations. \\
\hline
Agreeableness & Behavior & Cooperative, friendly, and considerate of others. \\
\hline
 & Thinking & Value harmony and relationships over competition or gain. \\
\hline
& Feeling & May experience feelings of empathy and compassion towards others. \\
\hline
Neuroticism & Behavior & May experience frequent or intense emotional distress. \\
\hline
 & Thinking & Tending to focus on negative aspects of a situation. \\
\hline
& Feeling & May experience a wide range of negative emotions. \\
\hline
\end{tabular}
}
\end{table}

\begin{table}[h]
\centering
\small
\caption{Domains and facets of the NEO personality inventory \cite{costa1992neo}}
\begin{tabularx}{\textwidth}{l|X}
\hline
\textbf{Domain} & \textbf{Facets} \\
\hline
Neuroticism & Anxiety, Angry Hostility, Depression, Self-Consciousness, Impulsiveness, Vulnerability \\
\hline
Extraversion & Warmth, Gregariousness, Assertiveness, Activity, Excitement-Seeking, Positive Emotions \\
\hline
Openness to Experience & Fantasy, Aesthetics, Feelings, Actions, Ideas, Values \\
\hline
Agreeableness & Trust, Straightforwardness, Altruism, Compliance, Modesty, Tender-Mindedness \\
\hline
Conscientiousness & Competence, Order, Dutifulness, Achievement Striving, Self-Discipline, Deliberation \\
\hline
\end{tabularx}
\label{table:big5_facets}
\end{table}

Not all trait-based models have five traits. The HEXACO model \cite{lee2004psychometric} of personality structure is a framework in psychology that extends the traditional Big Five personality traits by adding a sixth factor, Honesty--Humility. This model was developed by Kibeom Lee and Michael C. Ashton in the early 2000s, based on findings from several studies using different languages and cultures. The Honesty--Humility dimension represents traits related to fairness, sincerity, modesty, and greed avoidance. Individuals high in this trait are often viewed as sincere, fair, and modest, whereas those who are low might be more prone to manipulation, deceit, and arrogance.

\subsection{Correlation between personality models}
\label{subsection:background_personality_models_correlation}
The landscape of personality psychology is rich with various models, encompassing type-based systems such as MBTI and trait-based frameworks like the Big Five and HEXACO. Given that these models aim to map the underlying latent structure of personality, it is not surprising that they often exhibit correlations with each other. As they are conceptually aligned to some degree, models like the Big Five, the Five-Factor Model, and HEXACO show a higher degree of inter-correlation \cite{doi:10.1177/08902070211026793}.

Interestingly, even the MBTI, which structurally differs from the trait-based models, demonstrates significant correlations with the five Big Five dimensions. For instance, the Introversion/Extraversion dimension in MBTI directly correlates with Extraversion in the Big Five, while the Thinking/Feeling dimension aligns with the Big Five's Agreeableness. Similarly, Judging/Perceiving correlates with the Conscientiousness trait of the Big Five, and Intuitive/Sensing corresponds with Openness to Experience \cite{furnham1996big}.
However, a notable discrepancy exists in that the Neuroticism trait from the Big Five model lacks a direct counterpart in MBTI.

\section{Personality and demographics}
\label{section:background_personality_demographics}
\subsection{Personality and culture}
The complex relationship between personality and culture has many dimensions. The pancultural approach suggests that certain personality traits, such as those in the Five Factor Model, are universal across cultures \cite{mccrae2005universal}. This position is supported by multinational studies that consistently identify these traits in diverse cultural landscapes using various methods. Moreover, patterns have emerged, such as individualist countries exhibiting higher scores on extraversion and openness.

Cross-cultural psychologists have raised concerns, arguing that various factors can influence these countries' differences in personality traits \cite{heine2009personality}.
Cultural norms and values play a role in molding personality development. For instance, in collectivist societies, where community values and harmony are paramount, there is a pronounced emphasis on traits like agreeableness \cite{triandis2001individualism}. These traits include cooperation, empathy, and modesty. Individuals nurtured in such environments tend to prioritize group goals over individual aspirations, cultivating a personality that is attuned to communal needs and expectations.
Conversely, individualistic cultures often emphasize traits linked to extraversion, such as assertiveness and excitement seeking. These societies celebrate individual achievements and autonomy, encouraging personalities to prioritize self-expression and individual fulfillment.
Another finding is that bilingual individuals might show different personality traits based on their native language, indicating that cultural context can shape personality. Furthermore, while individualist countries might have a high concentration of extroverts open to experiences, there exists a distinction between genuine personality traits and cultural stereotypes. The dynamic relationship between personality and culture is one of mutual influence, each shaping the other in complex ways.
Finally, as with other demographic variables, there are many variations and nuances within a single cultural group. In addition to culture, personality traits, to some extent, depend on other sociodemographic variables, such as age and gender.

\subsection{Personality and gender}
In terms of gender differences, research \cite{costa2001gender, feingold1994gender} has suggested that, on average, males exhibit a propensity towards higher scores in traits related to assertiveness and dominance -- a characteristic discerned distinctly within the traits of extraversion and assertiveness outlined in the Big Five. In contrast, females frequently score higher on facets associated with nurturance and empathy, delineated predominantly in the agreeableness and warmth traits of the Big Five personality framework. Females also score higher than males on anxiety and trust.  This tendency identified in empirical studies can manifest itself in various spheres of life, including vocational preferences and leadership dynamics. This suggests that historical, cultural, and biological factors influence these differences in society.

However, research shows that gender differences in personality traits are not absolute, as the variability in personality traits within individual gender groups significantly overshadows the differences observed between genders \cite{carothers2013categorical, hyde2005gender}. 

\subsection{Personality and age}
While influential in predicting significant life outcomes, personality traits are not static throughout an individual's life. Historically, the emphasis has been on the stability of these traits, suggesting that they consistently predict life outcomes. However, recent research \cite{bleidorn2022personality} indicates that while personality traits exhibit stability, they never achieve complete stability. This suggests that personality traits are malleable and can evolve. During young adulthood, there is a notable shift toward traits that signify psychological maturity, such as emotional stability, conscientiousness, and agreeableness -- a phenomenon known as the maturity principle of personality development \cite{caspi2005personality}. Conversely, adolescents often witness a temporary decline in traits such as conscientiousness, extraversion, and agreeableness; a trend termed the \textit{disruption hypothesis} \cite{doi:10.1177/0963721415589345}. While young adulthood and adolescence have been extensively studied, there is not much research on the evolution of personality traits in middle-aged and older adults. However, some evidence suggests that the maturation trends observed in earlier life stages may reverse in late adulthood.
\label{section:background_personality_assessment}
\section{Personality assessment}
Most research on personality relies on the valid measurement of personality traits. Psychometrics aims to quantify various traits and human abilities, including aspects such as intelligence, aptitude, personality, and emotional and behavioral characteristics. Psychometric tests are standardized assessments designed to measure specific constructs, providing a numerical score or rating representing the individual's standing in that construct.

Personality questionnaires or inventories are tools designed to assess individual personality traits and are widely used in psychology, human resources, and personal development. 
The items are the individual questions or statements in a questionnaire. Each item is designed to elicit responses that reflect certain aspects of the respondent's personality. Items typically use a Likert scale (e.g., ``strongly agree'' to ``strongly disagree'') or other rating systems. The responses are then scored to reflect the presence or intensity of specific personality traits. Good items are clear, concise, and designed to minimize misinterpretation or bias.

Proprietary inventories are developed and owned by an individual or organization, often requiring payment or a license. An example of a proprietary inventory is the MBTI.
In contrast, open-source inventories are freely available to the public. They can be used, modified, and distributed without cost. The most notable examples of such inventories are the Big Five Inventory (BFI) \cite{john1991big} and International Personality Item Pool (IPIP), which provides a collection of items for various personality scales, including those aligned with the Big Five traits. Because of their reliability and validity, two commonly used self-report questionnaires are the BFI and the NEO Personality Inventory-Revised (NEO-PI-R) \cite{costa1992neo}. The BFI consists of 44 items related to each of the five domains and their facets, and individuals are asked to rate their agreement with each item on a five-point scale (e.g., ``strongly disagree'', ``disagree'', ``neutral'', ``agree'', ``strongly agree''). The responses to the items are then used to calculate scores for each of the five domains and the facets within each domain. The NEO-PI-R inventory comes in shorter versions of 60 items and longer versions of 240 items, each of which measures one of the facets of the Big Five Personality Traits: Openness, Conscientiousness, Extraversion, Agreeableness, and Neuroticism.
The IPIP NEO is a highly similar questionnaire that comes in versions of 120 and 300 items. The IPIP is a public domain inventory that is freely available for researchers and practitioners to use, and it has been translated into many different languages. The longer versions of these questionnaires provide a more in-depth understanding of an individual's personality and can help provide a more nuanced understanding of an individual's personality.

To be considered valid, psychometric tests must have reliable and accurate results, which means that they should produce consistent results when administered to the same person on different occasions or to different people. Additionally, psychometric tests should have content validity, which means that they cover all essential aspects of the construct they are intended to measure, and criterion-related validity, meaning that they predict real-world outcomes.

%Language responses contain valuable information that large language models can extract into scores converging with validated rating scales (O. Kjell et al., 2022). However, rating scales themselves are only an observable “proxy” and not a perfect true score (Figure 1). Most psychometric theories, such as classical test theory (e.g., Novick, 1966) or item-response theory (Reise & Waller, 2009), view self-report responses as an approximation of the true latent variable that is sought. Therefore, the validity of language-based assessments and rating scales should go beyond evaluating their convergence.
There are different ways to assess someone's personality; they can be asked to rate themselves or rated by someone else. The most commonly used type of ratings are \textbf{self-ratings} of personality, which refer to the assessments that an individual makes of their personality traits, providing valuable insights into their self-perceptions. 
Some challenges come with self-report questionnaires, including response biases, such as social desirability bias \cite{fisher1993social, king2000role}, where individuals may feel pressure to respond in a socially desirable way. This can lead to a potential underestimation of traits seen as less desirable, such as neuroticism, or an overestimation of traits seen as more desirable, such as agreeableness. 
An additional issue in self-reported personality assessments is the reference effect \cite{heine2010reference}. This problem arises because individuals interpret and respond to questions based on their unique experiences, perspectives, and internal standards. When a questionnaire asks about traits or behaviors, such as ``Are you a sociable person?'' individuals may have varying ideas of what ``sociable'' means. One person's interpretation of sociability might be attending a social event once a month, while another might consider daily social interactions as the standard.

This variability in reference points can lead to challenges in interpreting and comparing results among different people. For example, two individuals might rate themselves high in sociability, but their behavior might differ. This issue is particularly salient when assessments are used for comparative purposes, such as in organizational settings or psychological research.

Researchers may employ various strategies to address the reference problem. An approach is to use behaviorally anchored rating scales, which provide concrete behavioral examples for each trait level, helping to standardize responses. Another approach is to complement self-reports with other types of assessment, such as observer ratings or behavioral observations, to mitigate these biases and validate self-report scores.

\textbf{Other ratings} of personality refer to assessments of an individual's personality traits, such as friends, family members, or coworkers, made by others who know the individual well. In the latter case, other ratings are often named peer ratings. These ratings can be obtained using questionnaires, where the individual being rated is asked to nominate a few people who know them well, and these individuals are then asked to rate the individual's personality traits. Other ratings can provide valuable complementary information to self-ratings, as they can provide a different perspective on an individual's personality based on the observations and experiences of others.

\textbf{Zero-acquaintance ratings} pertain to evaluations of personality traits conducted by individuals without prior interaction with the person being assessed. These ratings can be acquired through various methods, including observing an individual's behavior during a laboratory task or assessing their social media profile. Such ratings offer valuable insights into a person's personality consistency across various situations and contexts, as they stem from minimal exposure to the individual and remain unaffected by prior interactions or relationships.

\subsection{Reliability and validity}
\label{subsection:background_validity_reliability}
In psychometrics \cite{nunnally1994psychometric}, validity is a concept that denotes the degree to which a test precisely measures what it is designed to measure. Ensuring the validity of the test confirms that the test accurately reflects the specific construct that it aims to assess. However, the precondition for testing validity is reliability. Reliability refers to the test's consistency over time and across different samples. This aspect of psychometrics tells us whether a test can consistently produce stable and repeatable results under varying conditions and at different times.
Table~\ref{table:validity_types} summarizes different aspects of validity.

\subsection*{Types of reliability}
\textbf{Test-retest reliability} refers to the consistency of test scores when the test is administered on two or more different occasions to the same individuals.
\textbf{Interrater reliability} refers to the consistency of test scores when the test is administered by two or more different raters. This is important to ensure that the results are not biased by the person administering the test.
\textbf{Internal consistency reliability} refers to the consistency of test scores within a single test. This involves evaluating the consistency of scores on different test items or subtests to ensure that the results are consistent and free from random measurement errors.

\subsection*{Types of validity}
Similarly to reliability, there are different types of validity.
\textbf{Construct validity} focuses on the degree to which a test measures the theoretical construct it is intended to assess. This involves evaluating the relationship between a test and other measures of the same construct and ensuring that test results align with theoretical and research-based expectations. Closely related to construct validity are \textbf{convergent validity} and \textbf{discriminant validity}. 
\textbf{Convergent validity} refers to the extent to which a test correlates with other measures of the same construct taken at different times. This ensures that the test produces consistent results across different assessments of the same concept.
\textbf{Discriminant validity} concerns the degree to which a test does \textit{not} measure constructs or traits it is not intended to assess. This ensures that results are not contaminated by unrelated factors.
\textbf{Content validity} refers to the extent to which a test comprehensively covers all important aspects of the construct it aims to measure. This requires evaluating test items to ensure they adequately represent the construct and are free from bias.
\textbf{Criterion-related validity} addresses how well a test predicts \textit{real-world} outcomes, such as future academic or job performance. This involves examining the relationship between test scores and relevant criteria, such as grades or workplace performance.
\textbf{Face validity} concerns the degree to which a test appears to measure what it is intended to measure based on its content and format. This involves evaluating the test from the perspective of test takers to ensure that the items are clear and relevant to the measured construct.
\textbf{Predictive validity} focuses on the extent to which a test accurately forecasts future behavior or performance. This requires analyzing the relationship between test scores and subsequent outcomes, such as academic achievements or job performance. 

\begin{table}[h!]
\centering
\small
\caption{Aspects of reliability and validity with their descriptions \cite{nunnally1994psychometric, rust2014modern}}
\label{table:validity_types}
\begin{tabularx}{\textwidth}{lX}
\toprule
\textbf{Aspect} & \textbf{Description} \\
\midrule
\multicolumn{2}{c}{\textbf{Reliability}} \\
\midrule
Test–retest reliability & Refers to the consistency of test scores when the test is administered on two or more different occasions to the same individuals. \\
\addlinespace
Interrater reliability & Refers to the consistency of test scores when the test is administered by two or more different raters. \\
\addlinespace
Internal consistency reliability & Refers to the consistency of test scores within a single test, evaluating the consistency of scores on different items or subtests. \\
\midrule
\multicolumn{2}{c}{\textbf{Validity}} \\
\midrule
Face validity & Concerns the degree to which a test appears to measure what it is intended to measure based on its content and format. \\
\addlinespace
Content validity & Refers to the extent to which a test comprehensively covers all important aspects of the construct it aims to measure. \\
\addlinespace
Construct validity & Focuses on the degree to which a test measures the theoretical construct it is intended to assess, by evaluating its relationships with other measures and theoretical expectations. \\
\addlinespace
Convergent validity & Refers to the extent to which a test correlates with other measures of the same construct taken at different times. \\
\addlinespace
Discriminant validity & Concerns the degree to which a test does \textit{not} measure constructs or traits it is not intended to assess. \\
\addlinespace
Criterion-related validity & Addresses how well a test predicts real-world outcomes, such as future academic or job performance. \\
\addlinespace
Predictive validity & Focuses on the extent to which a test accurately forecasts future behavior or performance. \\
\bottomrule
\end{tabularx}
\end{table}

\subsection{Frameworks for assessing personality judgments}
\label{section:background_personality_jugdment}
Other- and, to a greater extent, zero-acquaintance ratings rely on the ability of judges to infer someone's personality.
Brunswik Lens Model \cite{brunswik2023perception} and the Realistic Accuracy Model (RAM) \cite{letzringinpress} are two theoretical approaches used to understand human judgment and the process by which people form impressions and make predictions about others. 

\textbf{Brunswik Lens Model} offers a way to understand how people use incomplete or uncertain information in their environment to make judgments. Brunswik used the analogy of a lens to illustrate how information gets filtered through our cognitive processes before it affects our decisions. The model also considers how different environmental cues have varying degrees of validity (how well they predict the criterion) and reliability (how consistently they provide the same information). The main idea is that humans rely on various environmental cues to make probabilistic judgments.

The Lens Model consists of several interconnected elements as depicted in Figure~\ref{fig:vinci}. Environmental variables refer to underlying states or attributes in the environment that may not be directly observable. These variables are related to proximal cues, which are observable and correlated with distal cues, acting as signals or clues.
Perception and judgment involve recognizing and interpreting cues to form opinions or decisions. Functional achievement reflects how judgment correlates with actual environmental variables, indicating the accuracy of those judgments.
The model name is derived from its visual representation, consisting of two lenses. The left side illustrates the ecological correlation between environmental variables and cues, and the right side symbolizes the psychological correlation between cues and judgment. These lenses metaphorically demonstrate how cues act as a filter through which the world is perceived, with potential room for error and bias in judgment.
Brunswik's lens model has been instrumental across diverse fields. In clinical psychology, it clarifies the diagnostic process, while in decision-making studies, it helps to understand decisions under uncertainty. The model has also led to the development of quantitative methodologies, such as the Lens Model equation, offering researchers a precise way to model human judgment processes.

\textbf{The Realistic accuracy model (RAM)} was developed by Funder \cite{letzringinpress, funder1995accuracy} to enable a deeper understanding of the accuracy of judgments of personality traits. This model outlines that accurate judgments require the completion of four stages: relevance, availability, detection, and utilization. 

In the first stage, \textbf{relevance}, the target individual must display behavior that accurately reflects the trait being assessed. This behavior should be both observable and meaningfully connected to the judgment being formed. Additionally, the cues must be directly linked to the evaluation process, aligning with the concept of ecologically valid distal cues within the Lens Model framework. Ecologically valid cues refer to observable behavioral indicators that reliably correspond to an underlying trait in a given environment. These cues should not only be available but also interpretable in a way that accurately conveys trait-relevant information. Within the Lens Model, ecologically valid cues are essential because they serve as the medium through which perceivers infer personality traits. If a cue lacks ecological validity, it can lead to inaccurate or misleading judgments, reducing the accuracy of personality assessment.

Subsequently, the \textbf{availability} stage asserts that the judge should have access to the necessary cues, either through direct observation or other means. If the information remains hidden or unclear, it becomes unavailable for judgment.
The third stage, \textbf{detection}, involves the judge's ability to discern or notice relevant and available cues. This phase requires perceptual and cognitive skills, along with a focus on the pertinent cues to the trait under scrutiny. In the Lens model, this stage is equivalent to perception modeling, where proximal cues are utilized to identify distal cues.
In the final stage, \textbf{utilization}, the judge is expected to interpret the observed data correctly, correlate it accurately with the evaluated trait and form a coherent judgment.

The sequential nature of RAM stages indicates that any compromise can potentially lead to a less precise judgment. For instance, a failure to detect relevant information or improper utilization of it could result in skewed or incorrect judgments. Furthermore, the sequential arrangement of the stages can help diagnose the judgment process and allow us to pinpoint the most likely cause of the judgment inaccuracy.

The Lens Model and RAM underscore the importance of environmental cues and cognitive processes in judgment and decision-making, albeit with different emphases. While the Lens Model depicts a probabilistic relationship between cues and judgments, RAM outlines a sequential four-stage process. In Chapter~\ref{ch:simpa}, we will use RAM as the backbone of our proposed computational framework.

\chapter{Text-based personality analysis}
\label{ch:background_text_based_pa}

In Chapter~\ref{ch:background_nlp} and Chapter~\ref{ch:background_personality}, we discussed how the lexical hypothesis underpins modern personality models, emphasizing the central role of language in developing personality theory. However, the utility of language extends beyond theoretical frameworks; it also serves as a practical source of cues for individual personality assessment.

Practical objectives have predominantly driven the development of tools and methodologies for text-based personality analysis. An early application of text analysis is authorship identification, which examines linguistic patterns to determine a text’s probable author. A notable example is the Federalist Papers, 85 essays written under the pseudonym Publius to support the U.S. Constitution's ratification. In 1963, statistical methods were used to attribute authorship to Alexander Hamilton, James Madison, and John Jay \cite{mosteller1963inference}.

\section{Research landscape}
The field concerned with such questions is \textbf{author profiling}, a subfield of stylometry or the study of linguistic style. Stylometry has been used for centuries in various forms, going back to ancient Greek scholars who attempted to verify the authorship of classic works \cite{schironi2018best}.
As technology, and especially ML techniques, evolved, the scope of author profiling expanded. In addition to authorship attribution, it often involves determining other characteristics of the author, such as age, gender, native language, and personality traits from different text sources. This has applications in various fields, from forensics and security to marketing and literary analysis.

\begin{figure}[t]
    \centering
    %\vspace*{-1.5em}
    \hspace*{-0.5em}
    \includegraphics[scale=0.65]{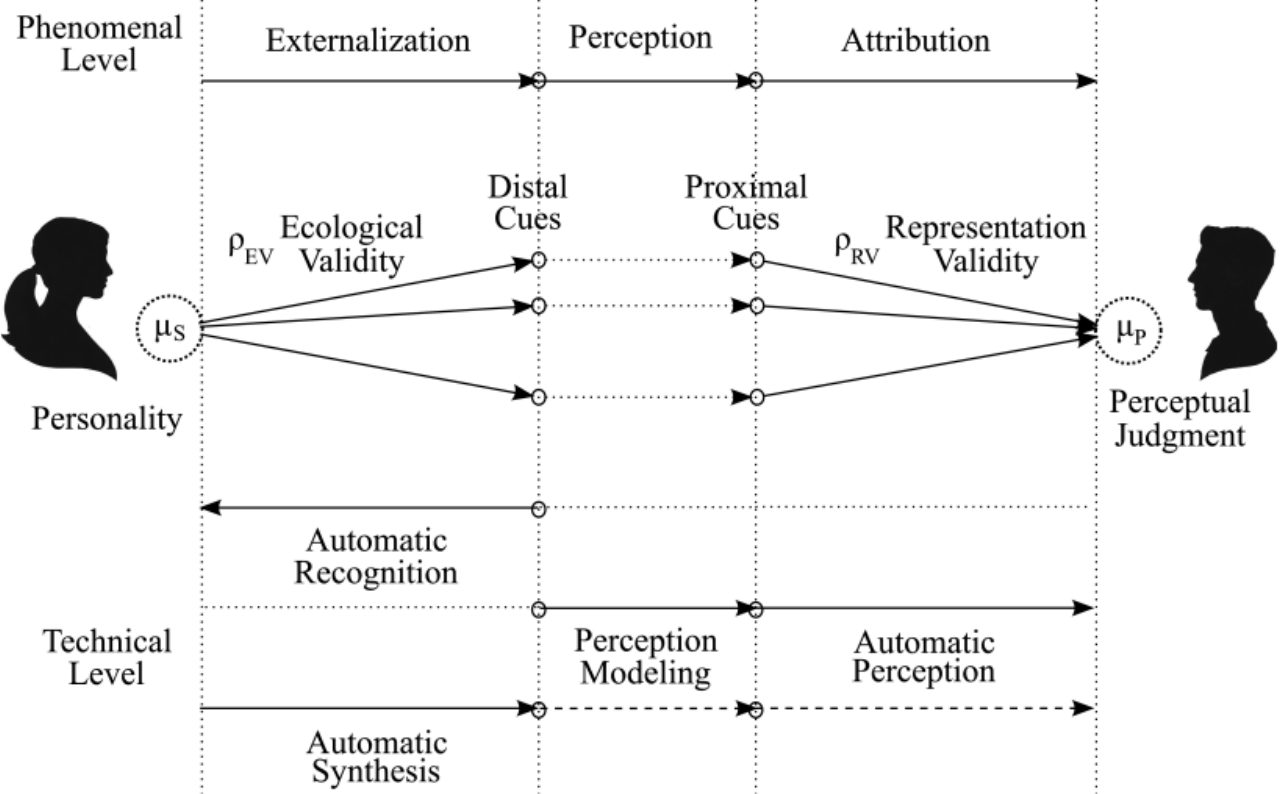}
    
    \caption{This figure illustrates the relationship between the Lens Model and the three main problems addressed in Personality Computing  \cite{vinciarelli2014survey}. ``Automatic Personality Recognition'' refers to the inference of self-assessments ($\mu_S$ in the figure) from distal cues. ``Automatic Personality Perception'' concerns the inference of assessments ($\mu_P$ in the figure) from proximal cues. ``Automatic Personality Synthesis'' involves generating artificial cues designed to elicit the attribution of predefined traits. Adapted from  \cite{vinciarelli2014survey}.}
    \label{fig:vinci}
\end{figure}

While author profiling uses personality as just one of many variables of individual differences, the field directly concerned with personality analysis is termed \textbf{Personality Computing} \cite{vinciarelli2014survey}.
Vinciareli and Mohammadi provided a survey on Personality Computing describing approaches that focus on three fundamental tasks: \textbf{Automatic Personality Recognition} (APR), \textbf{Automatic Personality Perception} (APP), and \textbf{Automatic Personality Synthesis} (APS). These tasks can be described by examining various aspects of the Lens Model (see Section \ref{section:background_personality_jugdment}) depicted in Figure~\ref{fig:vinci}. Originally introduced to explain how living beings gather information from their environment, the Lens Model was later adopted to describe the externalization and attribution of socially relevant characteristics during human-human \cite{scherer1978personality} interactions and, more recently, human-machine \cite{pianesi2012searching} interactions.

%\paragraph{Type-level prediction.} 

According to the Lens Model, individuals express their personality, a latent structure not directly observable through distal cues that others can perceive.

\textbf{Automatic Personality Recognition} (APR) involves inferring self-reported personalities from machine-detectable distal cues. The underlying assumption is that self-reports represent the actual traits of an individual and that computational models are used to recognize these traits. Typically, these models examine covariation measures between personality traits and distal cues, focusing on ecological validity, a measure of how test performance predicts behaviors in real-world settings. As with related fields like author profiling, APR incorporates various methodologies outside author profiling, from affective computing and social signal processing to other disciplines, including those outside computer science, such as sociolinguistics.

Certain signals are not visible outright; detecting them relies on one's skill in observing readily noticeable cues. For example, a gardener might gauge the health of a plant by examining its leaves (immediate signals), which can suggest underlying issues with roots or soil moisture (indirect signals). Within the realm of understanding human traits, this mechanism is known as personality perception.

\textbf{Automatic Personality Perception} (APP) involves inferring the personality that observers attribute to a specific target based on proximal cues. Unlike APR, which relies on self-reports, APP is based on other- and zero-acquaintance reports, meaning it depends on other people's perceptions of a target's personality traits. During the modeling of the perception process, the same methods are mostly employed. However, the emphasis shifts from ecological validity to the representation validity of the cues, focusing on the covariation between proximal cues and personality traits.
APP approaches typically aim to predict the traits attributed by multiple judges. However, individual differences between judges that influence their personality perception are often not modeled. If there are multiple judges, the target's personality score is usually calculated based on the mean scores.

Recently, with the increased use of dialogue systems, there has been an increased effort to make these systems exhibit human-like qualities. The objective is typically to instill desirable personality traits, such as good humor or openness to experience.
The \textbf{Automatic Personality Synthesis} (APS) task automatically generates distal cues based on the desired personality traits in personality computing. Developing APS models typically involves human judges to align human perception with machine-generated cues.

Before the rise of LLM-based dialogue agents, text personality analysis primarily focused on automatic personality recognition. However, this focus was driven by the goal of developing new assessment methods to complement traditional questionnaire-based approaches -- a goal that remains relevant today.

\section{The role of automated methods}
The promise of automated text-based personality assessment (ATBPA) lies in its ability to efficiently analyze large volumes of text data to identify explicit descriptions of behavior, thoughts, or emotions, while also recognizing implicit and nuanced personality traits with a level of consistency that often surpasses human performance. Implicit signals, such as language style, are typically subtle and challenging for humans to detect. By maintaining this consistency, ATBPA overcomes some of the limitations associated with traditional assessment methods, including respondent fatigue -- which can reduce the reliability of responses by limiting the number of items -- and the tendency of respondents to provide socially desirable answers.

Previous work has considered a wide range of relevant cues or features for personality prediction, which may broadly be categorized as features of content, style, or a combination of both (see Section \ref{section:background_features}). The most commonly used content features are words, phrases, word categories, and topics from a prespecified list (closed-vocabulary approaches) \cite{pennebaker2015development} or extracted from the text itself (open-vocabulary approaches) \cite{schwartz2013personality}. In contrast, stylistic features capture the linguistic style of the text and typically include subword-level features (e.g., character n-grams), punctuation and special symbols (e.g., emojis, exclamation marks), and discourse-level indicators (e.g., readability indices, cohesion metrics, type-token ratios). The more recent ATBPA methods rely on deep learning (DL) representations  \cite{mehta2020recent}, often in combination with the linguistic features mentioned above, as they have shown superior performance on various NLP tasks. The architectures used to model personality prediction tasks parallel the technological development in NLP. They include classical ML methods such as SVM and decision trees, DL architectures such as LSTM, CNN, and GNNs, and transformer-based models such as BERT  \cite{kazameini2020personality, mehta2020bottom}. 
Supervised approaches described in Section \ref{section:background_pred_models} learn to identify linguistic correlates of personality in text based on personality-labeled data. However, the apparent limitation of such models is that they require labeled data. 

\section{Data sources}

The availability of appropriate data arguably represents the primary bottleneck for many research directions in this field. While the field reaps on the development in ML and NLP communities, and it does not have to invest much effort in developing new models, these models are as helpful as the data which limits have they can be used to answer research questions related to personality.

The precondition for developing effective models is the availability of appropriate datasets. From an ML perspective, two primary components make the dataset appropriate: informative text for the task at hand and trustworthy labels. In the context of the RAM framework, the datasets are closely related to the availability stage, which precedes the detection and utilization stages, meaning that effective modeling depends on the availability of personality cues. Moreover, not all datasets are suitable for answering all personality research questions. Characteristics of the specific data source affect the volume and type of personality cues accessible \cite{mcfarland2015social}. These characteristics include factors such as anonymity, the nature of the content, its synchronicity, and the intended use of the source.

The degree of anonymity, for example, significantly impacts user discourse. Non-anonymous environments inhibit discussions on sensitive topics, leading users to alter the nature of their conversations. Personal issues on professional networks such as LinkedIn are rarely broached; when they are, they are usually for a specific purpose. At the same time, visually focused platforms like Instagram or TikTok predominantly feature concise textual content secondary to visual elements. Conversely, real-time communication platforms such as WhatsApp encourage concise messaging dependent on shared context. 
In contrast, asynchronous platforms include forums or text-focused social media platforms, such as Reddit, which often provide extensive and detailed content. However, specific platforms limit content length, potentially decreasing the level of expressiveness of communications.

A critical consideration across different platforms is the specific audience they attract, which relates to selection bias in the data collected from these platforms. For instance, Facebook's user base skews toward individuals over 30, while TikTok and Snapchat are more popular among younger demographics \cite{gottfried2024americans}. This demographic specificity often leads to homogeneous samples, presenting challenges in achieving a wider effect size in the study. Another form of selection bias is based on users' locations, as some platforms primarily attract users from specific countries, while others are more uniform globally. 

When acquiring datasets, researchers face constraints on the information that can be retrieved from particular sources. The company Meta, for example, prohibits the extraction of personal Facebook statuses or messages from private groups. The restrictions on accessing user data are for privacy, legal, ethical, and commercial reasons. Consequently, the most accessible and used datasets are those created before the implementation of such restrictions. Prominent examples include Twitter datasets such as TwiSty \cite{verhoeven2016twisty} and the Kaggle Personality Cafe dataset.\footnote{\href{https://www.kaggle.com/datasets/datasnaek/mbti-type}{https://www.kaggle.com/datasets/datasnaek/mbti-type}}

\subsection{Availability and use of personality labels}
Although the above-mentioned characteristics primarily influence the quantity and quality of the text collected from such sources, they also influence the possible labels and their trustworthiness.
Table~\ref{table:datasets} show that most datasets are labeled using the MBTI framework, considerably reducing their utility for psychologists due to well-documented concerns about the validity of the MBTI (see Section~\ref{subsection:background_typetraitmodels}). This labeling approach is less precise than the Big Five model, where trait scores typically originate from validated questionnaires and can model the normal distribution of personality traits. In many cases, MBTI labels are derived from authors' self-identification in their texts, often without having taken an official MBTI test. Instead, authors might rely on free online questionnaires or descriptions of different MBTI types.

Only a few datasets are labeled with Big Five scores from standard inventories. The myPersonality dataset \cite{kosinski2013private}, the only large-scale dataset derived from Facebook, was a significant research resource until 2018, when it became unavailable to the research community. A publicly available sample from this dataset is the CLEF workshop dataset \cite{celli2013workshop}, which includes only 10,000 texts from 250 users. Another available dataset for the Big Five is the student essay dataset by King and Pennebaker \cite{pennebaker1999linguistic}. However, the homogeneity of its sample, which includes individuals with similar ages and educational backgrounds, restricts the variety of research questions that researchers can investigate.

The lack of large-scale datasets, particularly regarding the number of texts per author, presents many unresolved challenges. One significant challenge is extracting relevant personality cues from a vast amount of text, particularly when navigating hierarchical structures in social media that share context, such as various subforums, threads, posts, comments, and replies. Traditional approaches, which utilize discrete features from the aggregated text of each author while overlooking much contextual information, handle these complex structures more effectively than deep learning methods. Nevertheless, they depend on features informed by expert knowledge. 
%\todo{add something about various features used in previous works, add how they modeled the hierarchical structure}

\begin{table}[h]
\small
\centering
\caption{Overview of widely used personality datasets. Datasets with their sources in bold indicate that access is available.}
\begin{tabularx}{\textwidth}{X|X|X|X|X|X}
\hline
\textbf{Source} & \textbf{Authors} & \textbf{Model} & \textbf{Languages} & \textbf{\# Users} & \textbf{\# Texts} \\ \hline
Twitter & Plank and Hovy, 2015 \cite{plank2015personality} & MBTI  & EN & 1500 & 1.2 million \\ \hline
TwiSty & 
%Twitter & 
Verhoeven et al., 2016 \cite{verhoeven2016twisty} & MBTI & GER, ITA, NED, FRA, PT, SPA & 18,168 & 34 million \\ \hline
%Kaggle & 
\textbf{Personality Cafe} & Kaggle, 2017 & MBTI & EN & 8600 & 430,000 \\ \hline
%MyPersonality & 
Facebook & Kosinski et al., 2013 \cite{kosinski2013private} & Big Five & EN & 74,941 & 15.4 million \\ \hline
\textbf{Facebook} & Celli et al., 2013 \cite{celli2013workshop} & Big Five & EN & 250 & 10,000 \\ \hline
%Essays & 
\textbf{Essays} & King and Pennebaker, 1999 \cite{pennebaker1999linguistic} & Big Five & EN & 2,467 & 2,467 \\ \hline
\end{tabularx}
\label{table:datasets}
\end{table}

\subsection{The role of demographics in text-based personality assessment}

Only a limited portion of the datasets incorporate essential demographic variables such as age and gender. These are vital confounders in personality and sociolinguistic research and influence different parts of text-based analysis, which can significantly impact the analysis.
On the level of language, which serves as a source of cues, authors with the same personality but from different cultural backgrounds, age groups, or genders can generate distinctly different text output. On the level of personality, there is a known change in personality scores in terms of age, gender, and culture as described in Section \ref{section:background_personality_demographics}.
The analysis becomes more complex when the individual domain traits of the authors are evaluated. Although these traits are largely orthogonal — possessing distinct discriminative properties — they can nevertheless lead to markedly varied impacts on the generated text's quality and style. For example, let us imagine similar personality profiles for two persons in which one person differs in only one domain: openness to experience. Person A is low in openness and could be more inclined to be religious, and person B is high on openness and could be more inclined to be spiritual, and based on that, they could have entirely different lives and hobbies. At the level of the text outputs of these people, we could see that person A writes about attending church and religious events, while person B writes about yoga retreats. The model predicting other domains, such as agreeableness, could make different predictions because these two people use entirely different vocabulary, which can impact features extracted from the texts.

The lack of demographic variables compromises the validity and reliability of models based on such datasets. Consequently, the predictive effectiveness of models developed from these sources is generally limited, exhibiting performance metrics ranging from nearly random to a maximum accuracy or F1 score of 86\%. This variability in predictive efficacy highlights the inherent limitations of the models based on these datasets.

%This need poses a practical problem because there are few publicly available datasets due to privacy issues (e.g., Facebook does not allow access to private statuses) and prohibitively high labeling costs. In addition, most of the available datasets have several shortcomings related to the small number of users or texts per user or the lack of demographic data, lack of anonymity (which makes users reluctant to express their true personality), limited expressiveness (e.g., on Twitter), low topic diversity, or a strong focus on personality-related topics (e.g., on personality forums) [11, 12]. 
%Until recently, the only Big 5 labeled datasets publicly available were the two small datasets from [13] and [14]. The former is based on essays, and the latter consists of a small portion of the MyPersonality dataset from Facebook statuses. While useful, these datasets are suitable for only a limited set of research questions and methods, consequently rendering much of the results obtained on these datasets of limited interest for personality research. 

\subsection{Open challenges}
%\todo{connect this to RAM stages?}
We can categorize the open challenges into two groups. The first group pertains to the insufficiencies of current datasets, which we discussed in the previous section. These include the absence of datasets labeled with the Big Five personality model, the lack of additional demographic labels, the limited number of users and texts per user, and the lack of topical diversity. The second group of challenges is directly connected to the first and concerns the lack of validity and interpretability of current methods for personality assessment based on text. For instance, the absence of demographic labels makes it impossible to control for confounding factors that influence language generation processes. The anonymity of users affects topical diversity, which in turn affects the generalizability of predictive models and limits the discovery of new relevant personality cues in the text. The limited number of users and texts generally restricts the types of models that can be used and hinders the development of new methods. For example, modeling user-level representations based on a small number of texts per user is much simpler than when users have many texts.
The inadequate validity and interpretability of the developed models discourage researchers in personality psychology, further widening the gap between the two communities.

While overcoming these challenges is essential, it is a complex undertaking due to the technical, legal, and ethical obstacles in generating new datasets. Technical obstacles may include accessing APIs, implementing web scrapers, or processing vast data while maintaining data quality and consistency. Legal constraints may arise from data protection regulations such as GDPR, which impose stringent data collection, storage, and processing requirements, particularly when handling sensitive information. Ethical considerations include preserving user privacy, preventing biases in dataset creation, and guaranteeing the responsible use of data for research purposes.

\section{Validity and reliability}

One of the main open challenges lies at the core of what separates the research communities of NLP and psychology. As discussed in subsection~\ref{subsection:background_validity_reliability}, while predictive accuracy is one aspect of validity, it represents only one form of validity evidence. This distinction reflects the differing priorities of personality psychology and NLP.

The scrutiny of the validity and reliability of the APR and ATBPA methods has been limited \cite{park2015automatic, kulkarni_latent_2018, tay2020psychometric, novikov2021inferred}, with limited literature that addresses the established norms of personality psychology research \cite{american2014standards}. These norms involve the assessment of different types of validity (such as content and discriminant) and reliability (for instance, test-retest). In ML, validity is associated with the model's interpretability and explainability. Interpretability refers to the degree to which a model allows the observation of cause and effect, whereas explainability denotes the ability to describe the internal mechanisms of a model in understandable terms \cite{gilpin2018explaining, gunning2019xai}. ATBPA models are expected to be both interpretable and explainable, as decisions regarding human traits should be just, grounded in robust evidence, and readily explainable as required.

%Using texts as a source of personality traits makes meeting these requirements even more difficult since language use is usually influenced by the personality of the author and other sociodemographic characteristics such as age, gender, and cultural background [17]. However, many of the existing ATBPA methods are too easily influenced by the varying distributions of some of these factors in the data sample, making the model's predictions susceptible to spurious associations between linguistic features and personality traits.

Furthermore, the lack of interpretability and the presence of confounding variables can lead to inconclusive or biased results and may even raise ethical challenges. According to  \cite{bleidorn2019using}, the evolution of ATBPA has unfolded in three generations. The first generation focused on identifying the linguistic correlates of personality within smaller samples. The second generation aimed at enhancing predictive power, while this new line of research that addresses the aforementioned challenges falls within the third generation. Tackling these challenges is essential as personality-based models become integral to widely used services, such as dialog systems \cite{roller-etal-2021-recipes} and recommender systems \cite{dhelim2021recommender}.

\part{Feature-based personality analysis from text}
\label{part:two}

This part describes our approach to overcome one of the main obstacles in the development of automated text-based personality assessment methods: the scarcity of personality-labeled datasets. The analysis and experiments described in the following chapters are based on work first described in \cite{gjurkovic-snajder-2018-reddit} and \cite{gjurkovic2021pandora}.

To address the limitations of existing datasets (see Chapter~\ref{ch:background_text_based_pa}), we have identified Reddit, a previously underutilized social media site, as a promising new data source. Reddit offers several advantages for personality analysis, including user anonymity, a diverse range of topics within specific subcommunities, and varying levels of interaction between users and content. These characteristics improve the availability of relevant cues necessary for an accurate personality assessment.

%This section outlines our methodology for collecting and utilizing flair-based labels to develop an initial personality-labeled Reddit dataset, which we named \textbf{MBTI9k}. 

The main prerequisite to unlocking the full potential of using the new data source is to find a reliable and effective way to extract personality and demographic labels. In this part, we will describe our methodology for labeling user-created Reddit texts using the most commonly used type-based personality model (MBTI), the trait-based personality model (the Big Five), and demographics. This methodology relies on previously used full-text information extraction techniques and a novel method using Reddit-specific descriptions named flairs.

To bridge the gap identified in previous research between the personality psychology and computer science communities, we developed a new Reddit dataset tailored for valid personality psychology studies. Incorporating Big Five scores and demographic labels was essential for two main reasons. First, known and unknown confounding variables can lead to incorrect conclusions and low generalizability. For instance, in a scenario where all Reddit users high in openness are males, the model may resort to an easier solution of predicting high openness by associating it with language typically used by males (such as discussions about sports and cars) rather than accurately predicting personality traits. Secondly, since this is a new data source, it was unknown to what extent personality and demographic variables are predictable.

In the subsequent chapters, we show the results of a preliminary analysis of language cues associated with personality on Reddit, compare these findings with related works derived from other data sources, and validate the dataset in terms of both face and predictive validity. To do this, we used traditional methods for text-based personality assessment. First, we extracted features using commonly used open- and closed-vocabulary approaches and analyzed user behavior patterns. We then employed these features in training and evaluating benchmark classifiers to predict the MBTI types and the Big Five scores of individual users based on their language cues.
Next, we present the benefits of the new dataset through three diverse experiments covering confirmatory study, the importance of considering demographic information in personality research, and a pragmatically oriented experiment in which we demonstrate that it is possible to predict the Big Five scores based on the MBTI labeled dataset. We end this chapter by discussing the implications and limitations of using Reddit data for personality analysis.
%Furthermore, we present baselines for personality and demographics prediction on \pandora. We treat targets such as Big Five personality traits and other demographic variables as targets for supervised machine learning, and we evaluate various benchmark models with different feature sets. 

\chapter{Personality and demographics on Reddit}
\label{ch:personality_demographics_reddit}
\section{Reddit}
Reddit is a social media and news aggregation platform that enables users to submit various forms of content, such as text posts, images, or links, and engage in online communities known as \textit{subreddits}, each dedicated to a specific topic. In 2024, Reddit had more than 365 million active weekly users,\footnote{\href{https://investor.redditinc.com/financials/sec-filings/default.aspx}{https://investor.redditinc.com/financials/sec-filings/default.aspx}} making it one of the largest and most popular websites since its inception in 2005.

On average, users spend about 15 minutes per visit on Reddit. The predominant demographic age is between 18 and 29 years old, representing 44\% of the site's user base. A 2024 survey\footnote{\href{https://www.statista.com/statistics/1255182/distribution-of-users-on-reddit-worldwide-gender}{https://www.statista.com/statistics/1255182/distribution-of-users-on-reddit-worldwide-gender}} revealed that most of the Reddit users are male (61\%), with the highest concentration of users residing in the United States, followed by the United Kingdom and Canada.

The Reddit platform facilitates information discovery and sharing by organizing content around subreddits. Users create and moderate these subreddits, each with its own rules and focus, covering topics such as news, politics, sports, gaming, and hobbies. Interaction with content occurs through upvoting, downvoting, commenting, or initiating discussions.

Reddit's anonymity policy is a key distinguishing feature that empowers users to express their thoughts freely and discuss sensitive topics without fear of judgment or ridicule. Reddit has a highly engaged community where lively discussions and debates are commonplace, enabling users to participate in online communities with similar interests.

\subsection{Accessing Reddit data}
There are different ways to programmatically access Reddit data. The Reddit API enables developers to access and extract data from the Reddit platform. The API provides access to all the data stored on Reddit, including posts, comments, votes, and user data. To use the Reddit API, developers must create a Reddit account and register an application. However, the terms and conditions for using the Reddit API have changed since its inception. Although once it was completely open to anyone wanting to access data, it is only granted on a by-case basis today. Implementations in different programming languages allow for easier use of the Reddit API. For Python, one can use the PRAW implementation.\footnote{\url{https://praw.readthedocs.io/}}

Until 2023, a service called Pushshift \cite{baumgartner2020} allowed searching and accessing a Reddit comment and submission dataset. The dataset includes comments and submissions from all Reddit communities, including public and private subreddits. 

The accessibility of Reddit data, the quality of the textual content, and the diversity of topics discussed contribute to its use in training NLP models, notably in developing large language models and conversational systems. A deeper understanding of the properties of these data can help identify biases inherent in the training datasets, thereby informing further development of NLP models.

\section{Labeling methods}
After recognizing Reddit as a potentially valuable source of digital cues about personality, the next step was to acquire labels relevant to personality analysis. In this section, we will detail how to obtain labels for the two most commonly used personality models in text-based analysis, MBTI and Big Five, as well as key demographic variables such as gender and age. Throughout this work, we refer to these acquired labels as \emph{gold labels} -- the ground truth annotations against which prediction models are compared.

Personality-relevant labels come in different flavors. Type-based personality models use short strings from a predefined set of labels. For example, MBTI has 16 different types. Similarly, location, gender, and age are also typically very short and can fit most platforms' description boxes. To support and even encourage descriptors on the user and comment levels, Reddit uses flairs -- descriptors commonly predefined on a subcommunity (i.e., subreddit) level. However, the Big Five scores are mostly shared by copy-pasting test results taken outside Reddit in the comments and are typically longer and contain both the names of traits and facets and their numerical scores. Because of that, we employ two methods for extracting personality-relevant labels: (1) flair-based methods and (2) full-text extraction methods. In the case of some labels, to increase coverage, we opted to use the combination of (1) and (2) as they are being reported in both ways.

\subsection{Flair-based labeling}
\label{section:flair_based_labels}
Despite acknowledging critiques about the validity of the MBTI personality model (Section~\ref{section:background_personality_models}), its widespread use as a tool for self-discovery made it an appropriate foundation for creating a new personality-labeled dataset. The popularity of this model is particularly evident on Reddit, where numerous subreddits dedicated to personality psychology encourage members to publicly declare their personality types, including MBTI, utilizing \textbf{flairs}. The flairs serve as specific markers that help categorize users and their contributions according to classifications relevant to the context of different subreddit communities. In forums focused on personality, participants often share their MBTI types and may complement this information with their demographic details.

Flairs can also tag content within a subreddit based on specific categories or topics. For example, a cooking subreddit may use flairs to categorize posts based on cuisine types or difficulty levels. In contrast, a music subreddit may use flairs to identify specific genres or artists. This tagging system lets users quickly find content relevant to their interests or needs.

In subreddits dedicated to exploring MBTI, such as \textit{/r/MBTI} and \textit{/r/ENFP}, participants frequently use flairs to denote their MBTI types. In addition to their MBTI type, these users often add more details such as hobbies, educational background, favorite books or movies, astrological signs, and classifications from other personality frameworks (for instance, Big Five, DISC, or Socionics). Some flairs even incorporate humorous or self-deprecating tags, like \textit{Overthinking Thinker} to represent thinking-dominant MBTI types or \textit{Chronic Daydreamer} for the INFP personality type. Nevertheless, flairs can be challenging since they are sometimes ambiguously phrased or require specific knowledge for correct interpretation. For instance, a flair might include jargon specific to a niche hobby or presume familiarity with the dimensions of the Big Five personality model.

Despite these challenges, we decided to utilize the MBTI types declared in the user's flairs as their personality labels. Our rationale was that users who specified their MBTI type in their flair likely had undergone at least one personality assessment or had self-identified with the descriptions of a particular MBTI type. In addition, users in the dataset are more likely to identify with a particular MBTI type because they have more prominent underlying traits, making the dichotomization of continuous traits less of an issue. Analysis of user comments reinforced this assumption, showing that most of the individuals who mentioned their MBTI type had not only taken multiple personality tests that gave them the same type, but also had a substantial understanding of MBTI concepts.

In creating our dataset, we prioritized precision over recall. We preferred having fewer users with verified MBTI labels rather than a larger pool with less certainty regarding their MBTI types. The collection process unfolded in several stages.

The initial step involved identifying users who had mentioned an MBTI type in their flair, which led to the compilation of various flairs per user. However, due to the vague nature of some flairs, we encountered numerous inaccuracies. This necessitated an additional step to discern and isolate flairs that indicated a user's MBTI type, filtering out ambiguous or misleading ones.

To accomplish this, we implemented a regex-based pattern matching strategy (regular expression) designed to filter the flairs effectively. This approach had three main objectives: (1) to pinpoint flairs that explicitly referenced MBTI types, (2) to classify flairs with ambiguous meanings, and (3) to exclude any flairs that did not align with our specific criteria.

\subsubsection{Manual refinement of flair-based labels}
We assessed the flairs deemed ambiguous, ignoring those that remained unresolved. For instance, a flair like \textit{ENXJ} suggested uncertainty about the user's thinking or feeling preference, making it unreliable for our purposes. Next, we organized the flairs by individual users and scrutinized the consistency of their MBTI types. This step was necessary because users might modify their flairs or use different ones across various subreddits. In our final selection, we excluded users whose MBTI type was not uniquely identifiable or consistent. This process ensured the integrity and reliability of our dataset, focusing on the collection of precise and clear MBTI labels.

Subsequently, we discovered that certain MBTI types were significantly underrepresented in the dataset, such as only 16 ESFJ and 23 ESTJ users. To address this issue, we augmented the dataset by performing a full text search of comments on the MBTI subreddit, excluding flairs. We used strict high-precision patterns, such as ``I am (an) $\langle$type$\rangle$'' and similar variations, to search for users' self-declaration of the specific underrepresented type. We manually evaluated the comments and removed any false positives before adding the remaining users to the dataset.
Finally, we gathered all posts and comments from the users identified in the previous steps, covering the period from January 2015 to November 2017.

\subsection{Full-text labeling}
\label{section:dataset_full_text_labels}
%\subsubsection{Big Five}
Labeling personality traits using the results of the Big Five test is more challenging than MBTI since the Big Five tests result in scores for each of the five traits. Furthermore, the scoring format is not standardized, and scores are typically reported in comments replying to posts that mention a specific online test, rather than in flairs. Normalization of scores presents several challenges. Firstly, different websites use different personality tests and inventories, such as HEXACO, NEO PI-R, and Big Five Aspect Scale, which can be publicly available or proprietary. The distribution of tests and inventories is shown in Table \ref{tbl:Big5_tests_distribution}. Secondly, the tests use different names for traits, such as emotional stability as the opposite of neuroticism, or use abbreviations, such as OCEAN, where N denotes neuroticism. Third, the format in which the scores are presented varies. Test scores may be reported as raw scores, percentages, or percentiles. Percentiles can be calculated based on the distribution of users that took the test or on the distribution of specific groups of offline test-takers, such as students, with the latter commonly adjusted for age and gender. Scores can be numeric or descriptive, with the former in different ranges such as \(-100\)--\(100\), \(0\)--\(100\), or \(1\)--\(5\), and the latter being different for each test. For example, descriptions such as typical and average may map to the same underlying score. In addition, users may copy their results, paraphrase them, make occasional spelling errors in trait names, or even mix both methods. Lastly, in certain instances, results are not derived from traditional inventory-based evaluations, but are instead generated by text-based personality prediction platforms, such as Apply Magic Sauce\footnote{\href{https://applymagicsauce.com}{https://applymagicsauce.com}} and Watson Personality.\footnote{\href{https://applymagicsauce.com}{https://web.archive.org/web/20200718043442/https://www.ibm.com/cloud/watson-personality-insights}}

%Table \ref{tbl:Big5_tests_distribution} shows the distribution of
%personality tests and inventories.

\subsubsection{Semi-automatic extraction of Big Five scores}

Automating the extraction of Big Five scores is challenging, given their non-standard nature and representation in full-text comments rather than flairs. Therefore, we opted for a semi-automated strategy. First, we identified candidate comments that contained the three most frequently correctly spelled traits: Agreeableness, Openness, and Extraversion. We then retrieved the corresponding post and used the provided link to identify the specific test to which the comment alluded. Comments linked to text-based prediction services were excluded and we used regular expressions, each tailored to a specific test, to extract personality scores from the comments. A detailed manual review of all extracted scores and associated comments was performed to ensure accuracy. Approximately 80\% of the scores were extracted correctly, with the remaining 20\% extracted manually. The result was the Big Five scores for 1,027 users in 12 different tests. Comments not belonging to a known test were omitted at this stage. To extract the ratings from these comments, we developed a test identification classifier trained on the reports of 1,008 users using character n-grams as features. This classifier achieved an F1 macro score of 81.4\% on reserved test data. Then, this classifier was used to determine the tests given in the remaining comments, and the score extraction process was repeated. In this way, we could identify the scores of 600 additional users, bringing the total number of users with Big Five scores to 1,608.

 \begin{table}
 \centering
{\small
\caption{\label{tbl:Big5_tests_distribution} The number of authors of comments containing reported scores for gold and predicted online tests}}
    \begin{tabular}{llrr}
    \toprule
    & &\multicolumn{2}{c}{\#\,Users}\\
    \cmidrule(lr){3-4}
    Online test               & Used Inventory          & Gold & Pred\\
    \midrule
    Truity               & Proprietary
     
   % \alert{Goldberg's unipolar markers ( \citeyear{goldberg1992development}) markers}  
    & 378      & 319       \\

    Understand Myself   & Big Five Aspects       & 268      & 146     \\ 
    IPIP 120             & IPIP-NEO      & 120      & 82      \\
    IPIP 300             & IPIP-NEO       & 60       & 17      \\
    Personality Assesor & BFI         & 66       & 10      \\
    HEXACO               & HEXACO-PI-R               & 49       & 1       \\
    Outofservice         & BFI               & 38       & 11      \\
    Qualtrics            & --                   & 19       & 8       \\
    123test              & IPIP-NEO       & 11       & 6       \\
    BFI-2                 & BFI & 1 & 0 \\
    See My Personality     & IPIP NEO& 1 & 0 \\
    \bottomrule
\end{tabular}
\end{table}

\subsubsection{Normalizing Big Five Test Scores}

To normalize the extracted scores, we first map the various descriptive labels to numerical values from 0 to 100 in increments of 10 (\textit{very\_low}: 0–10; \textit{low}: 10–20; \textit{low\_average}: 20–40; \textit{average}: 40–60; \textit{average\_high}: 60–80; \textit{high}: 80–90; \textit{very\_high}: 90–100). Since scores could be reported as raw values, percentiles, or descriptive labels, we used test‐specific information to interpret them correctly. The raw scores and percentiles reported by Truity\footnote{\url{https://www.truity.com}} and HEXACO\footnote{\url{http://hexaco.org/hexaco-online}} were converted to percentiles using the distribution parameters of each test (HEXACO publishes its parameters publicly, while Truity provided us with its participant data).

\subsection{Hybrid labeling}
\subsubsection{Demographic labels} 
\label{ref:demographics}

After analyzing MBTI-related subreddits, it became clear that Reddit users shared their MBTI types, discussed their results from other personality tests, and often provided context by sharing demographic information, such as age, gender, and location. This information was particularly prevalent in subreddits related to sports, fitness, or advice.
To determine the age, gender, and location of each user, we again rely on the textual descriptions in the flairs. To do this, we collected all unique flairs from all comments from the 10,228 users in our dataset and then manually checked for relevant information on age, sex and location. If a user disclosed their age in more than one flair at different times, we used the most recently provided age. In addition, we also performed a full-text search for self-reported comments on age (e.g., ``I am 18 years old'') and gender (e.g., ``I am female/male'') at the comment level.

Users frequently provide location information at various levels, such as countries, states, cities, continents, and regions. To standardize this information, we normalized the place names and matched countries with their corresponding country codes, along with associating countries and states with the appropriate continents and regions.
%\todo{add more details; what resources we used}

When mapping states to regions, we found that the United States and Canada were divided into different regions. To address this, we used five regions for the United States and three for Canada (with no users in one region). %Table~\ref{tab:types} shows the distribution of users per country and region. 

%\begin{table}
%\centering
%{\small
%    \begin{tabular}{lllr}
%	    \toprule
%    Country       & Region Name & Region Code & \#\,Users \\
%    \midrule
%    United States & West        & w           & 208     \\ 
%    ~             & Midwest     & mw          & 153     \\
%    ~             & Southeast   & se          & 144     \\
%    ~             & Northeast   & ne          & 138     \\
%    ~             & Southwest   & sw          & 100     \\
%    \midrule
%    Canada        & West        & cw          & 50      \\
%    ~             & East        & ce          & 44      \\
%    \bottomrule
%\end{tabular}
%}
%\caption{\label{tbl:region_distribution} Regional user distribution}
%\end{table}

\section{Descriptive statistics}
\label{section:dataset_descriptive_statistics}

Based on the acquired labels, we created two datasets that were made publicly available: (1) MBTI9k based on MBTI type prediction and analysis, and (2) \pandora dataset, which is additionally augmented with Big Five labels and demographic information.

The MBTI9k dataset is extensive, comprising 22,934,193 comments, which amount to 583,385,564 words, in 36,676 subreddits from 13,631 distinct users, along with 354,996 posts (totaling 921,269 words) from 20,149 subreddits by 9,872 unique users. At its creation, this dataset represented the largest compilation of personality-labeled words, surpassing more than eight times the word count in the largest study to date \cite{schwartz2013personality}.

In \pandora, we retrieved all comments from users with Big Five labels from 2015 and added them to the MBTI9k dataset. The enlarged user base and stricter time filtering resulted in a dataset of 17,640,062 comments posted by 10,288 users. \pandora includes 393 users that are labeled Big Five and MBTI.

The hierarchical structure of the Reddit data allows for analysis at various levels, including individual posts and comments, diverse subreddit communities, and temporal trends. We begin with a fundamental descriptive analysis of the dataset, then explore a more detailed analysis of features in Chapter~\ref{ch:personality_features} to showcase the applicability of the dataset in personality research.

\label{subsubsection:dataset_mbti}
\subsubsection{MBTI}
Table~\ref{tab:merged_mbti} provides insights into the distribution of Reddit users in the MBTI9k dataset based on their MBTI types and the individual MBTI dimensions. For context, the first column compares these distributions to those estimated for the U.S. population.\footnote{\href{https://www.capt.org/products/examples/20025HO.pdf}{https://www.capt.org/products/examples/20025HO.pdf}} Our data reveal a significant presence of Redditors who exhibit introverted, intuitive, thinking, and perceiving traits. This distribution resembles the personality traits of gifted adolescents \cite{sak2004synthesis} and aligns with the trend showing that Reddit users tend to be more educated than the average internet user.\footnote{\href{https://www.alexa.com/siteinfo/reddit.com}{https://www.alexa.com/siteinfo/reddit.com}\label{alexa}} The last columns of the table present the number of users for all 16 MBTI types. Four MBTI types (INTP, INTJ, INFP, and INFJ) represent 75 percent of all users, indicating a shift in personality distribution compared to the general public.

\begin{table}[t]
\centering
\small
\caption{Distributions of MBTI types and dimensions in US general public and on Reddit}
\begin{tabular}{lccccc}
\toprule
\textbf{Type} & \textbf{\% USA} & \textbf{\% comm} & \textbf{\% post} & \textbf{\% MBTI9k} & \textbf{\# Users} \\
\midrule
INTP & 3.3  & 22.3 & 26.8 & 25.3 & 2833 \\
INTJ & 2.1  & 17.2 & 20.6 & 20.0 & 1841 \\
INFJ & 1.5  & 11.2 & 12.9 & 11.1 & 1051 \\
INFP & 4.4  & 11.0 & 13.3 & 11.6 & 1071 \\
ENFP & 8.1  & 6.1  & 7.4  & 6.6  & 162  \\
ENTP & 3.2  & 6.1  & 7.4  & 6.7  & 627  \\
ENTJ & 1.8  & 5.3  & 2.8  & 3.9  & 319  \\
ISTP & 5.4  & 5.2  & 3.7  & 4.8  & 408  \\
ISTJ & 11.6 & 3.4  & 1.3  & 2.4  & 194  \\
ENFJ & 2.5  & 3.3  & 1.1  & 2.3  & 616  \\
ISFJ & 13.8 & 2.4  & 0.7  & 1.3  & 109  \\
ISFP & 8.8  & 2.3  & 0.7  & 1.6  & 123  \\
ESTP & 4.3  & 1.2  & 0.5  & 0.9  & 71   \\
ESFP & 8.5  & 1.1  & 0.3  & 0.7  & 51   \\
ESTJ & 8.7  & 1.0  & 0.3  & 0.5  & 43   \\
ESFJ & 12.3 & 0.8  & 0.2  & 0.4  & 29   \\
\midrule
\multicolumn{6}{l}{\textbf{Dimension‐level distribution}} \\
\midrule
\textit{Introverted}  & 50.7 & 75.1 & 80.0 & 78.1 & 7134\\
\textit{Extroverted}  & 49.3 & 24.9 & 20.0 & 21.9 & 1920\\
\midrule
\textit{Sensing}      & 73.3 & 17.4 & 7.7  & 12.6 & 1030\\
\textit{Intuitive}    & 26.7 & 82.6 & 93.3 & 87.4 & 8024\\
\midrule
\textit{Thinking}     & 40.2 & 61.7 & 63.4 & 64.4 & 5837\\
\textit{Feeling}      & 59.8 & 38.3 & 36.6 & 35.6 & 3217\\
\midrule
\textit{Judging}      & 54.1 & 44.6 & 39.9 & 41.8 & 3752\\
\textit{Perceiving}   & 45.9 & 55.4 & 61.1 & 58.2 & 5302\\
\midrule
\bottomrule
\end{tabular}
\label{tab:merged_mbti}
\end{table}

Table~\ref{tab:common_subs} provides an alternate perspective, showcasing the distribution of subreddits categorized by the variety of MBTI types participating in them. Notably, nearly 47\% of subreddits are frequented by users of a single MBTI type, while a mere 1.45\% of subreddits (534 in total) see participation from all 16 MBTI types. Despite being a small fraction of the total dataset, this data can be used for comparative analyses among different personality types.

%\FloatBarrier

\begin{table}[t]
	{\small
	\begin{center}
    	\caption {Distribution of subreddits by the number of distinct MBTI types of participating users}
		\begin{tabular}{rrr|rrr}
		\toprule
		\#\,types & \#\,subred. & \% & \#\,types & \#\,subred. & \% \\
		\midrule
		1              & 17222      & 46.96             &
		9              & 729        & 1.99              \\
		2              & 5632       & 15.36             &
		10             & 640        & 1.75              \\
		3              & 3105       & 8.47              &
		11             & 567        & 1.55              \\
		4              & 2034       & 5.55              &
		12             & 512        & 1.4               \\
		5              & 1540       & 4.2               &
		13             & 443        & 1.21              \\
		6              & 1217       & 3.32              &
		14             & 377        & 1.03              \\
		7              & 964        & 2.63              &
		15             & 362        & 0.99              \\
		8              & 798        & 2.18              &
		16             & 534        & 1.46              \\
	\bottomrule
	\end{tabular}
	\label{tab:common_subs}
	\end{center}}
\end{table}

%\FloatBarrier
%\subsubsection{Big Five Tests}

%\begin{table}
%  \centering
%{\small
%    \begin{tabular}{lr|lr}
%	    \toprule
%    Country        & \#\,Users  & Region            & \#\,Users \\
%    \midrule
%    US             & 1107       & US West           & 208       \\
%    Canada         & 180        & US Midwest        & 153       \\
%    UK             & 164        & US Southeast      & 144       \\
%    Australia      & 72         & US Northeast      & 138       \\
%    Germany        & 53         & US Southwest      & 100       \\
%    Netherlands    & 37         & Canada West       & 50        \\
%    Sweden         & 33         & Canada East       & 44        \\
%    \bottomrule
%\end{tabular}\caption{\label{tbl:location_dist} Geographical distribution of users per country and region (for US and Canada)}}
%\end{table}
%
%\begin{table}
%  \centering
%{\small
%    \begin{tabular}{lr|lr}
%	    \toprule
%    Country      & \#\,Users & Continent    & \#\,Users \\
%   \midrule
%   US  & 1107    & North America & 1299  \\
%   Canada         & 180     & Europe        & 580   \\
%   UK & 164     & Asia          & 103   \\
%   Australia      & 72      & Oceania       & 85    \\
%   Germany        & 53      & South America & 24    \\
%   Netherlands    & 37      & Africa        & 4     \\
%    \bottomrule
%\end{tabular}\caption{\label{tbl:location_dist} Geographical distribution of users per country and region (for US and Canada)}}
%\end{table}

\subsubsection{The Big Five}

\begin{figure}
    \centering
    \vspace{-1em}
    \hspace*{-0.5cm}
    \includegraphics[scale=0.28]{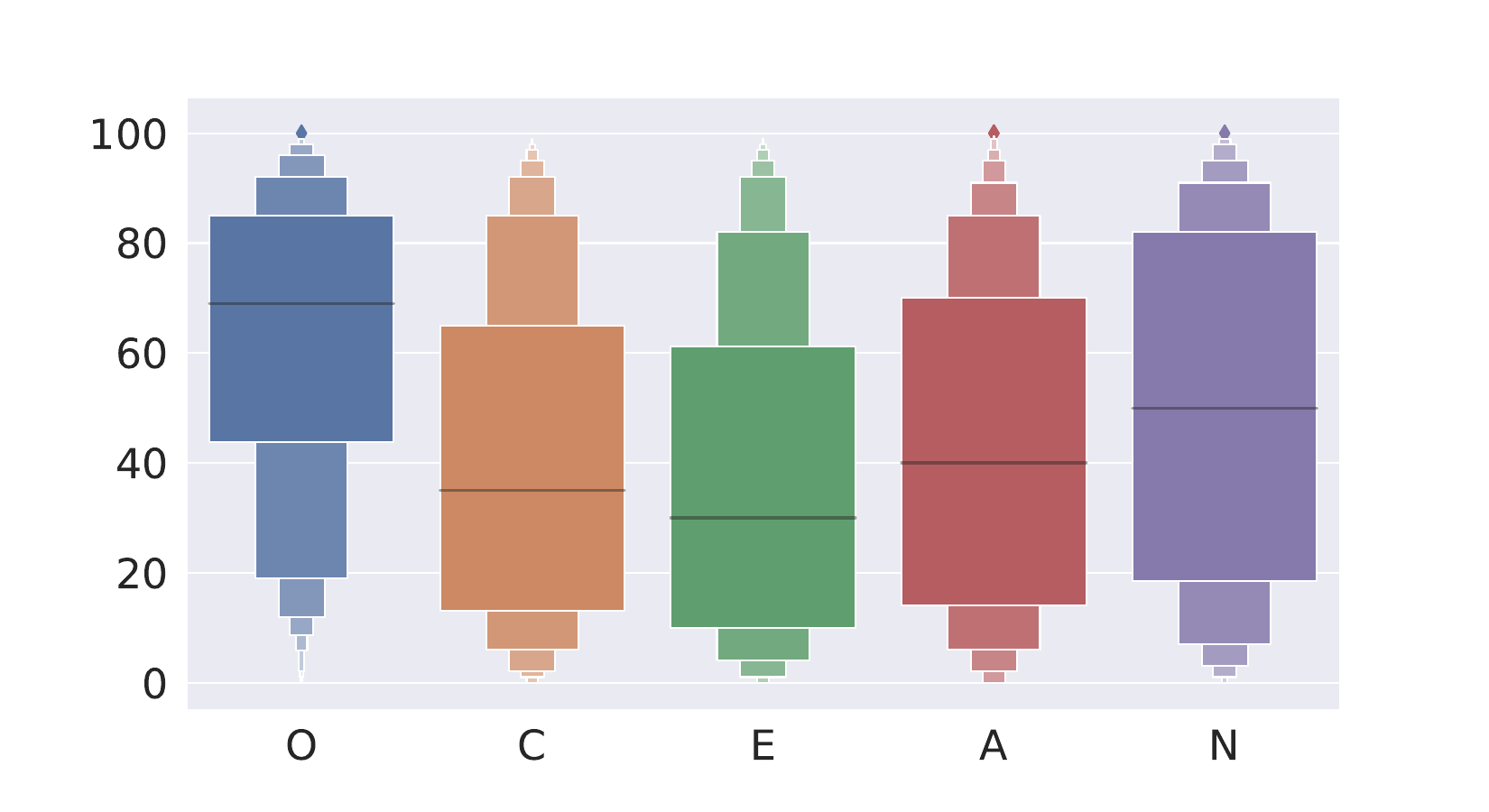}
    \vspace*{-0.5em}
    \caption{Distribution  of  Big  5  percentile  scores}
    \label{fig:distribBig5}
\end{figure}

Table~\ref{tbl:big5_mean_std_distribution} and Figure~\ref{fig:distribBig5} illustrate the distribution of Big Five scores across traits in the \pandora dataset. We find that the average user in our dataset has rather average scores for neuroticism, higher scores for Openness, and lower scores for Extraversion, Agreeableness, and Conscientiousness. Table~\ref{tbl:big5_mean_std_distribution} contains means and standard deviations for descriptions and percentiles of each Big Five trait. Table~\ref{tbl:Big5_tests_distribution} shows the distribution of tests and their inventories in \pandora. In addition, males in our dataset were significantly younger than females (mean age: males = 25.68 vs. females = 26.67; t = 3.356, p = 0.000807) and scored lower on agreeableness (mean agreeableness: males = 38.88 vs. females = 44.12; t = 2.026, p = 0.043319). Moreover, Table~\ref{tab:merged_mbti} shows that MBTI-labeled users are higher on introversion, intuition (i.e., openness), thinking (i.e., less agreeableness), and perceptiveness (less conscientiousness). This is not surprising given the high correlations between specific MBTI dimensions and Big Five traits, as shown in Table~\ref{tbl:mbti_big5_corr_valid}. These correlations are consistent with existing psychological research \cite{mccrae1989reinterpreting}.

%The correlations between the Enneagram and the Big Five are also informative. For example, Extraversion is positively correlated with Enneagram Type 7, which describes a cheerful and outgoing person, and negatively correlated with Type 5, which is typically more reserved and introverted.

Given the high correlations between the personality models and the multiple personality labels of some users, our next experiment focuses on examining these relationships in more detail. Additionally, since there are more MBTI users than Big Five users, we want to explore these models more deeply in the context of domain adaptation.

\begin{table}[t!]
   \centering
{\small
    \setlength{\tabcolsep}{5pt}
    %-- Now we have 6 columns: the new 2 columns + the original 4 columns
    \caption{\label{tbl:big5_mean_std_distribution} Means of normalized Big Five percentile scores (n=1608), age (n=2324), number of comments per user (n=10,255) and Big Five results distribution on different reported scales for 1,652 users. The Descriptions (n=155) and Percentile (n=365) columns show the number of users who reported their scores on the description and percentile scales, respectively.}
}
	\begin{tabular}{@{}lccrrr@{}}
	    \toprule
%	    & \multicolumn{2}{c}{All} & \multicolumn{2}{c}{Male} & \multicolumn{2}{c}{Female}\\
    \textbf{Big Five Trait} & \textbf{All} & \textbf{Females} & \textbf{Males} & \textbf{Descriptions} & \textbf{Percentiles} \\
    \midrule
    Openness                & 62.5  & 62.9 & 64.3& $67.37 \pm 26.76$     & $67.27 \pm 26.87$\\
    Conscientiousness     & 40.2  & 43.3 & 41.6& $41.29 \pm 27.97$     & $40.48 \pm 30.22$  \\
    Extraversion            & 37.4  & 39.7 & 37.6& $38.70 \pm 27.53$     & $37.09 \pm 31.16$\\
    Agreeableness           & 42.4  & 44.1 & 38.9& $50.10 \pm 29.10$     & $42.39 \pm 30.89$\\
    Neuroticism             & 49.8  & 51.6 & 46.9& $55.95 \pm 31.11$     & $52.82 \pm 31.97$\\
    \midrule 
    Age                                                      & 25.7  & 26.7 & 25.6 & -- & --\\
    \midrule 
    \# Comments                          & 1819  & 2004 & 3055&--&--\\ 
    \# EN comments                      & 1714  & 1908 & 2878&--&--\\
    \bottomrule
\end{tabular}

\end{table}

  \begin{table}
   \centering
{\small
\caption{\label{tbl:intersection} Intersection details for personality models and the total number of unique labels}}
    \begin{tabular}{lrrrr}
	    \toprule
       Variable &  Big Five  & MBTI  &  Unique \\
    \midrule
    Gender & 599 & 2695  & 3084 \\
    Female & 232 & 1184  & 1331\\
    Male &  367 & 1511  & 1753\\
    Age & 638 & 1890  & 2324\\
    Country & 235 & 1984  & 2146\\
    Region & 74 & 800  & 852  \\
    Big Five & -- & 393  &1608 \\
    MBTI & 393 & -- & 9084 \\

    \bottomrule
\end{tabular}

\end{table}

\begin{table}[t]

  \centering
{\small
%<<<<<<< HEAD
\caption{\label{tbl:mbti_big5_corr_valid} Correlations between gold MBTI, and Big Five. Significant correlations (p\textless{.05}) are bolded.}
}
\begin{tabular}{lrrrrr}
\toprule 
%Gold & & & Predicted & \\
MBTI / Big Five & O &C & E & A & N \\
\midrule
Introverted       & --.062  & --.062	 &\textbf{--.748}  & --.055 &\textbf{.157}	 \\
Intuitive         &\textbf{ .434}  & --.027 &--.042 & .030 &.065 \\
Thinking          &--.027	 & \textbf{.138}	 &--.043  & \textbf{--.554}&  \textbf{--.341}\\
Perceiving        &\textbf{.132} &\textbf{--.575}	 &\textbf{.145}  & .055& .031\\
%\midrule
%Enneagram 1 & --.139 & \textbf{.271}& --.012 & .004 & --.163 \\
%Enneagram 2 &.038 & \textbf{.299}& .042 & \textbf{.278} & --.034 \\
%Enneagram 3 & .188 & .004& .143 & --.069 & --.097 \\
%Enneagram 4 & .087 & --.078& --.137 & \textbf{.320} &\textbf{.342} \\
%Enneagram 5 & --.064 & .006 & \textbf{--.358} & --.157 & --.040 \\
%Enneagram 6 & --.026 & .003 & --.053 & --.007 & \textbf{.276} \\
%Enneagram 7 & .015 & \textbf{--.347} & \textbf{.393} & --.119 & \textbf{--.356} \\
%Enneagram 8 & --.127 & .230& .234 & \textbf{--.363} & --.179 \\
%Enneagram 9 & --.003 & --.155& --.028 & .018 & .090\\
\bottomrule
\end{tabular}

\end{table}

\subsubsection{Location and language}
Table~\ref{tbl:combined_location_dist} shows that most users in our dataset are from English-speaking countries, with an even regional distribution across the United States and Canada. To identify the languages each user speaks, we use language identification techniques \cite{joulin2016bag, joulin2016fasttext} at the comment level and calculate the percentage of comments written in English for each user. Table~\ref{tbl:lang_dist} shows the distribution of comments in the identified languages. 
Using this information, we estimate the number of languages users speak by setting a threshold of 30 comments per language to avoid false positives. %Table \todo{add ref to table} then shows the distribution of languages spoken by each user.
We also applied the langid\footnote{\href{https://github.com/saffsd/langid.py}{https://github.com/saffsd/langid.py}} language identification tool to all user-generated content. This analysis reveals that over 97\% of the users predominantly write in English. Moreover, web traffic data indicates that around 76.4\% of Reddit's visitors originate from countries where English is the primary language.\footref{alexa}

Table~\ref{tbl:big5_mean_std_distribution} shows the average number of comments per user, while Table~\ref{tbl:intersection} provides the number of intersections between personality models and other demographic variables.
%\FloatBarrier

\begin{table}
  \centering
{\small
\caption{\label{tbl:combined_location_dist} Combined geographical distribution of users per country, region, and continent}}
    \begin{tabular}{lr|lr|lr}
	    \toprule
    Country        & \#\,Users  & Region            & \#\,Users & Continent       & \#\,Users \\
    \midrule
    US             & 1107       & US West           & 208       & North America   & 1299      \\
    Canada         & 180        & US Midwest        & 153       & Europe          & 580       \\
    UK             & 164        & US Southeast      & 144       & Asia            & 103       \\
    Australia      & 72         & US Northeast      & 138       & Oceania         & 85        \\
    Germany        & 53         & US Southwest      & 100       & South America   & 24        \\
    Netherlands    & 37         & Canada West       & 50        & Africa          & 4         \\
    Sweden         & 33         & Canada East       & 44        & -          & -       \\
    \bottomrule
\end{tabular}

\end{table}

\begin{table}[h]
  \centering
{\small
\caption{\label{tbl:lang_dist} Language distribution}}
    \begin{tabular}{lr|lr}
        \toprule
    Language         & \#\,Comments    & Language        & \#\,Comments \\
    \midrule
    English          & 16637211        & Dutch           & 30219        \\
    Spanish          & 87309           & Esperanto       & 19501        \\
    French           & 72651           & Swedish         & 16880        \\
    Italian          & 64819           & Polish          & 15134        \\
    German           & 63492           & Portuguese      & 32037        \\
    \bottomrule
\end{tabular}

\end{table}

%\FloatBarrier 

\section{Limitations}
\subsection{Selection biases}

There are different reasons why the distributions of the extracted labels are not uniform and do not follow the distributions found in the general public and other data sources.
First, it is important to note that our datasets may not represent all Reddit users, Internet users, or the general population. While the method for labeling personality and demographics described earlier proved effective in creating a dataset with high precision, it is important to recognize the inherent selection bias in our dataset. This bias arises primarily because our dataset is limited to users knowledgeable about the MBTI or the Big Five and actively engage in MBTI-focused subreddits on a specific platform (i.e., Reddit). Furthermore, in the case of MBTI, it includes only users familiar with the concept of flairs and choose to use them to indicate their MBTI type. This approach automatically excludes users who have not commented or do not use flairs, potentially skewing the dataset. 
Similar selection biases are present when extracting the Big Five scores, as only users who had taken the Big Five tests and reported their results in the comments section were considered.

\subsection{Normalization biases}
There are additional sources of uncertainty when extracting and normalizing reported scores, as they depend both on the type of test taken and the norms used to convert raw scores to percentiles. In addition, scores were reported in several forms: descriptions, raw scores, or percentiles.

In the case of MBTI and demographics, applying a regex-based filtering system and eliminating users who exhibited uncertainty about their MBTI type or demographics may introduce additional biases. These factors collectively contribute to a dataset that may not fully represent the broader Reddit community. Consequently, when referring to a \textit{Reddit user} or \textit{Redditor} in this thesis, these terms should be understood within the framework of these specific limitations and biases. Maintaining this awareness can aid in interpreting the findings derived from this dataset.

%\subsection{Methodology}

\chapter{Personality-relevant features}
\label{ch:personality_features}
To evaluate the value of the Reddit-based MBTI9k and \pandora datasets for ATBPA, we performed a comparative analysis replicating several commonly used feature extraction methods from previous studies on personality detection described in Chapter~\ref{ch:background_text_based_pa} and Section \ref{section:background_features}. This allowed us to assess the degree to which our datasets provide new insights beyond what has already been established in the literature.

Furthermore, we performed a face validity check on the MBTI9k dataset to ensure that the extracted features accurately represent the expected MBTI dimensions and types. This step was critical in validating that the dataset effectively reflects the self-reported personality traits of the users. The following sections describes the feature extraction and analysis process in more detail.
\section{Resources and methods}
\label{sec:features}
%\subsection{Feature Extraction}

%\todo{move from datasets to methodology}
%\todo{refer to background sections and feature extraction methods}
Features that we extracted can be classified into two primary groups: linguistic features (see Section~\ref{section:background_features}) extracted from user comments and behavioral features related to user activity patterns. Most of the features can be used on any level of analysis, whether a single comment or on aggregated texts containing multiple or all user comments. In these cases, the implementation is similar. However, in some cases, there is an added value in meta-features based on different statistical calculations of lower-level feature values. The following section provides a detailed description and preliminary analysis of the extracted features.
%\todo{mention closed and open vocab methods}
\subsection{Closed vocabulary}
We utilized LIWC \cite{pennebaker2015development}, a text analysis software commonly used in text-based personality research to extract features based on various linguistic categories. The LIWC features are the results of the most commonly used implementation of closed-vocabulary methods in personality research on text and can aid in directly comparing findings from different studies. Using LIWC, we extracted a total of 93 features, which encompassed different aspects of language usage such as part-of-speech (e.g., pronouns, articles), topical preferences (e.g., bodily functions, family), and psychological categories (e.g., emotions, cognitive processes). 

In addition to LIWC, we incorporate several psycholinguistic word lists in our analysis. These included measures such as perceived happiness, affective norms (valence, arousal, and dominance), imageability, and sensory experience, as described in \cite{perspara17nlpcss}, along with two lists of word meaningfulness ratings from the MRC Psycholinguistic Database \cite{coltheart1981mrc}. We calculated the average ratings for each user based on the words in these dictionaries, resulting in 26 features, which we refer to as PSYCH.

Additionally, we computed the type-token ratio as a proxy for lexical variation, the ratio of English comments, and the ratio of British English words to American English words for each user, which can indicate the user's origin.

\subsection{Open vocabulary}
The linguistic features based on open vocabulary aim to capture both content and style. We employed character n-grams (lengths of 2-3) and word n-grams (lengths of 1-3), which were weighted by term frequency (tf) and term frequency-inverse document frequency (tf-idf). The reasoning for utilizing both character and word-level features is that character-based features more effectively capture individual style. Conversely, word-based features, when taken out of context, mainly convey information about content. Longer n-grams can also reveal user preferences, values, self-descriptions, and attitudes (e.g., \emph{I often smile} or \emph{I am well-spoken}). We leverage weighted n-gram features since they reduce the scores of frequently used words. Nevertheless, these n-grams can carry a significant personality-relevant signal \cite{pennebaker2015development}, so we integrate them with LIWC-extracted features that maintain frequency information. To minimize the number of word n-grams, we apply Porter's stemmer.
\subsection{Activity-based features}
We collected user activity data that reflect user behavior from comment and post metadata, dividing the extracted features into global and post features. The global features encompass the total number of comments the user makes across all subreddits, including MBTI-related subreddits, and the number of subreddits in which the user has commented. The post features include the user's overall post score, the number of posts made in \emph{over 18} subreddits, the number of self-posts (posts that link to other Reddit posts), and the number of gilded posts (posts that other users have awarded money).

We also extracted topical affinity features to further characterize the user's activity. We counted the user's comments across subreddits and represented them as a single vector, together with the entropy of the corresponding distribution. We employed LDA models \cite{blei2003latent} with 50 and 100 topics to derive topic distributions from the user's comments, manually grouping the top 200 subreddits into 35 semantic categories to encode them as 50-, 100-, and 35-dimensional vectors.

Additionally, we considered the temporal aspect of user activity, which may be necessary for personality type prediction. Therefore, we included time intervals between comment timestamps, such as mean, median, and maximum delay. We also encoded daily, weekly, and monthly comment distributions as vectors of the corresponding lengths.

\section{Feature analysis}

\subsection{Feature relevance}

We estimate the relevance of each feature for each MBTI dimension using a
two-sided t-test: we define feature relevance as inversely proportional to the p-value under the null hypothesis of no difference in feature values for the two classes.
Table~\ref{tab:feature_top30} shows the proportion of features from each
group of features in the top 30 most relevant features for each dimension of MBTI. For instance, tf-weighted character n-grams (char\_tf) account for about 29\% of top-30 most relevant features in the extravert-introvert (E/I) dimension. The main observation is that different features are relevant for different dimensions. Generally, tf-idf weighted character n-grams are the most relevant features for all dimensions except for E/I, for which tf-idf weighted word n-grams are most relevant. However, while LIWC, PSYCH, and LDA100 represent 48\% of the top 30 most relevant features for the T/F dimension, they have no relevance for the S/N dimension. Post features seem to be relevant only for S/N and J/P dimensions. 

\begin{table}[t]
	{\small
		\begin{center}
        			\caption {Percentage of each feature group in top-30 relevant features for each dimension}
			\begin{tabular}{lllll}
				\toprule
				Feature group  & E/I   & S/N   & T/F   & J/P   \\
				\midrule
				char\_tf & 29.03 & 45.16 & 35.48 & 51.61 \\
				word\_tf & 35.48 & 25.81 & 12.9  & 32.26 \\
				liwc         & 19.35 & 0.0   & 25.81 & 9.68  \\
				lda100       & 6.45  & 0.0   & 9.68  & 3.23  \\
				psy          & 3.23  & 0.0   & 12.9  & 0.0   \\
				word        & 3.23  & 9.68  & 0.0   & 0.0   \\
				char        & 0.0   & 12.9  & 0.0   & 0.0   \\
				posts        & 0.0   & 6.45  & 0.0   & 3.23  \\
				\bottomrule
			\end{tabular}
			\label{tab:feature_top30}
		\end{center}
	}
\end{table}

Table~\ref{tab:feature_perc_used} provides an additional perspective on the significance of features. Specifically, it displays the percentage of highly significant features (p-value \textless\ 0.001) from each feature group utilized for each dimension. The global, PSYCH, and LIWC features are used considerably (>50\%) for one or more dimensions. Unsurprisingly, the PSYCH and LIWC features are relevant, since they were designed to model psycholinguistic processes. Unlike the post features, these features appear to be most indicative of the T/F dimension while being the least relevant for the S/N dimension.

\begin{table}[t]
\centering
\small
\caption{Percentage of highly relevant features (\(p<0.001\)) in the total number of features per feature group and dimension.}
\begin{minipage}[t]{0.48\linewidth}
\centering
\begin{tabular}{lllll}
\toprule
\textbf{Feature group} & \textbf{E/I} & \textbf{S/N} & \textbf{T/F} & \textbf{J/P} \\
\midrule
global        & 33.33 & 33.33 & 100.00 & 66.67 \\
psy           & 25.00 & 41.67 & 70.83  & 41.67 \\
liwc          & 40.86 & 29.03 & 62.37  & 39.78 \\
day\_of\_week & 0.00  & 0.00  & 28.57  & 100.00 \\
word\_an\_tf  & 28.22 & 32.07 & 38.17  & 27.30 \\
char\_an\_tf  & 19.28 & 27.06 & 36.26  & 21.47 \\
word\_an      & 7.40  & 19.58 & 27.28  & 24.72 \\
char\_an      & 4.45  & 14.40 & 30.30  & 8.82  \\
meaning       & 0.00  & 0.00  & 50.00  & 0.00  \\
sub           & 0.04   & 0.48  & 0.14  & 0.00  \\
\bottomrule
\end{tabular}
\end{minipage}%
\hfill
\begin{minipage}[t]{0.48\linewidth}
\centering
\begin{tabular}{lllll}
\toprule
\textbf{Feature group} & \textbf{E/I} & \textbf{S/N} & \textbf{T/F} & \textbf{J/P} \\
\midrule
lda100        & 9.00   & 12.00 & 15.00 & 9.00  \\
posts         & 5.00   & 20.00 & 5.00  & 10.00 \\
char          & 0.12   & 0.88  & 28.99 & 0.24  \\
month         & 0.00   & 25.00 & 0.00  & 0.00  \\
word          & 0.16   & 1.23  & 21.67 & 1.12  \\
time\_diffs   & 0.00   & 16.67 & 0.00  & 0.00  \\
subcat        & 0.00   & 2.86  & 8.57  & 0.00  \\
lda50         & 0.00   & 6.00  & 4.00  & 0.00  \\
hour          & 0.00   & 0.00  & 0.00  & 4.17  \\
--            & --    & --    & -- & -- \\
\bottomrule
\end{tabular}
\end{minipage}
\label{tab:feature_perc_used}
\end{table}

\begin{figure}[t]
	\centering
	\vspace*{-1em}
	\begin{subfigure}[b]{0.5\textwidth}
		\centering
		\includegraphics[width=\textwidth]{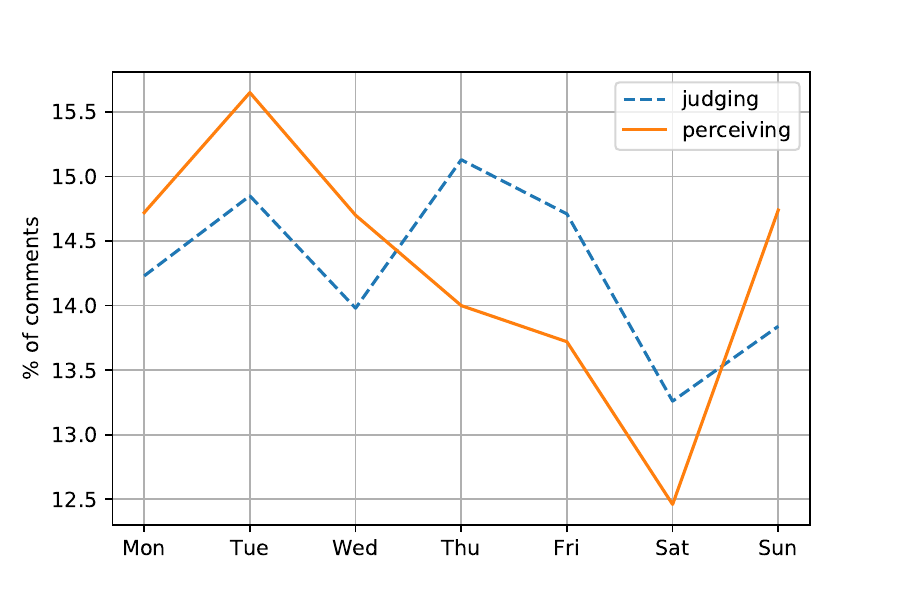}
		\vspace*{-2em}
		\caption{J/P distribution across days of week}
		\label{sf:day}
	\end{subfigure}\\
	\begin{subfigure}[b]{0.5\textwidth}
		\centering
		\includegraphics[width=\textwidth]{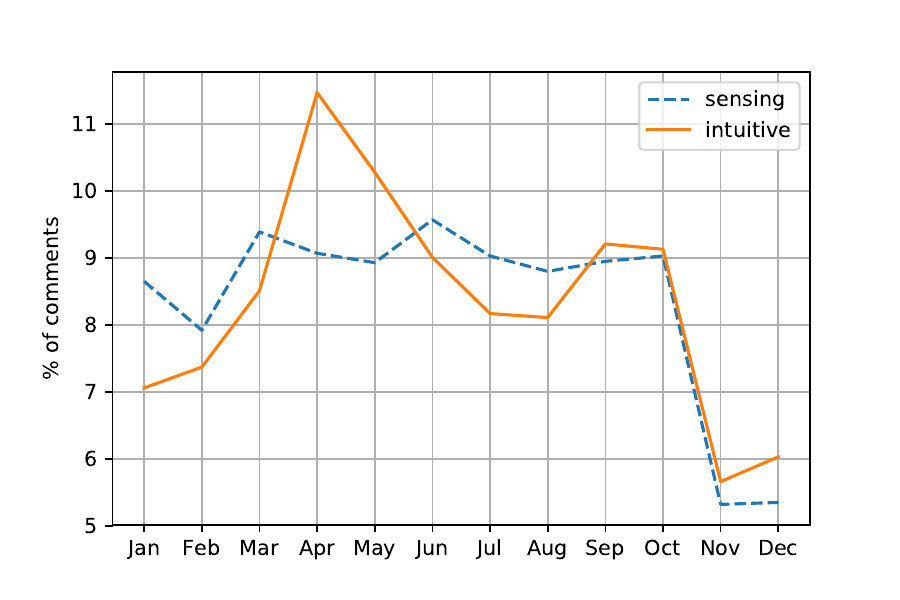}
		\vspace*{-2em}
		\caption{S/N distribution across months}
		\label{sf:month}
	\end{subfigure}
	\caption{Temporal distribution of comments}
	\label{fig:temporal}
\end{figure}

\subsection{Temporal features}

We examined the day-of-week and monthly distribution of comments to analyze the relationship between posting habits and the dimensions of the MBTI. Although day-of-week distribution was a good predictor for the T/F and J/P dimensions, posting time differences only seemed relevant for the S/N dimension. A day-of-week proportion of 100\% for the J/P dimension indicates that all points in the distribution indicate that particular dimension. Conversely, a monthly distribution proportion of 25\% suggests that only four months in a year are relevant for the S/N dimension. We provide more insight in Fig.~\ref{sf:day}, which displays the distribution of comments on days of the week for the J/P and S/N dimensions. We observed that perceiving types comment more on Tuesdays and Sundays, while judging types comment more on other days. Additionally, intuitive types are more active during April and May, whereas sensing types prefer to comment during January and July.

\subsection{Word usage} Using specific words or word classes is well correlated with personality traits. The face validity of these linguistic features reinforces their utility in personality analysis: Language patterns not only reflect statistically robust relationships but also make intuitive sense given what we know about the traits in question. Conversely, if word usage patterns were found to diverge from established personality concepts, it would signal a lower degree of face validity, thus decreasing our confidence in these features as reliable predictors in personality models. 
In our datasets, we found that Extraversion is characterized by the use of social- and family-related words \cite{schwartz2013personality} and the use of exclamation marks. This is consistent with the most relevant word features for the E/I dimension in our dataset: \textit{Friend}, \textit{Social}, \textit{comm\_mbti}, \textit{only}, \textit{i'm an extrovert}, \textit{fri}, \textit{at least}, \textit{drivers}, \textit{Affiliation}, \textit{Exclam}, \textit{origin}, \textit{!!} (word classes from LIWC and PSYCH are shown capitalized). The most relevant words for the S/N dimension are also somewhat expected: \textit{Is\_self\_mean}, \textit{Is\_self\_median},
\textit{–}, \textit{i}, \textit{’}, \textit{is a}, \textit{my\_}, \textit{it},
\textit{“a}, \textit{Avg\_img}, \textit{my}, \textit{\_he}, \textit{cliché},
\textit{Sixltr}, \textit{exist}. By definition, sensing types are more
concrete, while intuitive types are more abstract, which seems to be reflected in the imageability feature (e.g., \textit{Avg\_img}). Intuitive users tend to use rare (\textit{e.g., cliché}), more complex, and longer words (as indicated, e.g., by the \emph{Sixltr} feature: words with more than six characters).
Sensing types also seem to share posts with content they found outside of Reddit more than intuitive users (e.g., \textit{Is\_self} features).  Feelers tend to use more words about love, feelings, and emotions. They also use more social and affectionate words, as well as pronouns and exclamations, as evidenced by
the most relevant words for the T/F dimension: \textit{love}, \textit{Feel},
\textit{Posemo}, \textit{valence}, \textit{Emotion}, \textit{happy},
\textit{i}, \textit{polarity}, \textit{!}, \textit{i love}, \textit{Ppron},
\textit{SOCIAL}, \textit{Exclaim}, \textit{Affect}, \textit{Pronoun},
\textit{\_so}, \textit{e!}. The most relevant words for J/P also seem to
reflect the common stereotypes, such as that judgers are more plan, work, and
family oriented: \textit{Work}, \textit{husband}, \textit{Home}, \textit{help},
\textit{for}, \textit{plan}, \textit{sit}, \textit{hit}, \textit{joke},
\textit{fo}.

%\todo{something to add more}

\chapter{Personality assessment experiments}
\label{ch:part_two_experiments}

In this chapter, we describe ATBPA experiments designed to better understand the potential of newly developed datasets and the expected models' performance in predicting type- and trait-based personality and demographic labels. First, we developed baseline prediction models for MBTI and Big Five based on features described in the previous chapter. Next, we describe additional experiments motivated by both practical and theoretical implications. The first experiment focuses on domain adaptation and illustrates how models trained on labels from the MBTI model can predict labels from the established Big Five model. This experiment capitalizes on the abundance of data for MBTI and utilizes correlations between the traits of the various models and their expression in language. The second experiment demonstrates how the comprehensive psychodemographic profile can help detect biases in gender classification.
The issue of bias is of utmost importance in the case of Reddit, since Reddit data is used to train various NLP models. Our research demonstrates that a gender classifier trained on a large Reddit dataset struggles to predict gender for users with specific combinations of personality traits more frequently than for other users. The third experiment presents the practical applications of \pandora in the social sciences. Building upon existing psychological theories, we perform a confirmatory and exploratory analysis between the propensity for philosophy and various psychodemographic variables.

\section{Prediction models}
\label{sec:prediction}

After verifying that the computed features provide face validity (see subsection~\ref{subsection:background_validity_reliability}), the next step was to assess their predictive validity. We used traditional classification and regression models previously employed for personality prediction to establish benchmark performance and simplify the experimental setup. This decision was particularly driven by the challenge of aggregating individual comments into a user-level representation, given a large number of comments per user, with some reaching 100,000 comments in the dataset. To overcome this challenge, we opted to aggregate user comments into a single text and generate predictions at the user level. We repeat this approach for two developed datasets that we made publicly available: MBTI9k and \pandora. 
However, we did not use all the context we could have to better contextualize personality cues. Consistent with standard practice, we treated the MBTI personality prediction task as four distinct binary classification problems, one for each MBTI dimension. This design choice was driven by the assumption that these dimensions are orthogonal. However, this is an oversimplification and does not hold in reality.
In addition, we explored a 16-class classification problem that involved predicting the MBTI type by combining the predictions for the four individual dimensions.

\subsection{Experimental setup}

For MBTI classification using the MBTI9k dataset, we explored the performance of three commonly used classifiers: a support vector machine (SVM), $\ell_2$-regularized logistic regression (LR), and a three-layer multilayer perceptron (MLP). We used nested stratified cross-validation to select the best model with five folds in the outer loop and 10 (for LR) or 5 (for SVM) folds in the inner loop. The macro F1-score was used as the evaluation criterion during model selection. To assess the usefulness of the different features, we (1) trained all models with features selected using the t-test and (2) trained the LR model separately with each feature group. To prevent data leakage, feature selection and standard scaling were applied separately in the training set for each cross-validation fold, and the number of features was optimized. We accounted for class imbalance by using class weighting. A majority class classifier (MCC) was used as a baseline. We employed the Scikit-learn \cite{pedregosa2011scikit} implementation for all models.

We redid the experiments for MBTI using \pandora dataset along with the models for Big5 and demographic variables prediction. 
We used several sets of features, including: (1) \textbf{N-grams}: Tf-Idf weighted 1--3 word n-grams and 2--5 character n-grams, (2) \textbf{Stylistic}: Number of words, characters and syllables, monosyllabic/polysyllable words, long words, unique words, and all readability metrics implemented in Textacy, (3) \textbf{Dictionaries}: words mapped to Tf-Idf categories from LIWC, Empath, and NRC Emotion Lexicon dictionaries, (4) \textbf{Gender}: Predictions of the gender classifier from \S\ref{section:genderpred}, (5) \textbf{Subreddit distributions}: a matrix where each row is a distribution of the number of posts in all subreddits for a given user, reduced to 50 features per user using PCA, (6) \textbf{Subreddit other}: Number of downs, score, gilded, ups and controversiality scores for a comment, (7) \textbf{Named entities}: the number of named entities per comment extracted using Spacy, (8) \textbf{Part-of-speech}: Counts for each part-of-speech, and (9) \textbf{Predictions} (for Big Five trait predictions only): MBTI predictions made by a classifier based on withheld data. Features (2), (4), and (6--9) were computed at the individual comment level and aggregated to minimum, maximum, mean, standard deviation, and median values for each user, except for the subreddit PCA features, which were constructed at the user level, and the N-gram and dictionary features, which were computed by considering the concatenation of comments for each user as a single text and applying a tf-idf weighting.

Our study involved constructing six regression models to predict age and Big Five personality traits and eight classification models to predict four dimensions of MBTI, gender, region. To build these models, we experimented with various algorithms, including linear/logistic regression (LR) from the Sklearn library \cite{pedregosa2011scikit}, as well as deep learning models, particularly neural networks (NN).
%\todo{explain the difficulties with DL; the representation masked language model BERT - and why it wasn't fine-tuned -look at chat-gpt; wanted to preserve whole comment, the comments are lengthier than the BERT context; wanted to capture both the local and global context; reuse knowledge that ngrams work for this kind of problem}
We trained a separate neural network (NN) model for each task that represented a user as a matrix, each row representing one of his or her comments. To encode the comments, we used 1024-dimensional vectors generated by BERT \cite{devlin-etal-2019-bert}, fed into convolutional layers, max-pooling, and multiple fully connected (FC) layers. Due to limited computational resources, we used only the last 100 comments per user. The models consisted of a convolutional layer, a max-pooling layer, and several FC layers. The convolution kernels were as wide as the representation of BERT and slid vertically across the matrix to combine information from multiple comments. We experimented with different kernel sizes (2 to 6) and the number of kernels ($M$ = 4 to 6). The convolutional layer outputs were then decomposed into a fixed number of $K$ slices, and max pooled, resulting in $M$ vectors of length $K$ per user, one for each kernel. These were passed to multiple FC layers with Leaky ReLU activations, and regularization was applied only to the FC layers (L2 norm and dropout).

We evaluated the models using 5-fold cross-validation with a separate stratified split for each target. We used regression F-tests to select the top $K$ features, and the hyperparameters of the models and $K$ are optimized via grid search on held-out data in each fold.
We evaluated the models using the Pearson correlation coefficient and macro-f1, respectively,

\subsection{Results}

\subsubsection{Models based on the MBTI9k dataset}
\paragraph{MBTI per-dimension prediction.} 

The macro F1-scores for each MBTI dimension's prediction results are displayed in Table~\ref{tab:results}, averaged across the five folds. The achieved results are well above the baseline despite using relatively simple models. Models that combine all features, such as LR\_all and MLP\_all, perform the best across all dimensions.

In terms of individual dimensions, MLP\_all is the best model for the E/I dimension, with only a slightly higher score than the LR word n-gram model. Character n-grams, LIWC, and PSYCH also predicted the E/I dimension. For the S/N dimension, MLP\_all once again outperforms other models, while word n-gram features demonstrate good performance. Models based on topical and user-activity features did not achieve results above the baseline.

The T/F dimension had the lowest results, consistent with the previous findings \cite{capraro2002myers}. N-gram-based features perform slightly better than dictionary-based (LIWC, PSYCH) and topic-based (LDA) features, but the differences in model scores are insignificant. Finally, for the J/P dimension, the best-performing model is LR\_all, which outperforms all models that use a single feature group.

\begin{table}{}
	{\small
		\begin{center}
        			\caption{Macro F1-scores for per-dimension prediction and accuracy of type-level prediction for models with all features, LR models with a single feature group, and the MCC baseline}
			\begin{tabular}{lllllr}
				\toprule
& \multicolumn{4}{c}{Dimensions}\\
\cmidrule(lr){2-5}
				Model            & E/I   & S/N   & T/F  & J/P & Type  \\
				\midrule
				LR all          & 81.6  & 77.0    & \textbf{67.2} & \textbf{74.8} & 40.8 \\
				MLP all         & \textbf{82.8}  & \textbf{79.2}  & 64.4 & 74.0   & \textbf{41.7} \\
				SVM all         & 79.6  & 75.6  & 64.8 & 72.6 & 37.0 \\
				\midrule
				LR w\_ng & 81.0    & 73.6  & 66.4 & 71.8 & 38.0 \\
				LR chr\_ng & 62.2  & 64.0    & 66.4 & 65.8 & 26.5 \\
				LR liwc         & 55.0    & 49.8  & 65.0   & 57.4 & 14.2 \\
				LR psych        & 52.0    & 48.2  & 64.0   & 57.0 & 12.5   \\
				LR lda100       & 50.0    & 48.2  & 62.4 & 56.2 & 13.9 \\
				LR posts        & 49.4  & 53.2  & 48.0   & 51.8 & 9.5 \\
				LR subtf        & 49.6  & 49.6  & 50.4 & 50.2 & 13.2 \\
				\midrule
				MCC              & 50.04 & 50.04 & 50.0   & 50.02 & 25.2\\
				\bottomrule
			\end{tabular}

			\label{tab:results}
		\end{center}
	}
\end{table}

\begin{figure}[t]
	\begin{center}
		\vspace*{-1.5em}
		\hspace*{-0.5em}\includegraphics[scale=0.85]{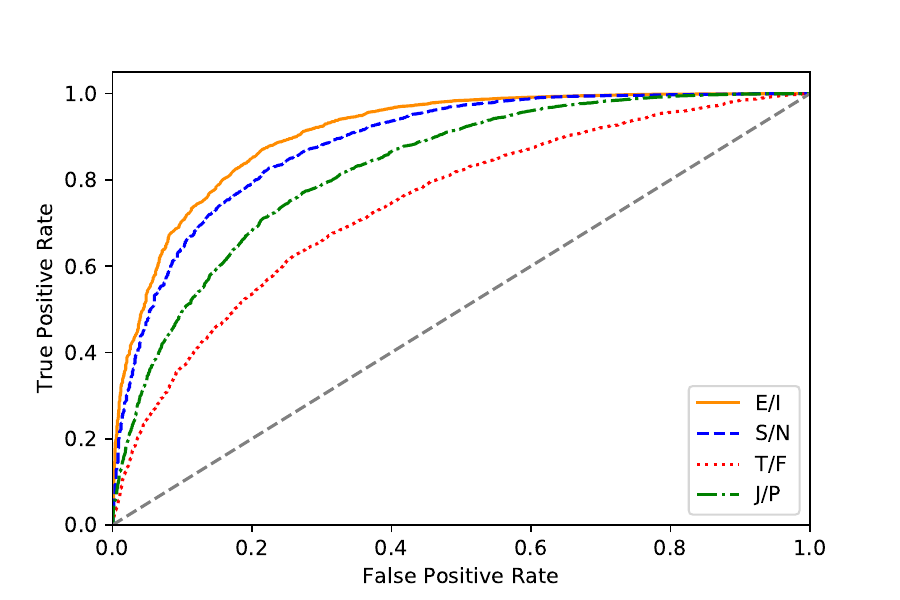}
		
		\caption{ROC curves for the LR\_all model}
		\label{fig:roc}
	\end{center}
\end{figure}

As personality traits exist on a continuous scale along each dimension, it is reasonable to assess type prediction as a confidence-rated classification task using ROC curves. The results are depicted in Figure~\ref{fig:roc}. The ROC curve illustrates the relationship between the true positive rate (recall) and the false positive rate (fall-out), both increasing as the classification threshold increases. For example, the ROC curve for the T/F dimensions informs us that we can identify approximately 70\% of T cases with a fall-out of approximately 40\%.

\paragraph{MBTI type-level prediction.} 

\begin{figure}
	\centering
	\vspace*{-1.5em}
	\hspace*{0.75em}\includegraphics[scale=0.85]{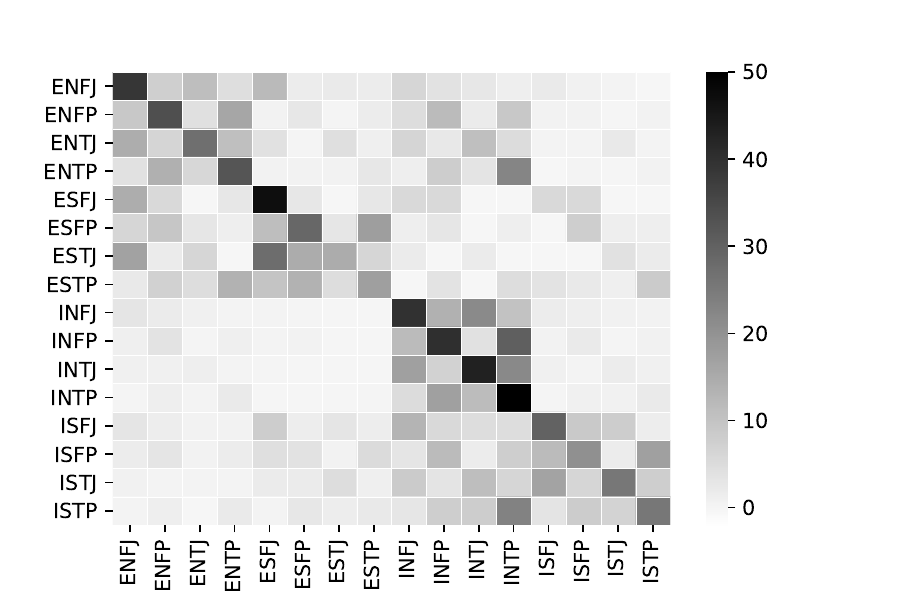}
		\caption
		{\small{Heatmap of the type prediction confusion matrix}}
		\label{fig:heatmap}
\end{figure}

To predict the MBTI type, we combined the outputs of the binary models for each individual dimension, and the accuracy results are presented in the last column of Table~\ref{tab:results}. Among all models, MLP\_all achieved the highest accuracy of 42\%, which is significantly better than the baseline accuracy of 25\%. Table~\ref{tab:mismatch} provides more detailed information about incorrect predictions made by the LR\_all model, including the number of mismatched dimensions. The results show that in 82\% of the cases, the model predicts either the correct type or makes an error on only one dimension, and in more than 97\% of cases, the model correctly predicts two or more dimensions. The heatmap in Figure~\ref{fig:heatmap} presents the likely mismatches based on the confusion matrix of the LR\_all model. We observed that similar types in the MBTI theory tend to get grouped, such as introverted intuitive types, including INTP and INTJ, which are often difficult to distinguish. However, the model successfully captured these nuances by showing that INTJ is more similar to INFJ, while INTP is more similar to INFP.

\begin{table}[t]
\centering
	{\small
    		\caption{The number and percentage of mismatched dimensions between predicted and actual types}
            
\label{tab:mismatch}
	\begin{tabular}{rrrrrr}
		\toprule
& \multicolumn{5}{c}{\#\,mismatches}\\
\cmidrule(lr){2-6}
		 & 0     & 1     & 2     & 3    & 4    \\
		\midrule
		Count & 3757 & 3715 & 1384 & 240 & 15 \\
		\%    & 40.83 & 40.77 & 15.61 & 2.63 & 0.16 \\
		\bottomrule

	\end{tabular}}

\end{table}

%results
\subsubsection{Models based on the \pandora dataset}
\label{sec:part_two_pandora_prediction}
\begin{figure}
    \centering
    %\vspace{-1em}
    %\hspace*{-0.5cm}
    \includegraphics[scale=0.7]{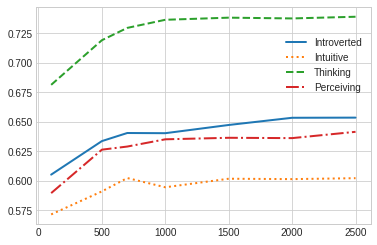}
    %\vspace*{-1.5em}
    \caption{Learning curves of logistic regression for MBTI trait prediction}
    \label{fig:lcmbti}
\end{figure}

\begin{figure}
    \centering
    %\vspace{-1em}
    %\hspace*{-0.5cm}
    \includegraphics[scale=0.7]{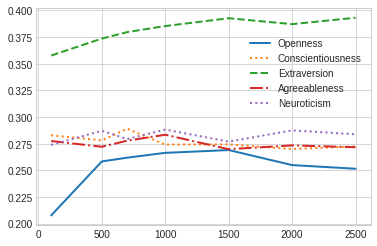}
    %\vspace*{-1.5em}
    \caption{Learning curves of logistic regression for Big Five trait prediction}
    \label{fig:lcbig5}
\end{figure}

We evaluated the models using the Pearson correlation coefficient and macro-f1, respectively, and the results are shown in Table~\ref{tbl:baselineres}. It can be observed that LR performs best when only n-gram features are used.
We also tried ExtraTreesRegressor, which benefits from other features but performs comparably, so we omit those results.
The prediction of the Big Five traits, unlike other tasks, showed significant improvement when the MBTI predictions were included as features, as reported in Section~\ref{sec:mbtibig5} and Table~\ref{tbl:mbti_big5_corr_valid}. In addition, increasing the number of comments used for training from the last 100 to 1000 increased the score by up to 5 points. We experimented with different numbers of comments per user and found that using more comments generally led to better results for all tasks.
Figures~\ref{fig:lcmbti} and \ref{fig:lcbig5} show the learning curves for a logistic regression model with unigram features: the x axis is the number of comments and the y axis is the macro score F1 of the model. Performance levels off at around 1000 comments and shows little significant change as the number of comments used for training increases above this value.

\begin{table}[]

  \centering
{\small
\setlength{\tabcolsep}{4.5pt}
\caption{\label{tbl:baselineres}Prediction results for the different traits for LR and NN models. For the LR model, we show the results for different feature combinations, including N-grams (N), MBTI predictions (P), and all other features (O). Scores are macro-averaged F1 scores for classification tasks and Pearson correlation coefficients for regression tasks.
The best results are shown in bold.}
}
\begin{tabular}{@{}lcccccc@{}}
\toprule 
  & \multicolumn{5}{c}{LR} & \\
  \cmidrule(lr){2-6}
  & NO & N & O & NOP & NP & NN \\
\midrule
\multicolumn{7}{c}{Classification (Macro-averaged F1 score)} \\
\midrule
Introverted &        .649 &        \textbf{.654} &        .559 & -- & -- & .546\\
Intuitive   &        .599 &        \textbf{.606} &        .518 & -- & -- & .528\\
Thinking     &       .730 &        \textbf{.739} &        .678 & -- & -- & .634\\
Perceiving    &      .626 &        \textbf{.642} &        .586 & -- & -- & .566\\
%Enneagram &          .155 &        \textbf{.251} &        .145 & -- & -- & .143\\
Gender         &     .889 &        \textbf{.904} &        .825 & -- & -- & .843\\
Region          &    .206 &        \textbf{.626} &        .144 & -- & -- & .478\\
\midrule
\multicolumn{7}{c}{Regression (Pearson correlation coefficient)} \\
\midrule
Agreeableness  &     .181 &        .232 &        .085 &  .237 &         \textbf{.270} & .210\\ 
Openness        &         .235 &        \textbf{.265} &        .180 &    .235 &         .250 & .159\\
Conscientiousness &        .194 &        .162 &        .093 &   .245 &         \textbf{.273} & .120\\  
Neuroticism        &       .194 &        .244 &        .138 &   .266 &         \textbf{.283} & .149\\ 
Extraversion &              .271 &        .327 &        .058 &  .286 &         \textbf{.387} & .167\\ 
Age &        .704 &        \textbf{.750} &        .469 & -- & -- & .396\\
\bottomrule
\end{tabular}

\end{table}

%\section{Learning Curves for the Logistic Regression Models}

\section{Predicting Big Five with MBTI}

%\todo{break down into motivation, experimental setup, results}
%\section{Predicting Big Five with MBTI/Enneagram}
\label{sec:mbtibig5}
%\todo{move results in separate subsection}
MBTI personality model is significantly more prevalent among social media users than the Big Five model. Consequently, acquiring MBTI labels (Section~\ref{section:flair_based_labels}) and developing effective predictive models using supervised machine learning are relatively straightforward. However, we acknowledge that the validity of the MBTI has faced significant criticism (see Section~\ref{section:background_personality_models}), and they are seldom employed in psychological research.

In this experiment, our objective is to explore the possibility of combining the MBTI that are more common in \pandora to predict the Big Five traits from the text. We hypothesize that moderate to strong correlations between the different personality models (Table~\ref{tbl:mbti_big5_corr_valid}) and the presence of a significant number of users with multiple labels (Table~\ref{tbl:intersection}) may compensate for the questionable psychological validity of the MBTI.

In this experiment, we approached the task as a domain adaptation problem by mapping MBTI to Big Five labels using a simple domain adaptation approach from \cite{daume-iii-2007-frustratingly}. We first partitioned \pandora into three subsets: comments of users for which we have both MBTI and Big Five labels (M+B+, n=382), comments of users for which we have the MBTI but no Big Five labels (\mbox{M+B-}, n=8,691), and comments of users for which we have the Big Five but no MBTI labels (M-B+, n=1,588).
We then proceeded in three steps.  In the first step, we trained on M+B- four text-based MBTI classifiers, one for each MBTI dimension (logistic regression, optimized with 5-fold CV, using 7000 filter-selected, Tf-Idf-weighed 1--5 word and character n-grams as features).  We then applied these classifiers to a subset of \pandora users who have MBTI and Big Five labels. We used MBTI scores as features to train five regression models, one for each Big Five trait. Finally, we applied both classifiers to the subset of \pandora users with Big Five labels but without MBTI labels, which serves as a target set for domain adaptation. We used MBTI classifiers to obtain scores for the four MBTI dimensions and then fed them with Big Five models to obtain predictions for the five traits. 

\subsection{Results}
We combined the outputs of four classifiers and achieved a type-level MBTI prediction accuracy of 45\%. Table~\ref{tbl:mbti_big5_pred_corr_valid} presents the correlations between the gold MBTI and Big Five labels and the predicted MBTI labels. As expected, the overall correlations are lower than those between the gold labels (Table~\ref{tbl:mbti_big5_corr_valid}). The main difference is that extraversion moderately correlates with the predicted intuitive dimension of the MBTI. The correlations displayed in Table~\ref{tbl:enne_mbti_big5_corr} clearly indicate that predictions based on MBTI labels can assist in predicting the Big Five traits.

Furthermore, the results support the use of regression models, as the predicted Big Five traits correlate more strongly with self-reported Big Five traits than the predicted MBTI dimensions, except for Conscientiousness, which significantly correlates with the Perceiving/Judging MBTI dimension. For instance, predicted openness predicts openness more effectively than the intuitive dimension. We observe a similar effect for agreeableness, which is negatively correlated with thinking (and positively correlated with feeling) but shows a positive correlation with intuitive preference, as anticipated. Additional experiments indicate that these predictions provide valuable features in predictive models.

%Unexpectedly, the Enneagram's predictions made with a low-performing model (f1\_macro: 21\%) demonstrate strong correlations with various traits. The value of the Enneagram's predictions is most apparent in neuroticism, which is not included in the MBTI model.
%
%The datasets used in these experiments are significantly smaller: we train the Enneagram classifier with 730 users and test it with 64 users. We find that openness does not correlate with any of the Enneagram types. Types 1 and 2 exhibit higher conscientiousness, unlike Type 7, which is more extroverted than Type 5. Types 2 and 4 are agreeable, while Type 8 is not, consistent with its description as powerful, dominant, confident, determined, strong-willed, and confrontational. Surprisingly, we observe a strong positive correlation between neuroticism and Types 4 and 6, along with a negative correlation with Type 7, which is not included in the MBTI model.
%
\begin{table}[t!]
\centering
\small
\caption{\label{tbl:mbti_big5_pred_corr_valid}Correlations between predicted MBTI (rows) and self-reported ``Gold'' Big Five (columns) among users who reported both. 
Significant correlations (\(p<.05\)) are in bold.}
\begin{tabular}{lrrrrrrrrr}
\toprule
& \multicolumn{9}{c}{\textbf{Gold}} \\
\cmidrule(lr){2-10}
\textbf{Predicted} & O & C & E & A & N & I/E & N/S & T/F & P/J \\
\midrule
I/E & --.094 & --.003 & \textbf{--.516} & .064 & .076 & \textbf{.513} & .046 & --.061 & \textbf{--.108} \\
N/S & \textbf{.251} & .033 & \textbf{.118} & .068 & --.026 & --.096 & \textbf{.411} & --.036 & --.033 \\
T/F & --.087 & .085 & --.142 & \textbf{--.406} & \textbf{--.234} & .023 & --.043 & \textbf{.627} & .083 \\
P/J & .088 & \textbf{--.419} & --.002 & .003 & .007 & --.066 & .032 & \textbf{.141} & \textbf{.587} \\
\bottomrule
\end{tabular}
\end{table}

% 
% \begin{table}[t!]
%
%  \centering
%{\small
%\begin{tabular}{lrrrr}
%\toprule 
%  & \multicolumn{4}{c}{Predicted} \\
%  \cmidrule(){2-5}
% %\cmidrule(c){2-5} 
%%\cmidrule(l){1}\cmidrule(rrrr){2-5}
%%Gold & & & Predicted & \\
%Gold  & I/E & N/S & T/F & P/J \\
%\midrule
%O   & --.094 &\textbf{.251}  & --.087 & .088  \\
%C & --.003 & .033  & .085  & \textbf{--.419} \\
%E      &\textbf{--.516} & \textbf{.118}  & --.142 & --.002 \\
%A     & .064  & .068  & \textbf{--.406} & .003  \\
%N       & .076  & --.026 & \textbf{--.234} & .007  \\
%\midrule
%I/E       & \textbf{.513}  & --.096 & .023  & --.066 \\
%N/S      & .046  & \textbf{.411}  & --.043 & .032  \\
%T/F          & --.061 & --.036 & \textbf{.627}  & \textbf{.141}  \\
%P/J        & \textbf{--.108} & --.033 & .083  & \textbf{.587} \\
%\bottomrule
%\end{tabular}\caption{\label{tbl:mbti_big5_pred_corr_valid} Correlations between predicted MBTI and Big Five with gold Big Five traits among users who reported both MBTI and Big Five. Significant correlations (p\textless{.05}) are shown in bold.}
%}
%\end{table}

% \usepackage{graphicx}
\begin{table}[]
\centering
{\small
\caption{\label{tbl:enne_mbti_big5_corr} Correlations between predicted MBTI, and Big Five with gold Big Five traits among users who reported only Big Five traits. Significant correlations (p\textless{.05}) are shown in bold.}
}
\begin{tabular}{crrrrr}
\toprule
Predicted & O & C & E & A & N \\
\midrule
I/E                       & \textbf{--.082}   & .039             & \textbf{--.262}       & --.003        & --.002      \\
N/S                       & \textbf{.127}    & --.021            & .049        & \textbf{.060}         & .001       \\
T/F                       & --.001   & .038             & --.039       & \textbf{--.259}        & \textbf{--.172}      \\
P/J                       & .018    & \textbf{--0.41}            & .007        & .034         & .039       \\
\midrule
O                       & \textbf{.147}    & \textbf{--.082}            & \textbf{.212}       & \textbf{.145}       & \textbf{.070}    \\
C                       & --.007   & \textbf{.237}            & .013        & \textbf{--.112}        & \textbf{--.090}     \\
E                       & .098    & --.028            & \textbf{.272}      & .044         & .022       \\
A                       & .006   & \textbf{--.079}           & .023        & \textbf{.264}         &\textbf{.176}       \\
N                       & --.048   & --.025            & --.042       &\textbf{.231}         & \textbf{.162}       \\
%\midrule
%Enn. 1                          & .002    & .032             & -.028       & .047         & .025       \\
%Enn. 2                          & -.011   & \textbf{.108}             & .030        & \textbf{.135}         & \textbf{.046}       \\
%Enn. 3                          & \textbf{.085}   & .014             & \textbf{.071}        & -.064         & \textbf{-.069}      \\
%Enn. 4                          & -.041   & -.017            & -.033       & \textbf{.166}         & \textbf{.159}       \\
%Enn. 5                          & \textbf{.067}    & -.035            & \textbf{-.060}       & \textbf{-.121}        & \textbf{-.076}  \\
%Enn. 6                          & -.051   & .004             & -.035       & .046         & \textbf{.113}       \\
%Enn. 7                          & -.043   & -.019            & \textbf{.078}        &\textbf{ -.085}         & \textbf{-.088}     \\
%Enn. 8                          & .022    & -.044            & \textbf{.063}        &\textbf{ -.129}         &\textbf{ -.075}      \\
%Enn. 9                         & -.034   & -0.016            & \textbf{-.102}       & .041         & -.005     \\
\bottomrule
\end{tabular}
\end{table}

%\section{Confirmatory Study on Philosophy Propensity and Personality Traits}
\section{Confirmatory study on philosophy propensity and personality traits}
%\todo{break down into motivation, experimental setup, results}
%\begin{itemize}
%    \item motivate: we want to show how this dataset can be used by psychologists
%    \item describe method and findings
%\end{itemize}

%\paragraph{Language, location, and demographics.}

This section describes an experiment designed to demonstrate the potential of \pandora for social science research in both confirmatory studies and exploratory analyses. In confirmatory studies, researchers can replicate existing theories and findings using a dataset that differs from those typically used in the field. Additionally, they can assess different types of validity, including external and convergent validity (see Subsection~\ref{subsection:background_validity_reliability}). In exploratory studies, researchers can identify new relationships among psychodemographic variables manifested in online conversations. We present a use case combining both studies, focusing on the Reddit users' propensity to philosophy. We examined the relationship between this propensity (measured as a propensity for philosophical topics in online discussions) and openness to experience, cognitive processing (e.g., insight), and readability index, as previous research has suggested a positive correlation between these variables \cite{johnson2014measuring, dollinger1996on}. We hypothesized that we can confirm this relationship, as all four variables share a predisposition toward higher intellectual engagement. In addition, we conducted an exploratory analysis of emotional variables.

We conducted a hierarchical regression analysis to examine the relationship between the propensity of Reddit users for philosophical topics and several predictors, including demographics, personality traits, emotions, cognitive processing, and text readability. The propensity to philosophical topics was measured by the frequency of philosophical words using the Empath tool, while the Posemo, Negemo, and Insight features of the LIWC and the Flesh-Kincaid Grade Level Readability Score measured predictors. Emotion variables were included for exploratory analysis.

In the regression analysis, control variables (demographic data) were added in the first step, Big Five traits in the second step, emotion variables in the third step, insight trait as a cognitive inclination variable, and the F-K GL readability index in the last step. The sample included 430 Reddit users, 273 males and 157 females, with a mean age of 26.79 years (SD =7.954) and gold labels for gender, age, and Big-5 traits.

\subsection{Results}
We checked that multivariate normality and multicollinearity were satisfied, and homoscedasticity was satisfied after 14 outliers based on the standardized residuals were removed.
Table~\ref{tbl:philosophy} shows that the 11 predictors explain 41\% of the variance in the trait \textit{philosophy} feature. As expected, Openness to Experience, Readability Index, and Insight were significant and positive predictors of \textit{ philosophy} feature. Agreeableness was a significant negative predictor before adding emotion variables. The results show significant associations with the emotion variables. Negative emotions were positive predictors of the frequency of discussing philosophical topics. In contrast, positive emotions were a significant predictor until the final step when F-K GL was added to the model. This was due to a moderate correlation between posemo and F-K GL ($-0.40$). Finally, males had a higher frequency of philosophy-related words than females. In conclusion, the hypothesis is confirmed and the exploratory analysis provides interesting results that could stimulate further research.

\begin{table}
  \centering
{\small
    \setlength{\tabcolsep}{4.9pt}
    \caption{\label{tbl:philosophy} Hierarchical regression of propensity for philosophical topics (``philosophy'' feature from Empath) on gender, age, Big Five personality traits, Flesh-Kincaid Grade Level readability scores, positive and negative emotions features, and insight feature as predictors (n=430). The table shows regression coefficients and the goodness of fit measured by $R^2$, adjusted $R^2$, and $R^2$ change. Significant correlations: *p\textless{.05}, **p\textless{.01},  ***p\textless{.001}}.}
    \begin{tabular}{lrrrrr}
	    \toprule
	    &\multicolumn{5}{c}{Regression coefficients}\\
	    \cmidrule(lr){2-6}
Predictors      & Step 1	& Step 2 &	Step 3 &	Step 4 &	Step 5\\
    \midrule
Gender &	--.26\nospacetext{$^{**}$} &	--.24\nospacetext{$^{**}$} &	--.20\nospacetext{$^{**}$} &	--.19\nospacetext{$^{**}$} &	--.17\nospacetext{$^{**}$}\\
Age &	--.01 &	--.03 &	--.02 &	.00 &	.01\\
O &	-- &	.20\nospacetext{$^{**}$} &	.19\nospacetext{$^{**}$} &	.15\nospacetext{$^{**}$} &	 .10\nospacetext{$^{**}$}\\
C &	-- &	.01 & 	.05 & 	.08 & 	.07\\
E & 	-- &	.02 &	.03 & 	.04 & 	.04\\
A &	-- &	--.12\nospacetext{$^*$} &	--.05 &	--.05 &	--.06\\
N &	-- &	--.04 &	--.03 &	.01 &	.02\\
posemo &	-- &	-- &	.15\nospacetext{$^{**}$} &	.17\nospacetext{$^{**}$} &	.03\\
negemo &	-- &	-- &	.29\nospacetext{$^{**}$} &	.27\nospacetext{$^{**}$} &	.29\nospacetext{$^{**}$}\\
insight &	-- &	-- &	-- &	.36\nospacetext{$^{**}$} &	.27\nospacetext{$^{**}$}\\
F-K GL  &	-- &	-- &	-- &	-- &	.34\nospacetext{$^{**}$}\\
    \midrule
$R^{2}$ &	.07 & 	.12 & 	.22 &	.34 & 	.43\\
Adjusted $R^{2}$ &	.06 & 	.11 & 	.20 &	 .32 &	.41\\  
$R^{2}$ change &	.07\nospacetext{$^{**}$} &	 .06\nospacetext{$^{**}$} &	.10\nospacetext{$^{**}$} &	 .12\nospacetext{$^{**}$} &	.09\nospacetext{$^{**}$}\\
\bottomrule
\end{tabular}
\end{table}

%\section{Biases in Gender Classification Linked to Personality Traits}

\section{Biases in gender classification linked to personality traits}
\label{section:genderpred}
%\todo{break down into motivation, experimental setup, results}

%\begin{itemize}
%    \item we already know that there are many confounding variables: is personality one of them in another author profiling task?
%    \item we use gender classification task as these models work the best
%    \item describe a collection of the new dataset and training procedure and performance of the gender classifier
%    \item describe our findings
%    \item We train a gender classifier on a separate balanced dataset of 24,954 users with self-reported gender labels, and obtain 89.9% accuracy
%We then use this classifier on Pandora (n=3074) with an accuracy of 89.3%
%We find a statistically significant difference (z-test) between male and female users in the misclassified group of users: 8.1% of male users (142/1743) and 14.4% of female users (192/1331) 
%\end{itemize}

Gender classification is a crucial aspect of author profiling, especially on social media, and has attracted considerable attention in the NLP community \cite{bamman2014gender, sap2014developing, ciot2013gender}. This is in part because of its importance for advertising and recommender systems but also because age and gender must be considered in many computational sociolinguistic studies. In addition, gender is often a research topic on fairness and bias in NLP \cite{sun-etal-2019-mitigating}. Bias can arise from demographic and other imbalances in training data. To further validate \pandora dataset, we examine personality profiles as a possible source of bias and test whether a basic gender classification model trained on Reddit has biases that can be attributed to personality traits. This is a critical issue because Reddit is often used as a data source to train NLP models \citep{zhang-etall-2017-char, cheng-etal-2017-factored, henderson-etal-2019-repository, sekulic-strube-2019-adapting}.

To create a gender classifier, we obtained a unique Reddit dataset and automatically labeled it by gender using flairs with the strings ``/f/'' and ``/m/'' as female and male indicators, respectively. Although gender is not a binary construct, we restricted our analysis to users who specify a binary gender to obtain a more balanced dataset in bias analysis. This approach resulted in a precision of 98.5\% in \pandora. From the 34k users with these flair patterns, we selected a balanced dataset of 24,954 users and collected more than 30 million comments, excluding quoted text and comments with less than five words. We then summarized the user comments and divided the users into 80\% training and 20\% test groups.

\subsection{Results}
For classification, we used logistic regression with 500-dimensional SVD vectors from Tf-Idf word n-grams, achieving a test accuracy of 89.9\%. The accuracy of the classifier on 3,084 \pandora users with known gender is 89.3\%.
We first evaluate the accuracy of the classifier in \pandora to analyze bias. Table~\ref{tbl:fm_confusion} shows that the classifier did not correctly predict the gender of 8.1\% male (142/1743) and 14.4\% female (192/1331) users, the difference between the two groups being statistically significant (p\textless0.05 with a two proportion Z-test). To further examine this bias, we divide male and female users into correctly and incorrectly classified cases and test for statistically significant differences in psychodemographic variables between the groups. For binary variables, we use a two-proportion Z-test, whereas for continuous variables, we use the Kruskal-Wallis H test.
%\alert{The motivation behind this is to detect biases in the
%classifier with respect to such variables and differences in these biases for
%males and females.}
The table in Table~\ref{tbl:fm_predictors} shows the results of our bias analysis. Statistically significant differences were found for the dimensions of thinking and perceiving of MBTI in males and females, the extraversion of the Big Five trait in males and age in females. Females with a thinking or perceiving preference were more likely to be misclassified as males, while males with these preferences were more likely to be misclassified as females. In addition, the gender of more extroverted males was more likely to be misclassified and younger females were more likely to be in the misclassified group. These results suggest that a comprehensive psychodemographic profile may be valuable in detecting biases in machine learning models trained on social media texts.

Differences in the total number of comments, the number of English comments (for males), and the percentage of English comments in all comments (for both genders) are also statistically significant. We further analyze our dataset and find that the number of comments is positively and significantly correlated (with p\textless0.05) with thinking (r=0.13 for males, 0.07 for females) and age (r=0.12, 0.11) for both genders and negatively correlated with intuition ($-0.06.$) for females. These correlations may explain the relationship between the number of comments and biased performance. These results highlight the importance of considering the potential impact of psychodemographic variables on machine learning models trained on social media texts.

\begin{table}[]
  \centering
{\small

    \setlength{\tabcolsep}{4.9pt}
    \caption{\label{tbl:fm_predictors} Differences in means of psycho-demographic variables per gender and classification outcome. Significant correlations (*p\textless{.05}, {***}p\textless{.001}) are in bold.}
}
\begin{tabular}{@{}lrrrrrr@{}}

\toprule
& \multicolumn{3}{c}{Female} & \multicolumn{3}{c}{Male}\\
\cmidrule(lr){2-4}\cmidrule(lr){5-7}
Variable  & \ding{51} & \ding{56} & $\Delta$  & \ding{51} & \ding{56} & $\Delta$ \\
\midrule
Age                &  26.78    &  25.83  & \textbf{0.95\nospacetext{$^*$}}   & 25.46   & 26.90 & --1.44  \\
\midrule
I/E        & 0.78    & 0.72    & 0.06   & 0.76    & 0.82  & --0.06 \\
N/S            & 0.86    & 0.91    & --0.05   & 0.92    & 0.93 & --0.01 \\
T/F           &  0.47   &  0.64   & \textbf{--0.17\nospacetext{$^{***}$}}   &  0.61     &  0.29   &  \textbf{0.32\nospacetext{$^{***}$}} \\
P/J         &  0.39    &  0.56  &\textbf{--0.17\nospacetext{$^{***}$}}   &  0.53     &  0.39  & \textbf{0.14\nospacetext{$^{***}$}} \\
\midrule
O           & 61.40   & 68.18 & --6,78   & 64.11   & 67.20 & --3.09 \\
C         & 45.28   & 36.44 & 8.84  & 41.10   & 47.50 & --6.40 \\
E       & 40.67   & 36.44 & 4.23  &  36.68    &  49.60  &\textbf{--12.92\nospacetext{$^*$}}\\
A      & 45.07   & 40.78 & 4.29  & 38.43   & 44.70  & --6.27\\
N        & 50.95   & 53.72 & --2.77 & 46.81   & 47.50  & --0.69\\
\midrule
\#\,com      & 1899 & 2626 & -727 &  3145  &  2034  & \textbf{1112\nospacetext{***}}\\
\#\,en\_com       & 1824 & 2403 & -579 &  2964  &  1900  & \textbf{1064\nospacetext{***}} \\
\%\,en\_com      &  96.17    &  94.43  & \textbf{1.74\nospacetext{**}}   &  94.51    &  95.75  & \textbf{-1.24\nospacetext{**}} \\
\midrule
Gender   & 0.88    & 0.29 & 0.59    & 0.12    & 0.69  & -0.57 \\
\bottomrule
\end{tabular}
\end{table}

%#increased awarness of gender fairness. biased classificators
%we motivaate the need of complete psychodemographics profile to show that even there is fair classficator if we %look only at gender target, it is not so if take complete profile

 \begin{table}
  \centering
{\small
\caption{\label{tbl:fm_confusion} Gender classification confusion matrix on \pandora dataset}}
\begin{tabular}{lrr}
\toprule
  Gold/Predicted& Male & Female \\
 \midrule 
 Male & 1611 & 142 \\
 Female & 192 & 1139 \\
 \bottomrule
\end{tabular}
\end{table}

%\section{Limitations}
\section{Summary}

In this part, we described how we developed two new datasets for personality analysis based on Reddit: MBTI9k and \pandora. These datasets address several deficiencies of existing datasets. Firstly, both are based on a novel data source, enabling comparing insights with other data sources. Secondly, Reddit offers new insights due to its structure revolving around topically focused subreddits and the anonymity of its users, which enables more extensive coverage of topics and increases the availability of personality-related cues.

One of the main obstacles in personality research on text is that most datasets are not publicly available. We have made available the \pandora dataset and two versions of the MBTI9k dataset: (1) a dataset comprising all comments and posts, each annotated with the author's MBTI type, and (2) a subset of this dataset, known as the MBTI9k dataset, which includes comments from all users who contributed over 1000 words. The datasets are available on request.\footnote{\href{https://psy.takelab.fer.hr/datasets/all/pandora/\#dataset-request-form}{https://psy.takelab.fer.hr/datasets/all/pandora/\#dataset-request-form}} and come with a User Agreement aiming to ensure the privacy of Reddit users.

\subsection{Implications}
The experiments demonstrate the potential for recognizing and predicting Redditor personalities based on simple features like word n-grams. These experiments also showed face validity, with most features aligning with expected associations at the dimension and type levels of personality. A key finding is the unique personality distribution on Reddit, which differs from the general population, suggesting a higher prevalence of introverted, thinking, and intuitive types among Redditors. This implies a trend towards higher openness to experience, lower agreeableness, and increased introversion compared to the general population's Big Five personality trait distribution.
The experiment on personality bias in gender classification demonstrated that demographics and personality are crucial for personality-relevant tasks and other tasks that can be influenced by personality. This has implications for various tasks, especially since Reddit data is often used to train conversational models. Our example of a confirmatory study on the propensity for philosophy showed that the \pandora dataset could be used to gain more insights on previous findings from other data sources or theories. On the other hand, our study on using abundant MBTI data to improve Big Five predictions has practical implications for developing more performant systems.
However, it is unlikely that the personality prediction system will perform as well as those used to predict demographic variables. Our analysis shows a clear difference in performance, ranging from over 90\% for prediction gender, down to ~0.7 Pearson correlation coefficient for age, and finally to a range of 0.2 to 0.4 Pearson's correlation coefficient for personality traits. Most of the performance differences come from the construct itself, as used personality models are made to be parsimonious models of the latent structure of what personality is, and we know from research that the more specific personality traits, the higher correlation is expected. This implies that we need to start developing systems at the level of personality facets to improve performance. The other part of the performance differences is mostly due to the availability and quality of personality cues. Some traits are more likely to be expressed on Reddit than others, which can be seen from the prediction metrics, where extraversion is easier to detect than other traits. Future research should shift from averaging performance across all data instances (e.g., users or texts) and instead focus on identifying instances with sufficient personality-relevant cues to output predictions with confidence while preserving face validity.

%\todo{move this part}
%To eliminate topic bias, we removed comments from 122 subreddits focused on MBTI-related topics (representing 7.1\% of all comments) and replaced explicit mentions of MBTI types (and related terms, such as cognitive functions  \cite{mascarenas2016jungian}) with placeholders. Along with comments, we provide each user with their MBTI type and precomputed features (refer to Section~\ref{sec:features}). We have publicly released both datasets,\footnote{\href{http://takelab.fer.hr/data/mbti}{http://takelab.fer.hr/data/mbti}} and have utilized the MBTI9k dataset for subsequent analyses.
%

\subsection{Limitations}
The MBTI9k dataset, with its large sample size, is a valuable source of personality-labeled data. However, we acknowledge the inherent limitations in both the dataset and the models derived from it. First and foremost, the dataset is based on the MBTI personality model, which, despite its popularity in non-academic circles, is not extensively adopted in scientific research due to its lack of empirical validation. The primary critique of MBTI stems from its dichotomous approach to personality traits, which can oversimplify these traits' complex, continuous nature (Section~\ref{section:background_personality_models}).

Another notable limitation of the MBTI9k is the lack of demographic data. This omission raises concerns about the association between language use and personality dimensions, as numerous confounding factors, such as age, gender, and geographic location, can significantly influence these associations (as addressed in the previous section). The data collection method is biased, focusing solely on authors who voluntarily disclosed their MBTI types through flairs. This approach limits the dataset to users who are aware of their MBTI type, actively participate in related subreddits, choose to share this information, have written at least one comment or post, and are familiar with using flairs. Consequently, the distribution of personality types within the dataset may not accurately reflect a larger population.

The self-reported nature of the MBTI types introduces another layer of potential bias, as individuals may have misconceptions about their personality traits or portray themselves differently online. The variability in MBTI tests taken by users further adds to the uncertainty of these labels. Moreover, the temporal scope of the data set (2015-2018) could exclude relevant authors from other periods, and contemporary trends and topics could influence the use of language.

The use of online language may not always reflect an individual's personality, as the specific subreddit, audience, or nature of the online presentation may influence it. However, on Reddit, where anonymity is encouraged, users might have less motivation to misrepresent themselves. Often, Redditors use their personality types to contextualize discussions or seek advice, which could lead to more honest representations of their personalities. However, the language used in online forums can vary significantly from other forms of communication, limiting the generalizability of findings to other contexts.

%\todo{integrate all discussion into the text directly?}
Compared to the MBTI9k data set, the \pandora dataset represents progress because it uses the widely accepted Big Five personality model and includes accompanying demographic variables that may help control for some confounding factors. However, it shares certain limitations with the MBTI9k dataset while introducing some new ones.

One limitation of the \pandora dataset is the potential for selection bias among Redditors participating in discussions about personality, demographics, language, and topics. In addition, processing the Big Five test results included several steps that may have contributed to selection bias. First, only results of personality tests that were in written form were processed, and no screenshots or images of the results were included. Second, a candidate pool of reports was created using the Google Big Query service by identifying the three Big Five traits (i.e., agreeableness, openness, neuroticism). However, comments that were spelled differently or used abbreviations were not processed. Third, for some reports, the type of test or inventory used was uncertain, and a classifier was trained to identify the type of test based on the format or structure of the report. However, the accuracy of this classifier is not perfect, with a macro F1 score of 81.4\%, and some tests may have been misclassified. This could lead to errors in normalizing scores to percentiles, as scores from different tests are not directly comparable.
Despite these limitations, we believe the final scores are as accurate as possible given the ambiguity of the various web services that offer tests, the tests themselves, and how Redditors have reported them in personality-related subreddits.
Finally, the Big Five personality traits are often considered too general to show significant correlations with specific life outcomes, and a more detailed approach using facets would be preferable. Fortunately, facet scores are available in certain personality reports, although they are currently not included in the \pandora dataset. However, they can be added subsequently.

Other limitations are associated with demographic labels in the \pandora dataset. First, gender labels are based on a single pattern (/m/f/) with the highest precision, meaning that users who indicated their gender differently were not labeled. In addition, gender labels were extracted from self-reports, such as ``I am (fe)male'', but there are several ways users can indicate their gender. There are also limitations with the location-based extractors, as only certain locations were captured, and sometimes it was unclear which location the data referred to. For example, some places have the same name but are in different countries and continents. Multiple databases and libraries were used to assign cities to countries, which can also be a source of errors.

In addition, our experiments mainly used simple models, which still leave much room for performance improvement. In this respect, \pandora is particularly well suited for constructing more advanced deep learning models. However, a major unresolved obstacle is creating an efficient user representation based on a potentially large number of comments.

Finally, as in the case of the MBTI9k dataset and the MBTI classifier, we emphasize that predictive models for the Big Five have some predictive validity, but are unreliable at the individual level.

\part{Statement-level Personality Assessment}
\label{part:three}

\chapter{Statement-to-Item Matching Personality Assessment (SIMPA)}
\label{ch:simpa}

Part~\ref{part:two} focuses on developing datasets to overcome the limitations commonly encountered in text-based personality research. The MBTI9k and \pandora datasets are distinguished by their large user base, the quantity and quality of textual content, reliable personality labels or scores, and metadata encompassing key demographic variables that affect the production of written language. The datasets enabled experiments that have highlighted several challenges associated with traditional approaches to personality analysis. 

In this part, we propose an interpretable computational framework that addresses these challenges -- Statement-to-Item Matching Personality Assessment (SIMPA). The work described in this chapter has been first published in \cite{GJURKOVIC2022114}.

\section{Motivation}

A key challenge in ATBPA is the influence of confounding variables on model predictions. These variables, although not directly related to personality traits, can inadvertently skew results. For instance, demographic factors such as age, gender, cultural background, conversational context, or the specific subreddit in which the text appears can all affect language use and complicate personality trait analysis.

Another significant challenge is the high level of noise typical in social media data. This noise stems from diverse factors, including multilingual posts, informal language, varying topics, and short, context-dependent replies. Such variability can obscure meaningful patterns, hindering the extraction of features relevant to personality assessment. Existing feature extraction methods often fail to capture these nuances, as they typically overlook contextual information. For example, while users may explicitly disclose their personality traits in phrases longer than three words, traditional trigram-based models are ill-equipped to detect these signals.

The hierarchical nature of personality traits further complicates modeling user -- generated content from platforms like Reddit. Signals that relate to specific facets of personality may be overlooked when traits are aggregated at the domain level. For instance, wordplay might reflect higher intellect or imagination within the openness domain, yet models focused on surface-level features may dismiss such signals. Similarly, common phrases like ``exactly'', which could indicate agreeableness, are easily lost in aggregate analyses across users.

Additionally, personality cues vary in detectability across traits, leading to more confident predictions for certain traits over others. This variation also extends to individual users, reflecting the complexity of personality expression in text.

To address these challenges, we propose an interpretable methodology that improves the validity of personality assessments. By operating at a granular taxonomy level -- encompassing nuances, facets, and aspects -- this approach aligns more closely with underlying patterns of thought, feeling, and behavior. Rather than relying solely on domain-level scores, our method derives them by aggregating lower-level measures, improving the detection of inadequate or insufficient evidence in personality assessments.

Integrating insights from personality psychology, we aim to bridge the gap between NLP and psychological research. Our model-agnostic framework is designed to keep pace with rapid NLP advancements while ensuring interpretability and explainability.

This chapter introduces a novel approach to ATBPA that addresses the aforementioned challenges. Unlike current methods, which often prioritize predictive accuracy and emphasize convergent validity alone, our approach balances interpretability with comprehensive validity evidence. It integrates the psychometric robustness of questionnaire-based assessments with the efficiency and scalability of text-based methods.

Central to our approach is the use of reliable questionnaire items linked to specific personality traits. By matching these items to user-generated statements on social media, we provide a transparent decision-making process. This alignment not only improves interpretability but also clarifies the rationale behind model predictions. Employing entire statements, rather than isolated words, reduces topic dependence and ambiguity -- common pitfalls of existing methods that primarily focus on lexical features.

\begin{figure*}
    \centering
    \vspace{-1em}
    \hspace*{-0.5cm}
    \includegraphics[scale=0.5]{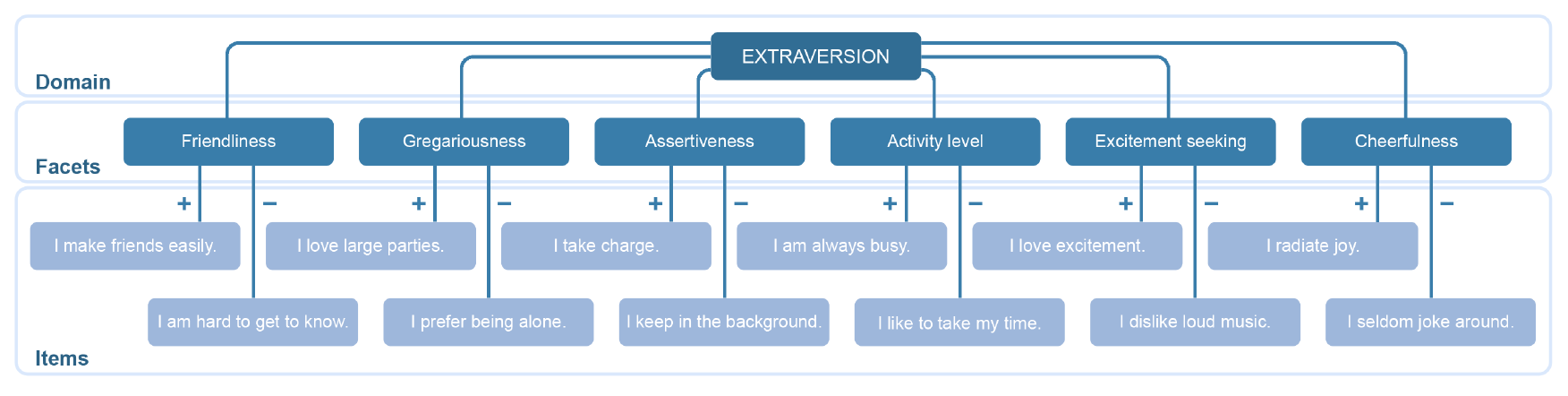}
    \vspace*{-0.5em}
    \caption{Trait hierarchy for the extraversion domain in the IPIP-NEO questionnaire. Questionnaire items are linked to facets, which are grouped into domains. The links between the items and the facets are positively ($+$) or negatively ($-$) keyed.}
    \label{fig:extraversion}
\end{figure*}

Once statements corresponding to questionnaire items are identified, they can be aggregated across various trait levels -- nuances, facets, aspects, or broader dimensions -- as illustrated in Figure~\ref{fig:extraversion}. The polarity of associations, determined by item keys, mirrors standard personality questionnaire scoring, thereby reinforcing the psychometric integrity of our approach.

\section{Conceptual foundations}
SIMPA involves finding textual statements containing self-reported personality descriptions in the target's texts and matching these statements with items from a personality questionnaire. We utilized the RAM (Section~\ref{section:background_personality_jugdment}) as the conceptual basis for SIMPA, which outlines four key stages for accurate personality judgments: cue relevance, availability, detection, and utilization. By following RAM, we draw parallels between how a human judge evaluates the personality of a target who has written a text and SIMPA's operation. Specifically, we begin with relevant questionnaire items for traits (relevance), extract the target's statements from a particular text source (availability), detect the statements that match the items (detection), and use them to assess personality (utilization).

Unlike the original RAM, which relies on a single pass through the four stages to make a judgment, SIMPA utilizes a feedback loop mechanism to allow multiple passes. This feedback loop adjusts the framework's implementation to a particular data source, such as social media texts, by enhancing its sensitivity in identifying relevant cues.

Regarding the technical implementation of the framework, the primary challenge in SIMPA is identifying statements from the target's texts that correspond to questionnaire items. A statement is deemed a good match for an item if its semantic meaning is highly similar to that of the item, indicating the same trait. We express this concept as trait-constrained semantic similarity, which merges semantic similarity with knowledge about how a particular trait may be expressed.

Several NLP tasks, such as paraphrase identification, semantic textual similarity, textual entailment, and natural language inference, are linked to semantic similarity (see Section~\ref{section:background_relevant_tasks}). Recent advancements in NLP, especially in large language models built on deep representation learning, have led to significant progress in these areas. We leverage these advancements by utilizing pre-trained NLP models that can match statements to items with high precision. This allows us to propose a framework that offers interpretable, explainable, and valid personality assessments.

The SIMPA framework is based on matching a target's statements to either questionnaire items or similar trait-relevant statements. This matching process connects a \emph{trait-indicative statement (TIS)} -- a statement taken from the target's texts that serves as a clue for personality assessment -- with a \emph{trait-relevant statement (TRS)} -- a statement with validated links to a specific trait. Initially, the set of TRSs consists only of questionnaire items; however, it can be expanded to include TISs considered significant for a particular trait.

\begin{figure*}
    \centering
    %\vspace{-1em}
    %\hspace*{-0.5cm}
    \includegraphics[scale=0.75]{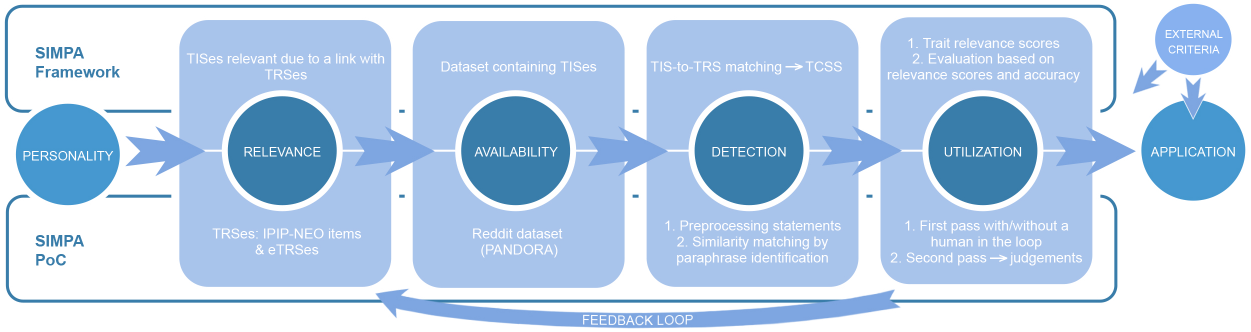}
    %\vspace*{-0.5em}
    \caption{A schematic illustrating the SIMPA framework and one possible implementation of the framework (Chapter~\ref{ch:implementation}), divided into four stages: relevance, availability, detection, and utilization, along with a feedback loop and the application of acquired personality judgments. SIMPA thus conveys the concept of the iterative flow of relevant textual cues (TISes) through all stages, resulting in personality judgments that can be applied in various contexts applications.}
    \label{fig:framework}
\end{figure*}

Figure~\ref{fig:framework} presents an overview of the SIMPA framework. This framework aligns with RAM, which divides the personality judgment process into four stages: cue relevance, availability, detection, and utilization. The relevant cues are the TISes, which must be present in sufficient quantity in the source text for accurate judgments to be made. During the detection phase, TISes are matched to TRSes based on the concept of trait-constrained semantic similarity. Finally, in the utilization stage, the identified TISes are assigned relevance scores, which are then combined to produce trait-level scores.

By adopting RAM as the foundation for SIMPA, we can identify weaknesses in the pipeline, specifically the amount and types of errors transmitted from one stage to the next. In SIMPA, we distinguish between two types of errors: \emph{errors of omission} and \emph{errors of commission}. Errors of omission refer to failing to detect a relevant cue or to link it to a trait, while errors of commission involve detecting an irrelevant cue or incorrectly linking a relevant cue to a trait. Both types of errors can negatively impact the accuracy of personality judgments. To address these errors, we designed a framework that includes a feedback loop mechanism, allowing multiple passes through the four RAM stages to obtain data-specific TRSes, thereby increasing the chances of identifying TISes.

\section{Stages}
\subsection{Relevance}
\label{sec:framework-relevance}

RAM starts with the idea of relevance, which asserts that for a person to make accurate trait judgments, the individual must display behavior, thoughts, or emotions related to the trait being assessed \cite{funder1995accuracy}. For example, to accurately judge someone as extroverted, that person must exhibit behaviors or thoughts associated with extroversion; otherwise, it would be difficult, if not impossible, to make such a judgment. In SIMPA, we consider relevance as an inherent property of a TIS and assume that its relevance can only be established through external knowledge that validates its accuracy. This contrasts with most ATBPA research, where cue relevance is equated with feature weights from a supervised model trained to predict questionnaire scores as labels, which lowers interpretability and generalizability, threatening the validity of these methods.

SIMPA resolves this issue by matching TISes to TRSes, where TRSes are questionnaire items that psychologists have carefully crafted and validated to capture specific personality traits. The reasoning is that if a target utilizes a statement that conveys the same meaning as a questionnaire item, then the judge can consider that statement a valid personality cue supported by content-based validity evidence. In other words, a TIS is deemed relevant for a trait if it corresponds well with the TRS associated with that trait.

When comparing our RAM-based approach to the Brunswik Lens Model (Section~\ref{section:background_personality_jugdment}), TRSes serve as proximal cues for identifying distal cues emitted by a target, offering ecological validity (i.e., relevance) regarding the target's actual personality trait.

\label{sec:tis-types}
\subsubsection{Types of Trait Indicative Statements}

We classify TISes into two main categories: (1) \emph{self-concepts of personality} and (2) \emph{personality manifestations}. Self-concepts of personality are statements in which the target explicitly defines themselves in terms of personality (e.g., ``I am a disciplined person''). In contrast, personality manifestations are not as self-categorizing and can be subdivided into trait references and individual acts. A trait reference is a statement in which a speaker makes a generalizable reference to emotions, cognition, or behaviors known to be indicative of personality (e.g., ``I never procrastinate'' as a cue for high self-discipline). Individual acts refer to personality-related actions described by the target (known as act reports) or observed by the target (known as act observations) (e.g., ``I finished all of my chores before noon'' as a cue for high self-discipline, or ``I will answer you after I finish my work'' as a cue for high self-discipline).

Using TISes in ATBPA has several advantages over other textual cues used in previous work. Firstly, TISes are explicit and strong signals of personality. Secondly, they are less prone to contextual meaning alterations, as they usually include enough context to reliably determine the semantic relationship between a TIS and a TRS. Lastly, TISes are, to some extent, domain-independent, as the same TISes can appear across different social media platforms, although their distribution may depend on platform characteristics. We further discuss TIS availability and other related aspects in the following section.
\subsection{Availability}
\label{sec:framework-availability}

Relevant cues must be accessible to the judge to allow for accurate trait judgment. In ATBPA, the availability of these cues primarily depends on the textual data source, which ideally maximizes both the quantity and quality of relevant cues. In SIMPA, this enhances the likelihood of identifying pertinent TISes for a particular set of TRSes. TIS availability is influenced by several factors related to the communication context, target, and trait.

When utilizing social media texts as the data source, environment-related factors encompass a range of platform-specific characteristics (e.g., reliance on textual communication, format restrictions, subject matter focus, synchronous communication) and user demographics and preferences \cite{mcfarland2015social}. For instance, communication on Twitter and Reddit typically occurs through text messages, whereas on Instagram and TikTok, it is mostly visual. Moreover, some platforms impose length limits on messages (e.g., Twitter), while others do not (e.g., Reddit). User anonymity on the platform can also influence TIS availability, as users tend to provide more self-related information online when they believe they are anonymous \cite{joinson2001self, chiou2006adolescents}. Another essential characteristic is the platform's subject matter focus, which may impact the distribution of available TISes. For instance, LinkedIn is predominantly used for work-related activities, making work-related TISes more readily accessible on LinkedIn than on a platform with a broad range of topics like Reddit. The topical focus is also associated with another factor that affects TIS availability, i.e., the need for self-introduction, as some discussions require users to provide more background information about themselves. For example, in conversations about personal relationships, users tend to offer more personality-related descriptions of themselves.

Additionally, the availability of TIS is influenced by varying social norms across platforms and their sub-communities. For example, in some communities (e.g., Facebook groups for first-time parents), a critical aspect of interaction is providing encouragement, which can boost the availability of TIS related to individual acts. Finally, the choice of platform can also affect user demographics, which, in turn, may correlate with personality differences and TIS availability \cite{tay2020psychometric}. For instance, the average Reddit user is a young male \cite{duggan20136}, potentially increasing the availability of TIS associated with this age or gender group (e.g., ``I enjoy being part of a loud crowd'').

The availability of TIS is influenced by target-related factors, including social desirability and the situations in which targets find themselves, as well as the environmental factors mentioned earlier. Some individuals are more inclined to express themselves in a socially desirable way, which affects the availability of TIS. Furthermore, specific situations can evoke particular traits, impacting cue availability. For instance, being accused of lying can trigger cues related to the cooperation facet, which may be either positively or negatively keyed.

The target's personality traits also influence TIS availability. Traits associated with many highly visible and frequently available cues are judged more accurately than those associated with less visible and less likely available cues. However, not all traits have the same likelihood of being manifested in online text communication, and some traits may be inherently difficult for a person to express as a TIS. For instance, individuals low on the self-consciousness trait may express fewer negatively keyed TISes related to self-consciousness than positively keyed ones.

Moreover, the factors affecting TIS availability may interact with one another. For example, anonymity on online platforms can encourage users to discuss different, potentially embarrassing topics, thus reducing the impact of social desirability bias. The social media platform Reddit is a topically diverse and anonymous platform that may allow users to be more flexible in their behavioral choices and express TISes that they might otherwise hide.

%\todo{prehoc vs posthoc }

\subsection{Detection}
\label{sec:framework-detection}

RAM's detection stage involves the judge identifying relevant and available personality cues \cite{letzringinpress}. In SIMPA, the detection stage seeks to identify as many TISes as possible within the target's text. This is accomplished by comparing the target's statements to each one in the set of TRSes, which initially consists of questionnaire items. However, the lexical gap between TISes and TRSes creates a challenge, as TISes are colloquial expressions taken from naturalistic texts, while TRSes are carefully crafted statements that convey the core meaning of emotional, cognitive, or behavioral personality-related patterns. To tackle this issue, relying solely on verbatim matching of TISes to TRSes would lead to numerous errors of omission in SIMPA.

To tackle this challenge, SIMPA operationalizes TIS-to-TRS matching as a similarity measure between two statements. If a statement is identical or highly similar to a given TRS, it is likely a TIS and relevant to the same trait as the corresponding TRS. This similarity measure represents a graded concept that extends beyond semantic equivalence, encompassing both semantic similarity and the trait-indicativeness of the two statements. This measure is referred to as Trait-Constrained Semantic Similarity (TCSS). TCSS ensures that the similarity between a TIS and a TRS captures semantic similarity and signifies the same underlying trait. For example, the statement ``I just hate crowded places'' aligns closely with the TRS ``I avoid crowds' and also reflects the same extraversion facet (gregariousness) with the same key. In contrast, the statement ``I work in a crowded place'' is similar to the TRS but does not indicate the same trait. Likewise, the statement ``I don't like being in shopping malls on Saturdays'' signifies the same trait but is less semantically similar to the TRS.

In essence, TCSS provides a context-agnostic measure of similarity between two statements, capturing their semantic equivalence in typical or default contexts. Despite its limited scope, TCSS integrates diverse domains of human and expert knowledge, including natural language semantics, pragmatics, common-sense reasoning, and sociocultural understanding relevant to personality trait assessment. If we compare our RAM-based approach with the Brunswik Lens Model (see Section~\ref{section:background_personality_jugdment}), TCSS is related to perception modeling as it tries to bridge the gap between proximal and distal cues. Ideally, TCSS should indicate the degree of agreement between a target's statement and a TRS, measured on a Likert scale. However, providing a more precise definition of TCSS is challenging due to the complex interactions between various linguistic phenomena. While future work may endeavor to develop a formalized characterization of TCSS that adheres to certain similarity measures' desired properties (e.g., a decrease in similarity with negation or epistemic hedging), we leave this pursuit for future research.

%\todo{alternative to tcss is using implicit detection; generate answer and reasons which increase confidence}

The models developed for tasks such as paraphrase detection and semantic textual similarity (see Section~\ref{section:background_relevant_tasks}) can be used off-the-shelf for the detection stage of the SIMPA framework, as they effectively reduce both omission and commission errors. However, some errors may still occur since these models were not explicitly designed for TCSS. Adapting these models to the TCSS task using transfer learning \cite{ruder2019transfer} could reduce these errors. Still, it would require a dataset labeled with TCSS scores, and the adaptation process would not be straightforward.

The selection of the NLP model for TCSS plays a central role in determining the number of errors of omission and commission. An error of omission occurs when a relevant cue is not detected, which primarily happens when the model fails to assign a high TCSS score to a vernacular statement that shares the same meaning as a particular TRS. Conversely, an error of commission arises when the model mistakenly identifies irrelevant statements as relevant, assigning them an excessively high TCSS score. Such errors can occur when two statements that are very similar in wording convey different meanings, particularly when one statement contains negations or antonyms. In this situation, the TIS does not indicate the same trait as the TRS, even though they are similar. However, it remains trait-relevant, indicating that errors of commission can be divided into two types: (1) informative errors, where the TIS does not align well with an existing TRS but is still trait-relevant to a judge and may even be considered a new TRS, and (2) noninformative errors, where the TIS fails to provide any useful trait-relevant information to a judge (e.g., a TIS ``I feel welcomed'' matched to the TRS ``I make people feel welcome''). During the utilization stage of SIMPA, we use informative commission errors to improve the cue detection process iteratively.

\subsection{Utilization}
\label{sec:framework-utilization}

In the SIMPA framework, the final stage is utilization, where the identified TISes are employed to accurately assess a target's personality traits. This stage presents several challenges, including the fact that the relevance of a TIS for a specific trait is a matter of degree. Additionally, multiple TISes may serve as cues for a single trait, cue relevance from lower-level traits combines into cue relevance for higher-level traits, and errors from earlier stages must be taken into account for accurate judgments.

To address these challenges, the utilization in SIMPA involves two steps: (1) recursively determining the relevance scores of cues for traits across all levels of the trait hierarchy for each target, starting from TISes at the lowest level and moving up to the level of trait domains, and (2) deciding whether to make another pass through the RAM stages using the feedback loop mechanism based on the obtained relevance scores and external criteria. These steps acknowledge the hierarchical nature of personality traits, where nuances at the bottom level connect to facets, which connect to aspects, and finally to trait domains. Additionally, they offer insights into the contribution of each lower-level trait since the evaluation of traits at higher levels is estimated based on aggregate judgments of traits at lower levels.

\paragraph{Relevance scoring.} 
We define the relevance score of a cue for a trait as the output of a scoring function that combines the relevance scores of cues for traits at lower levels of the trait hierarchy. Starting from the lowest level of TRSes, the relevance score for each TRS is obtained by aggregating the relevance scores of the matching TISes, using the TCSS score as a measure of similarity (Section \ref{sec:framework-detection}). The relevance of a cue for a TRS is determined by considering the combined evidence of TISes that match the TRS, with attention to the strength of that match as indicated by the TCSS score. The resulting relevance score of TRSes can then be further aggregated to produce the relevance score of cues associated with traits at higher levels (e.g., facets or aspects) linked to these TRSes. This enables the relevance scores of cues to be propagated up the trait hierarchy to the level of trait domains, allowing judges to select the hierarchical level of the traits for which they wish to obtain relevance scores.

Furthermore, the relevance score of a cue generally depends on the context, which includes both the number of cues and extralinguistic factors, such as social norms within a sub-community and demographic characteristics. The number of cues is particularly important, as it influences confidence in judgment. For example, a judge can be more certain that a target scores highly on the intellect facet if the target's TISes include multiple statements such as: ``I have read Nietzsche's elaboration on that'', ``I have read a scientific paper on that topic'', and ``I just finished a book by Stephen Hawking''. However, even when the number of cues is low, high relevance can compensate for this limitation. For instance, a single TIS that matches the TRS ``I love to read challenging material'' may serve as a sufficiently strong cue to infer that the target is high in intellect.

Comparing our approach to the Brunswik Lens Model, the relevance scoring process aligns with perception modeling, while the utilization process involves verifying representational validity before proceeding with judgment.

\section{Feedback loop} 
\label{section:simpa_feedback_loop}
To improve judgment accuracy despite the lexical gap between TISes and TRSes, SIMPA incorporates a feedback loop from the utilization stage to the availability stage, which allows for iteration through the RAM stages until a specified criterion is met. The feedback loop serves two purposes: (1) to adapt TIS detection to the source-text language by incorporating TISes into the set of TRSes and (2) to increase confidence in judgments when desired criteria are not met, such as the minimum number of detected TISes per trait or the minimum relevance score. The decision to loop back relies on relevance scores and judgment accuracy, which is determined by external criteria such as validity evidence.

Before looping back, the set of TRSes is expanded with TISes that contain sufficient trait-relevant information to function as TRSes. The TISes considered for promotion to TRSes fall into two categories: (1) TISes with high TCSS scores that are paraphrases of existing TRSes, and (2) TISes resulting from informative errors in the detection stage, which are highly indicative of a trait but do not match existing TRSes closely. The first category can be viewed as domain adaptation, while the second category encompasses new nuances of certain traits. The enlarged set of TRSes is applied in the next iteration of the RAM stages, where available and relevant cues are again detected and used to determine relevance scores. The loop may be initiated multiple times until no new TISes can be promoted to TRSes or the desired confidence criteria are satisfied.

The iterative feedback loop in SIMPA underscores its adaptability and robustness in personality assessment. By continually refining the set of TRSes through the integration of relevant TISes, SIMPA enhances its ability to capture nuanced personality cues across various contexts and linguistic expressions. However, in addition to improving judgment accuracy, the feedback loop serves a secondary function: systematically labeling expressions used by real individuals. This process effectively transforms large volumes of unstructured textual data into a structured database of labeled TISes, which holds significant value beyond its immediate application in personality assessment. By incorporating naturally occurring linguistic expressions that align with validated personality indicators, SIMPA can aid in developing a comprehensive repository of real-world personality cues.

%\todo{expand a bit}

%\section{Interpretability in SIMPA}

%\todo{add some diagram showing how we translate unstructured info to structured}

%- demonstrate and unroll a judgment from the results toward the start
%- compare SIMPA to previous methods, feature-based machine learning, and deep learning
%- show how 

%\section{ Implementation of a statement-level model for interpretable personality assessment on Reddit}
%\begin{itemize}
%    \item framework is nice, but we need to operationalize it
%    \item describe what we did to implement it
%    \item pandora dataset: availability: initial; we'll report back real results
%    \item how we chose TRSes; expert TRSes
%    \item feedback loop in real life
%    \item unsupervised setup: why we want it or need it
%    \item how we validate it
%    \item supervised setup: why
%    \item results
%\end{itemize}

\chapter{Implementation of the SIMPA framework}
\label{ch:implementation}

The SIMPA framework provides a structured and adaptable approach for analyzing different aspects of authors' personalities based on their written language. Its flexibility arises from its modular, staged design, allowing for variations in the implementation of stages or even merging stages to suit specific research objectives or practical applications. However, the interdependence of stages necessitates a holistic approach, as decisions at one step propagate to and influence subsequent ones and vice versa because of the feedback-loop mechanism. For instance, while the availability stage primarily focuses on the choice of data source, this decision must align with selecting appropriate TRSes, as lexical or contextual mismatches may limit their utility. Furthermore, the TRSes chosen affect how subsequent stages, such as detection and utilization, are executed. These interconnected decisions are often grounded in the specific application at hand. For example, in an unsupervised analysis aimed at predicting facet-level personality traits on Reddit, the choice of data source and aggregation method for TISes would be constrained early in the process, guiding the design of all subsequent stages.

In this chapter, we will present a set of experiments to explore the implementation of SIMPA for supervised and unsupervised ATBPA. First, we will explain design choices that simplify the implementation of different SIMPA stages while getting a valid and interpretable personality assessment. We will then focus on exploring different types of TRSes or relevant cues as they directly influence interpretability and serve as a grounding factor for the use of computational methods and personality psychology. Finally, we will present the applications of the model to two different use cases for supervised and unsupervised ATBPA.
%\todo{add more}

%All of the experiments will use the data from the social media site Reddit.
%n Chapter 3\todo{change}, we discussed the benefits of using Reddit as a valuable data source for personality research, including (1) user anonymity, which promotes free expression of thoughts, (2) topical diversity, with over two million subreddits covering many aspects of personality, and (3) the high quality and quantity of text per user, which increases the likelihood of finding personality-relevant cues.
%
%e will use the \pandora dataset (\todo{backref to dataset chapter}) for all experiments. We will rely on user self-reported Big Five scores as they provide evidence of validity for new ATBPA methods and enable direct comparison of predictive accuracy. The \pandora dataset includes a total of 1.3 million comments written by 1,608 users, with a total of 14.3 million sentences.
%
%Extracting and aggregating personality-relevant cues from such a large amount of text for each user poses a challenge. However, we believe that this will not impede our approach, as SIMPA should benefit from more data, increasing the likelihood of finding TISes.

\section{General methodology and design choices}

%- The relevance and availability stages are in large orthogonal to the choice of technology. However, to get availability estimates that can serve as a proxy for quality of TRSes, the detection step has to be implemented and run against the dataset of choice.
%- for that reason, before discussing different types of TRSes we have to decide on the implementation of the detection step
%- we focus more on establishing methodology to increase interpretability than on the performance of the model
%- because of that, we developed annotation schema to check the quality of detection step 
%- we choose to simplify the process by using semantic similarity metrics with underlying embedinng model based on simple paraphrase based models 
%

%The stages of relevance and availability in the SIMPA framework are largely independent of the specific technology used. However, the availability stage requires estimates that can act as proxies for the quality of TRSes. To generate such estimates, the detection step must first be implemented and applied to the chosen dataset. Before exploring different types of TRSes, it is essential to define the methodology for the detection step. Similarly, to generate user-level personality assessment, the detected TISes have to be combined in an aggregated estimate in the utilization stage. In this section, we will define and explain design choices that will be used in subsequent experiments exploring practical implementation for 
%

%\section{Detection}
%\label{sec:implementation-detection}
%\todo{again, write this as it was planned}

\subsection{Dataset}

In Chapter~\ref{ch:personality_demographics_reddit}, we discussed the benefits of using Reddit as a valuable data source for personality research. These benefits include (1) user anonymity, which encourages the free expression of thoughts, (2) topical diversity, with over two million subreddits covering various aspects of personality, and (3) the high quality and quantity of text per user, which increases the likelihood of discovering personality-relevant cues.

For all experiments, we will use the \pandora dataset (Chapter~\ref{ch:personality_demographics_reddit}). We will rely on user self-reported Big Five scores, as they provide evidence of validity for new ATBPA methods and allow for direct comparison of predictive accuracy. The \pandora dataset consists of 1.3 million comments written by 1,608 users, totaling 14.3 million sentences.

\subsection{Availability}
%\todo{expand this session; explain how people are often outputting tis-es while explaining something; more people-related subreddits have more TIS-es (can we get some numbers); explain other ways of estimating availability; prehoc and posthoc - and motivate what we will do}
SIMPA relies on the availability of TISes, which can be influenced by various factors related to the environment, the target, and the trait, as discussed in Section~\ref{sec:framework-availability}. For ATBPA on Reddit, several environment-related factors favor the availability of TISes, including user anonymity, the large number of users and texts per user, and the topical diversity of subreddits.

To estimate the availability of TIS candidates in the data, we can use pre-hoc analysis and post-hoc analysis. In pre-hoc analysis, we approximate the number of potential TISes using count-based methods that rely on closed-vocabulary approaches. For example, one can identify sentences that may contain TISes by counting those that include the pronoun ``I''. In this case, nearly 5.2 million such sentences exist in the \pandora dataset, corresponding to 36.3\% of all sentences. This high percentage suggests an increased likelihood that a large number of TISes are present in Reddit texts.

However, this approach serves only as a quick but unreliable method for assessing the actual availability of TISes. A more rigorous strategy for accurately estimating availability involves post-hoc analysis, which consists of conducting multiple iterations of SIMPA loop runs followed by enumerating the detected TISes. A detailed availability analysis is provided in the upcoming sections of this thesis.

\subsection{Detection}
\label{section:simpa_impl_detection}
Instead of focusing on the performance of detection models (e.g., maximizing accuracy or precision-recall metrics), our main goal is to develop a methodology that enhances the interpretability of the detection process. This is particularly important in the context of personality assessment, where transparency and explainability are vital for validating both the framework and its outputs.

To achieve this goal, we outline a comprehensive annotation schema for evaluating the quality of the detection step. This schema facilitates a systematic analysis of the types of errors encountered during detection, including errors of omission (failing to identify relevant TISes) and commission (false positives). It offers insights into the alignment between detected TISes and the theoretical constructs underlying the TRSes, which can be employed in the feedback loop to enhance detection (i.e., by adjusting detection thresholds) and TRS design.

To streamline the implementation of the detection step, we opted for a simplified approach based on semantic similarity metrics. This decision was motivated by the need to balance interpretability with computational efficiency. Specifically, we employed paraphrase-based models that generate embeddings for input texts and compare their similarity scores to predefined thresholds. The rationale for choosing paraphrase-based models is that they offer a straightforward way to quantify the similarity between user-generated text and TRSes and can be a strong baseline against which more complex detection methods can be compared in future experiments.

\subsubsection{Semantic similarity matching}
The detection step relies on TCSS (subsection~\ref{sec:framework-detection}, which can be approximated by calculating semantic similarity between TRS-es and TIS candidates (see Section~\ref{section:background_relevant_tasks}). However, for added interpretability in some use cases, a human-in-the-loop scenario can be desirable. In this section, we will describe both scenarios in more detail, which will be used in subsequent experiments.

The typical computational scenario for detecting TISes uses TCSS to match the target's statements to TRSes. This is done in two steps: (1) text preprocessing and (2) similarity matching. In the preprocessing stage, we begin by splitting the comments of each Reddit user into sentences using Spacy's Sentencizer \cite{spacy2020}. Each sentence is then treated as a potential TIS candidate. Although clauses may more closely align with TRSes, we choose to use entire sentences to simplify and expedite preprocessing, as most NLP models are trained on sentence-level data, and the clause-level processing mandates the use of syntactic parsers or part-of-speech tagging. However, we acknowledge that this simplification may lead to increased errors of both omission and commission. Additionally, we filter the sentences to include only those that contain the pronoun ``I''. However, demanding the use of the first-person pronoun ``I'' diminishes recall efficiency, given that certain self-referential statements do not necessarily require its inclusion.

Once we have obtained a set of sentences for all targets, we proceed to the similarity-matching step. To operationalize TCSS, we utilize paraphrase detection and semantic textual similarity, employing off-the-shelf pretrained models for these tasks. This simplification allows us to avoid model fine-tuning, which would necessitate a dataset labeled with correct TIS-to-TRS matches. Such a dataset is likely unavailable at this stage, especially if it originates from a new data source with its own linguistic characteristics. However, this simplification also increases commission errors (where some TISes are incorrectly matched to a TRS) and omission errors (where some TISes are not correctly matched to a TRS). While numerous NLP models are available for these tasks, we evaluate three distinct and commonly used models from the SentenceTransformer package 
\citep{reimers-2019-sentence-bert}: (1) semantic textual similarity models based on Siamese networks 
\citep{reimers-2019-sentence-bert}, (2) the Komnios word2vec model 
\citep{komninos-manandhar-2016-dependency}, and (3) the RoBERTa-based paraphrase detection model 
\citep{reimers-2019-sentence-bert}. We specifically select these models because they employ different conceptual approaches and may therefore yield varied performance when applied to our task.

We follow the NLP practice described in \cite{reimers-2019-sentence-bert} to determine the similarity between a TIS candidate and a TRS. We start by encoding each TIS as a vector in a high-dimensional vector space. Then, we calculate the semantic similarity between all TIS-TRS pairs using the cosine similarity metric between their corresponding vectors. We save these similarity scores in a matrix with rows for each target sentence and columns for each TRS. For each Reddit user, we create one matrix for each of the three NLP models. This process generates a distribution of similarity scores for each sentence and all TRSes. However, to simplify the matching process, we only consider the TRS with the highest similarity score as a match for each sentence. This means that each TIS can match only one TRS. Although this approach allows TIS to inherit TRS keys and traits, it also means that we pass on the opportunity to describe each TIS by the distribution of related keys and traits.

Despite being simplified, the detection stage still requires several design decisions, such as choosing the NLP model and the cosine similarity threshold. To further explore these decisions, we selected 100 sentences with the highest similarity scores for each model and qualitatively evaluated their performance. Our analysis revealed two significant findings: First, the similarity threshold directly affects the number and relevance of TISs that pass the threshold. Second, the similarity scores of different models vary significantly between TRSes. Although applying different thresholds for each TRS would be the best way to ensure consistent performance and optimal precision and recall of TISes, there are no practical and straightforward ways to implement it. Because of that, we evaluated the selected models in search of the most performant and practical in terms of smaller variance in thresholds between TRSes. The paraphrase detection model performed the most consistently in relevance and quantity at a fixed similarity threshold of the three models. Additionally, it had the highest overall TCSS and a better ratio of informative to noninformative errors of commission. Therefore, we only used the paraphrase detection model to simplify the implementation.

%We investigate this in the second scenario by involving a human expert (a personality psychologist) in the loop. For each TRS in the original set of 300 IPIP-NEO items, the expert was asked to annotate 20 statements from the dataset that were most similar to it according to cosine similarity.
%\todo{ensure this is mention somewhere}
\subsubsection{Human-in-the-loop}
\label{subsubsection:simpa_impl_human_in_the_loop}
The above-described setup simplifies both the detection and utilization stages by using semantic similarity as a proxy for TCSS and equating relevance scores with TCSS. However, this raises the question of how much more accurate judgments could be made if more sophisticated methods were used in these two stages.

We establish a performance baseline by incorporating a human expert (a personality psychologist) into the evaluation process. Table~\ref{tab:annotation_scheme} presents the annotation scheme employed by the expert, which operates on a binary distinction between correct and incorrect matches. Incorrect matches are further classified into six subcategories: five informative errors and one noninformative error. This results in a total of seven classification categories.

\textbf{Correct match}—The TIS maintains the same level of generality and polarity as the TRS, accurately reflecting the intended personality trait without any shift in specificity or sentiment (e.g., TRS: ``I'm always prepared.'' $\rightarrow$ TIS: ``I'm always prepared for whatever comes my way.'').  
\textbf{Same level of generality but opposite polarity}—The TIS directly contradicts the TRS while preserving the same level of specificity (e.g., TRS: ``I'm always prepared.'' $\rightarrow$ TIS: ``I'm never prepared.'').  
\textbf{Less general but same polarity}—The TIS conveys a more specific instance of the TRS while maintaining its original sentiment (e.g., TRS: ``I'm always prepared.'' $\rightarrow$ TIS: ``I am prepared.'').  
\textbf{Less general and opposite polarity}—The TIS represents a more specific instance of the TRS but reverses its polarity, contradicting the intended trait expression (e.g., TRS: ``I'm always prepared.'' $\rightarrow$ TIS: ``I came unprepared.'').  
\textbf{Points to the average score item}—The TIS expresses a more neutral stance rather than indicating a strong presence or absence of the trait (e.g., TRS: ``I'm always prepared.'' $\rightarrow$ TIS: ``I'm never fully prepared, but I'm not unprepared either.'').  
\textbf{Other item/facet of the same domain}—The TIS pertains to the same personality domain but does not directly correspond to the TRS (e.g., TRS: ``I'm always prepared.'' $\rightarrow$ TIS: ``I always arrange things in order.'').  
\textbf{Other (a noninformative error)}—The TIS is unrelated to the intended personality trait and does not provide a valid personality cue (e.g., TRS: ``I'm always prepared.'' $\rightarrow$ TIS: ``I prepared a meal.'').  

\begin{table}[ht]
\centering
\small
\caption{\label{tab:annotation_scheme} Annotation scheme for detection step}
\begin{tabular}{c|l|p{8cm}}
\hline
\textbf{Category} & \textbf{Description} & \textbf{Example (for TRS: ``I'm always prepared'')} \\ \hline
1 & Correct match & ``I'm always prepared for whatever comes my way.'' \\ \hline
2 & Same generality, opposite polarity & ``I'm never prepared.'' \\ \hline
3 & Less general, same polarity & ``I am prepared.'' \\ \hline
4 & Less general, opposite polarity & ``I came unprepared.'' \\ \hline
5 & Points to average score item & ``I'm never fully prepared, but I'm not unprepared either.'' \\ \hline
6 & Other item/facet of the same domain & ``I always arrange things in order.'' \\ \hline
7 & Other (noninformative error) & ``I prepared a meal.'' \\ \hline
\end{tabular}
\end{table}

\subsection{Utilization}
\label{sec:utilization}

In both detection scenarios, multiple passes through the RAM stages are conducted, and the set of TRSes may be expanded before beginning a second pass. In the first scenario, a threshold is applied to determine which TISes should be promoted to TRSes. In the second scenario, TISes are promoted only if an expert judges that they accurately match a particular TRS. In both cases, the newly promoted TRSes are assigned to the same trait(s) as the original TRSes they resembled.

Once the relevant TISes are identified for each target, the next step is to calculate trait scores at various levels of the trait hierarchy. To keep the process simple, we assign a relevance score of \(1\) to every detected TIS and then aggregate these values to obtain both facet- and domain-level scores. First, we sum all TIS scores corresponding to a specific facet key (e.g., \(+1\) or \(-1\)), thereby producing a \emph{facet score}. We then sum these facet scores to obtain \emph{domain-level} trait scores.

Let \(\{tis_1, tis_2, \ldots, tis_n\}\) be the set of \(n\) TISes detected for a given user. We fix each TIS relevance to:
\begin{equation}
\mathit{tis}_i = 1 \quad (i = 1, 2, \ldots, n).
\end{equation}
We then compute the facet-level scores, where \(m\) is the total number of facet keys:
\begin{equation}
f_j = \sum_{i=1}^{n} tis_i 
\quad (j = 1, 2, \ldots, m).
\end{equation}
Next, we compute domain-level scores by summing all facet scores associated with each domain, where \(p\) is the number of domains:
\begin{equation}
d_k = \sum_{j=1}^{m} f_j 
\quad (k = 1, 2, \ldots, p).
\end{equation}

After calculating domain scores for every target, we derive their \emph{relative expressions} of each trait by comparing these scores across the entire sample. To do this, we first compute the proportion of positively keyed versus negatively keyed TISes for each target. Let \(p_{t,i}\) be the proportion of positively keyed \(tis_i\) for target \(t\):
\begin{equation}
p_{t,i} = 
\frac{\text{number of positive } \mathit{tis}_i \text{ for target } t}
{\text{total number of } \mathit{tis}_i \text{ for target } t}.
\end{equation}

We sort all targets based on \(p_{t,i}\). The percentile score for target \(t\) and trait \(i\) is then:
\begin{equation}
\mathit{percent}_{t,i} 
= 100 \times \frac{\mathit{rank}_{t,i}}{N},
\end{equation}
where \(\mathit{rank}_{t,i}\) is the rank of target \(t\) for trait \(i\) (in ascending order of \(p_{t,i}\)), and \(N\) is the total number of targets. This ranking approach creates a relative percentile for each user, indicating how strongly they express a given positively keyed (or negatively keyed) trait compared to others in the dataset.

By systematically applying multi-pass detection with TRS expansion, aggregating TISes into facet and domain scores, and computing percentile ranks, we can generate interpretable trait profiles that position each target within the overall sample.

\section{Relevance: Different types of TRSes}

The design choices outlined for all stages of the SIMPA framework form the basis for a series of experiments aimed at providing essential insights into the implementation details necessary for conducting precise and interpretable personality assessments.

In the relevance stage, we focus on selecting and evaluating TRSes, which serve as versatile tools for linking textual cues to personality traits. Unlike traditional questionnaire items, TRSes offer greater gradation and variety, making them more adaptable to diverse personality analysis tasks. In the following experiments, we examine four distinct categories of TRSes, each contributing uniquely to the SIMPA framework.

The first category includes items from established personality questionnaires, such as the IPIP-NEO 300. These pre-validated items provide a comprehensive and balanced representation across facets, enabling systematic testing, replication, and benchmarking of SIMPA’s performance without requiring specialized psychological expertise.

The second category comprises expert-crafted TRSes, designed to address targeted research questions or practical applications. These statements leverage domain expertise to improve relevance and interpretability. By including this category, we aim to assess the additional value that expert knowledge can bring to the framework.

The third category explores TRSes generated using large language models (LLMs). This involves analyzing the performance of LLM-generated TRSes and validating them through strategies such as human annotation or comparative assessments with other LLMs. By incorporating this approach, we aim to expand the repertoire of TRSes and evaluate the scalability of automated generation methods.

Finally, we investigate TRSes promoted from TISes that match previously described TRS types. These statements are notable for their organic emergence from how individuals naturally describe themselves or others in text. We explore the feasibility of labeling these TISes using human-in-the-loop or ML models.

\subsection{IPIP based TRSes}
\label{section:simpa_impl_trs_ipip}
%\todo{practical implementation; list and explain all versions of TRSes and motivate what we will use in our experiments}

One of the obvious choices for TRSes are items from personality questionnaires designed and validated by experts to target specific personality traits (see Section \ref{sec:framework-relevance}). This implementation uses 300 items from the IPIP-NEO questionnaire \cite{goldberg1999ipip}, which are open source and available on the International Personality Item Pool website. There are other questionnaires for personality assessment; however, we decided to use IPIP-NEO to make the number of TRS-es and matched TISes reasonably low so we can manually check the quality of the matching process. Furthermore, most of our users' personality tests with Big Five scores were based on this questionnaire. To enhance the TIS-to-TRS matching process, we transform each item into a self-referential sentence by adding the pronoun ``I'' at the beginning of the item. This conversion (e.g., ``often feel blue'' to ``I often feel blue'') has been determined to improve the performance of the NLP models used in the detection stage.

%As we will show later, we expanded the initial set of TRSes employing a feedback loop mechanism and human-in-the-loop that identified additional TRSes more adjusted to the language of Reddit.
%\todo{write this as it was planned}

For each TRS in the original set of 300 IPIP-NEO items, the expert is asked to annotate 20 statements from the dataset that were most similar to it according to cosine similarity, which will serve as a TCSS score. The annotation scheme is described in the Section~\ref{subsubsection:simpa_impl_human_in_the_loop}.

The analysis consists of two parts, motivated by two practical considerations. The first consideration pertains to the quality of TRSes, where quality is defined by their semantic representation closely matching TISes associated with the same facet and keys. We approximate the quality of different TRSes by calculating the proportion of correct matches among the top 10 TIS candidates. The second analysis aims to improve understanding of the distribution of error types associated with the chosen TRSes. In this analysis, we calculate the distribution of annotation labels across the top 20 annotated TISes for each TRS and each domain.

\subsubsection{Results}

%\todo{write as we will do annotation using this scheme and then compare expert TRSes vs IPIP NEO, wrt tis-es we will explain that can expand the pool of trees using the same scheme by promoting 1 and 2; for LLM-based we devise experiment using LLM vs human annotators}

\begin{figure}
    \centering
    {
    %\vspace{-1em}
    %\hspace*{-.5cm}
    % \includegraphics[scale=.4]{mbti_kak.pdf}
    \includegraphics[scale=.85]{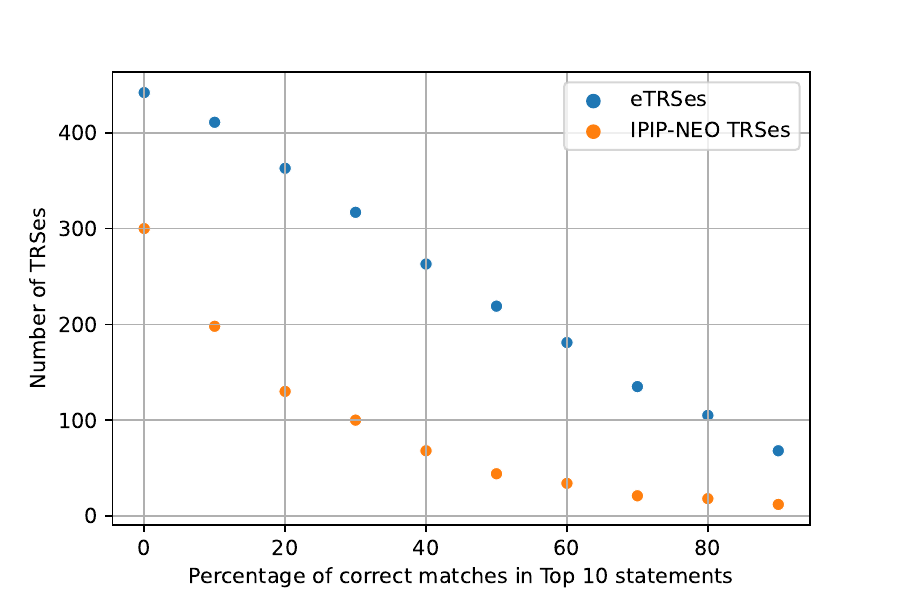}
    %\vspace*{-.5em}
    \caption{The number of TRSes (IPIP-NEO TRSes in orange and eTRSes in blue) for the different proportions of correct matches in the top 10 TIS candidates with the highest TCSS. For example, there are 100 IPIP-NEO TRSes with more than 30\% correctly matched TISes among the 10 top-ranked TISes ranked by the TCSS.}
    \label{fig:trstop10}
    }
\end{figure}

Figure~\ref{fig:trstop10} shows the number of TRSes for various proportions of correct matches among the top 10 TIS candidates with the highest TCSS. This metric serves as a proxy for TRS quality. There is a significant discrepancy in the quality of the detected TISes per TRS. Only 44 out of the 300 IPIP-NEO TRSes (indicated in orange) have more than 50\% correct matches. Table~\ref{tbl:annotations} details the proportions of annotated TIS candidates (the top 20 statements matched to TRSes using TCSS) for each personality domain and annotation category. We note considerable variations in the proportions of correct matches and informative/non-informative errors across the personality domains identified using IPIP-NEO TRSes. These discrepancies may arise from the skewed trait distributions in \pandora or because some domains have fewer IPIP-NEO TRSes that align well with TISes from social media texts. One potential solution to this issue could involve adapting TRSes to social media language.

\begin{table}
  \centering
{\small
\setlength{\tabcolsep}{4.5pt}
\caption{\label{tbl:annotations} The proportions of annotated TIS candidates (top 20 statements matched to the IPIP-NEO TRSes and eTRSes by means of TCSS) broken down by annotation categories for each Big Five domain (O -- openness; C -- conscientiousness; E -- extraversion; A -- agreeableness; N -- neuroticism). Categories (columns): OK -- a correct match; G$-$ -- statement of the same level of generality as the TRS but opposite polarity; LG$+$ -- less general statement with the same polarity; LG$-$ -- less general  with the opposite polarity;  Avg -- statements that point to the average TRS expression, SD -- statements relevant for other items/facets of the same domain; NOK -- a noninformative error.}}
\begin{tabular}{@{}crrrrrrrrr@{}}

%\topline
%& &  \multicolumn{5}{{|c|}}{Informative errors}\\

\toprule
& & & \multicolumn{5}{{c}}{Informative errors}\\
\cmidrule{4-8} \
 %& & &  General & & Less general && 
%\cline{3-6}%\cline{4-5}\\
& &  OK & G$-$ &  LG$+$ &  LG$-$ &  Avg  &  SD &  NOK \\
Domains & TRSes &  \multicolumn{1}{c}{1} & \multicolumn{1}{c}{2} & \multicolumn{1}{c}{3} &   \multicolumn{1}{c}{4}& \multicolumn{1}{c}{5}  & \multicolumn{1}{c}{6} & \multicolumn{1}{c}{7} \\
 
\midrule
O & IPIP-NEO &  24.0 & 7.7 &  21.2 & 5.0 & 1.1 &  \multicolumn{1}{c}{3.8}  &  37.2\\
 & eTRS &  36.2 & 4.5 &  16.1 &  4.0 & 2.6 &  \multicolumn{1}{c}{2.3}  &  34.3 \\
\midrule
C & IPIP-NEO &  13.2 & 1.8 &  27.9 &  4.5 & 0.3 & \multicolumn{1}{c}{2.8}  &  49.5 \\
 & eTRS &  33.6 & 5.2 &  22.9 & 5.1 & 1.2  &  \multicolumn{1}{c}{2.4} &  29.6 \\
\midrule
E & IPIP-NEO &  21.2 & 3.4 &  25.1 & 5.6 & 0.6  & \multicolumn{1}{c}{0.8}  &   43.2 \\
 & eTRS &  33.9 & 5.7 & 23.5 &  4.6 & 4.0 & \multicolumn{1}{c}{1.1}   & 27.2 \\
\midrule
A & IPIP-NEO &  18.9 & 2.8 &  17.8 & 5.9 & 0.2 & \multicolumn{1}{c}{1.4}  &  53.1 \\
 & eTRS &  33.5 & 5.0 &  20.1 & 5.3 & 0.6   & \multicolumn{1}{c}{2.5} &   33.0 \\
\midrule
N & IPIP-NEO &  22.8 & 1.9 &  35.2 & 4.3 & 1.2 & \multicolumn{1}{c}{1.0}  &   33.5 \\
 & eTRS &  29.1 & 3.6 &  32.7 & 4.8 & 2.8   &  \multicolumn{1}{c}{0.6} &  26.3 \\
\bottomrule
\end{tabular}

\end{table}
%\todo{write this in advance; explain what can be done and what we will do and why}

\subsection{Expert TRSes}
\label{sec:extending_trses}
%\todo{this is dependent on the utilization step - should we do general experiment setup first, and then describe the experiments}
%The second scenario's expert-based annotations revealed that many original IPIP-NEO items resulted in a lower rate of correctly matched TISes. Therefore, we investigated whether we could enhance the set of TRSes that consist of IPIP-NEO items by adding more expert-crafted statements. These additional items are called eTRSes.

While convenient, using IPIP items as TRSes can result in a higher error rate when matching TIS candidates. This happens mainly for two reasons. First, IPIP items are designed for interpretation by human respondents within the specific context of their own experiences. Second, these items are not phrased in a way that reflects the natural language or writing style commonly used by social media users, which decreases their relevance in this area.

In this study, we ask an expert psychologist to interpret the items that make up the IPIP-NEO facets and envision how individuals who scored high or low on each facet might describe themselves in a social media comment on a platform like Reddit. We refer to these additional items as eTRSes.

The eTRSes are divided into three categories:  
(1) paraphrases of an exact IPIP-NEO item (for example, ``I don't try to be successful'' for the IPIP-NEO item ``I am not highly motivated to succeed''),  
(2) new items associated with an existing nuance (such as ``I don't try hard'' for a potentially existing nuance comprising ``I do just enough work to get by'' and ``I put little time and effort into my work''), and  
(3) items that pertain to a new nuance (such as ``I don't care if I win or lose'' as an item representing a new nuance for indifference to winning).  

In constructing the eTRSes, the expert is guided by the following principles:  
(1) an eTRS should be a strong indicator of a facet,  
(2) more colloquial than standard items,  
(3) general (specific only when specificity enhances the relevance of the cue, for instance, ``I hit stuff when I'm angry''),  
(4) phrased to increase the likelihood of an NLP model relying on semantic similarity to match TISes to it (i.e., avoid using metaphors, be concise), and  
(5) negated not only through negations (such as ``I'm always late'' instead of ``I'm never on time'' for low self-discipline).

Consequently, eTRSes mostly take the form of references to personality traits and self-concepts (refer to Section~\ref{sec:tis-types}). With eTRSes written as trait references, we anticipate primarily capturing trait reference TISes and, to a lesser degree, individual acts and self-concepts (for instance, an eTRS ``I read a lot of nonfiction'' could correspond to an equivalent self-reference TIS, an act report TIS ``I've read the best nonfiction book of 2020'', an act observation TIS ``I recommend you read more nonfiction'', and a self-concept TIS ``I'm a nonfiction kind of person''). The eTRSes phrased as self-concepts are expected to match self-concept TISes. Still, they may also correspond to trait references and, to a lesser extent, individual acts (such as an eTRS ``I'm a creative person'' matching with a self-reference TIS ``I do a lot of creative DIY''). Please note that eTRSes of the self-concept type are entirely new compared to IPIP items.

To enhance the accuracy of detecting correctly matched TISes, we refine eTRSes in two steps by scrutinizing the semantic similarity between (1) eTRSes and (2) a set of statements from a held-out portion of the \pandora dataset consisting of comments from targets without Big Five ground truth scores (n=8684). In the first step, we examine the similarity scores between eTRSes and add highly semantically similar sentences linked to two different facets or domains to the set of eTRSes. This is expected to reduce informative errors in identifying TISes relevant to other TRSes or facets of the same domain, thereby enhancing the detection of correct matches. In the second step, we rank eTRSes and a set of sentences from the held-out Reddit dataset by cosine similarity. We then expand the set of existing TRSes with those statements if they had the same meaning but were phrased differently. We eliminate or refine existing TRSes unsuitable for that trait (i.e., where the most similar sentences were not a good match). While executing this last step, we examine semantic similarity and whether they correspond to the presumed traits.

We compare the quality of eTRSes using the same procedure as in Section~\ref{section:simpa_impl_trs_ipip}. First, we assess the quality of different TRSes by calculating the proportion of correct matches among the top 10 TIS candidates. This analysis will indicate whether the eTRSes perform worse or better than the IPIP-based TRSes. Secondly, we break down the statistics by annotation label across the top 20 annotated TISes for each TRS and each domain, which informs us about the differences in the distribution of error types between eTRSes and IPIP-based TRSes.

\subsubsection{Results}

The expert generated a set of 453 novel items relating to the Big Five domains and 30 IPIP-NEO facets. The overall time investment in constructing and improving eTRSes was around 100 hours, and the annotation took approximately 40 hours. However, the resulting list of eTRSes is preliminary, and it requires further validation in future studies, despite the preliminary findings, such as correlations between personality scores based on the matches with these items and gold-labeled scores, which suggest that the list is reliable.
%\todo{move to results section; experiments -> experiment definition -> results}
We compared the TISes detected using TRSes from the original IPIP-NEO items with those detected using eTRSes in terms of both relevance and quantity. To estimate relevance, we annotated the detected TISes matched to TRSes in both sets (as described in Section~\ref{subsubsection:simpa_impl_human_in_the_loop}). We also examined the number of TISes at various TCSS thresholds. Figure~\ref{fig:trstop10} shows that eTRSes outperform IPIP-NEO items regarding the number of correctly matched TISes. For instance, among the top 10 most similar statements, only 44 out of 300 (14.7\%) IPIP-NEO items had 50\% or more correct matches, while 219 out of 453 (48.3\%) eTRSes did. Table~\ref{tbl:annotations} provides a more detailed comparison of the relevance of detected TISes for IPIP-NEO and eTRSes. In addition to identifying more correct TISes, eTRSes generated fewer noninformative errors and maintained a more balanced proportion between correct matches, informative, and noninformative errors than IPIP-NEO TRSes. These findings suggest that creating TRSes better aligned with how personality is expressed in language on social media is valuable.

\subsection{LLM-Based TRS generation and evaluation}
\label{sec:llm-trs}

Sections~\ref{section:simpa_impl_trs_ipip} and \ref{sec:extending_trses} described two methodologies for obtaining TRSes. The first used IPIP questionnaire items, which frequently failed to reflect the casual style observed in online communities such as Reddit. The second relied on expert-crafted eTRSes, a process that demanded considerable time and effort from a personality psychology expert. In this experiment, we investigate whether a LLM (see subection~\ref{section:background_lm}) can generate and evaluate TRSes for personality assessment, reducing reliance on either purely formal or manually created statements.

We employ the OpenAI's \emph{Davinci003} model to generate 50 statements for each key of every facet in the Big Five domains as defined by the IPIP-NEO. In total, the model produces 3000 candidate TRSes. Each generation prompt explicitly requests statements written in a style characteristic of Reddit posts, in order to capture a more informal register:
\begin{quote}
\textit{list 50 simple statements that could be personality questionnaire items measuring \{key\} \{domain\}'s facet \{facet\} but written as if someone wrote them on Reddit. Make them as diverse as possible.}
\end{quote}

After each generation, we implement a three-step process to assess the quality of the resulting statements. First, all statements undergo a basic \textit{face validity} check, where they are evaluated for any obvious irrelevance or lack of clarity. Next, for a more thorough evaluation, 69 final-year psychology students at the University of Zagreb annotate each TRS candidate using the Alanno platform 	\cite{jukic-etal-2023-alanno}. Annotators assess whether a candidate statement truly reflects the intended facet and domain. The prompt for annotation is formulated as:
\begin{quote}
\textit{Does the following statement indicate the facet of adventurousness in the openness to experience domain?}
\end{quote}
Annotators choose from four labels: \emph{another facet of the same domain}, \emph{yes, in the direction of a less pronounced facet}, \emph{yes, in the direction of a more pronounced facet}, or \emph{no, it's not a signal for the domain}. Each TRS candidate is assigned to three independent annotators, and the final step involves computing Krippendorff's alpha and pairwise agreement to evaluate inter-annotator reliability.

\textbf{Krippendorff's alpha (\(\alpha\))} is defined as
\[
\alpha = 1 - \frac{D_{o}}{D_{e}},
\]
where \(D_{o}\) is the observed disagreement -- calculated as the sum of squared differences between annotation labels for all pairs of elements, weighted by the frequency of each label -- and \(D_{e}\) is the expected disagreement derived from marginal distributions. In this task, \(\alpha\) accounts for the ordinal nature of the labels (e.g., distinguishing between \emph{yes, in the direction of a more pronounced facet} and \emph{no, it's not a signal for the domain}). Its values range from \(-1\) (perfect disagreement) to \(1\) (perfect agreement), with \(0\) indicating no better-than-chance agreement.

\textbf{Pairwise agreement} measures the proportion of items on which two annotators assign the same label. Let \(N_{\text{agree}}\) be the number of items for which the annotators concur, and \(N_{\text{total}}\) be the total number of items:
\[
A = \frac{N_{\text{agree}}}{N_{\text{total}}}.
\]
Although intuitive, pairwise agreement does not adjust for chance alignment and is therefore less robust than Krippendorff's alpha.

Furthermore, we compare another LLM (OpenAI's ChatGPT 3.5 Turbo) to the student annotators by translating the original annotation prompt into English and presenting it as follows:
\begin{quote}
\textit{You are a psychologist, and you should answer the question about the statement using possible answers. question: Does the following statement indicate the facet of \{facet\} of the \{domain\} domain? statement: \{statement\} possible answers: [\textbf{'another facet of the same domain'}, \textbf{'yes, in the direction of less pronounced facet'}, \textbf{'yes, in the direction of more pronounced facet'}, \textbf{'no, it's not a signal for the domain.'}]}
\end{quote}
We compute the model’s accuracy by comparing its labels to the majority label assigned by the students, and we also calculate the model’s mean pairwise agreement with the annotators.

\subsubsection{Results}
\label{sec:llm-trs-results}

The annotations yielded a Krippendorff's alpha of 0.57, accompanied by a mean pairwise agreement of 0.71. These values reflect a moderate-to-low consensus among annotators. The ChatGPT 3.5 Turbo model achieved an accuracy of 66.6\% when compared to the students’ majority labels, with concordant judgments on 1999 TRS candidates. The model’s mean pairwise agreement score was 0.63, surpassing that of eight student annotators. However, most students outperformed the model: the overall mean pairwise agreement among students themselves was 0.701 (70.1 \%).

Table~\ref{table:traits_facets_options_values} summarizes accuracy across different traits, facets, and keys. The results illustrate marked variations within a single facet: for example, the model’s accuracy reached 88\% on TRSes that indicate higher Cooperation but dropped to 64\% on TRSes that indicate lower Cooperation. In addition, statements for which students unanimously agreed yielded higher model accuracy (76.2\%) than statements marked by inter-annotator disagreement (51.6\%). When \emph{soft misses} (e.g., \emph{another facet of the same domain}) were regarded as \emph{correct}, the overall accuracy of the model increased to 79\%.

These findings suggest that while LLMs can generate and evaluate TRSes with a moderate level of reliability, their performance does not exceed that of an average human annotator. Caution may be warranted before using LLMs as a substitute for expert review in contexts demanding fine-grained personality assessment.

%\todo{add graphics describing the process of evaluation and comparing; also add examples of alanno UI}
\FloatBarrier
\begin{table}[h]

\centering
\caption{Proportion of TRSes that LLM identified as identical to aggregated label by human annotators}
\begin{tabular}{llcc}
\hline
\textbf{Trait} & \textbf{Facet} & \textbf{High} & \textbf{Low} \\ \hline
Agreeableness & Altruism & 0.70 & 0.40 \\
 & Cooperation & 0.88 & 0.64 \\
 & Modesty & 0.74 & 0.40 \\
 & Morality & 0.50 & 0.64 \\
 & Sympathy & 0.82 & 0.94 \\
 & Trust & 0.96 & 0.56 \\
Conscientiousness & Achievement-Striving & 0.96 & 0.96 \\
 & Cautiousness & 0.96 & 0.92 \\
 & Dutifulness & 0.80 & 0.40 \\
 & Orderliness & 0.74 & 0.52 \\
 & Self-Discipline & 0.90 & 0.84 \\
 & Self-Efficacy & 0.84 & 0.82 \\
Extraversion & Activity Level & 0.58 & 0.34 \\
 & Assertiveness & 0.82 & 0.82 \\
 & Cheerfulness & 0.74 & 0.10 \\
 & Excitement-Seeking & 0.96 & 0.54 \\
 & Friendliness & 0.74 & 0.66 \\
 & Gregariousness & 0.84 & 0.82 \\
Neuroticism & Anger & 1.00 & 0.66 \\
 & Anxiety & 0.94 & 0.46 \\
 & Depression & 0.74 & 0.08 \\
 & Immoderation & 0.52 & 0.34 \\
 & Self-Consciousness & 0.84 & 0.44 \\
 & Vulnerability & 0.76 & 0.26 \\
Openness & Adventurousness & 0.98 & 0.34 \\
 & Artistic Interests & 0.96 & 0.76 \\
 & Emotionality & 0.60 & 0.10 \\
 & Imagination & 0.46 & 0.56 \\
 & Intellect & 0.82 & 0.80 \\
 & Liberalism & 0.56 & 0.20 \\ \hline
\end{tabular}
\label{table:traits_facets_options_values}
\end{table}
\FloatBarrier
%- good as we already have labels and readily available

%\todo{improve this}
\subsection{Promoting TISes to TRSes using a feedback loop}

The creation of feedback-looped TRSes is orthogonal to the choice of TRSes and is specific to SIMPA. This process may incorporate previously established IPIP items, expert-crafted eTRSes, or those generated via LLMS. A key feature that distinguishes SIMPA is its capacity to dynamically expand the set of TRSes through an iterative feedback loop (see Section~\ref{section:simpa_feedback_loop}), thus adapting to the language characteristics of a given data source. Each original TRS is associated with a set of TISes and their corresponding TCSS scores. In an ideal scenario with perfect detection accuracy, every TIS would precisely match the key, facet, and trait of its corresponding TRS. However, such perfection is unattainable in practice, resulting in various informative and noninformative errors (see Section~\ref{section:simpa_impl_detection}). Some TISes may capture nuances or keys not present in the original TRSes and, because they emerge naturally in real-world text, these TISes may exhibit higher availability and thus improve detection rates.

To address this, we experiment with two methods for promoting TISes to TRSes. The first method applies a fixed TCSS threshold, assigning each promoted TIS the labels of its matched TRS. The second method engages expert annotators, following the annotation scheme described in Section~\ref{subsubsection:simpa_impl_human_in_the_loop}, who may adjust the key or facet when promoting TISes to TRSes.

\subsubsection{Results}
\label{sec:feedback_loop_results}

\begin{figure}
    \centering
    %\vspace{-1em}
    %\hspace*{-.5cm}
    % \includegraphics[scale=.4]{mbti_kak.pdf}
    \includegraphics[scale=0.85]{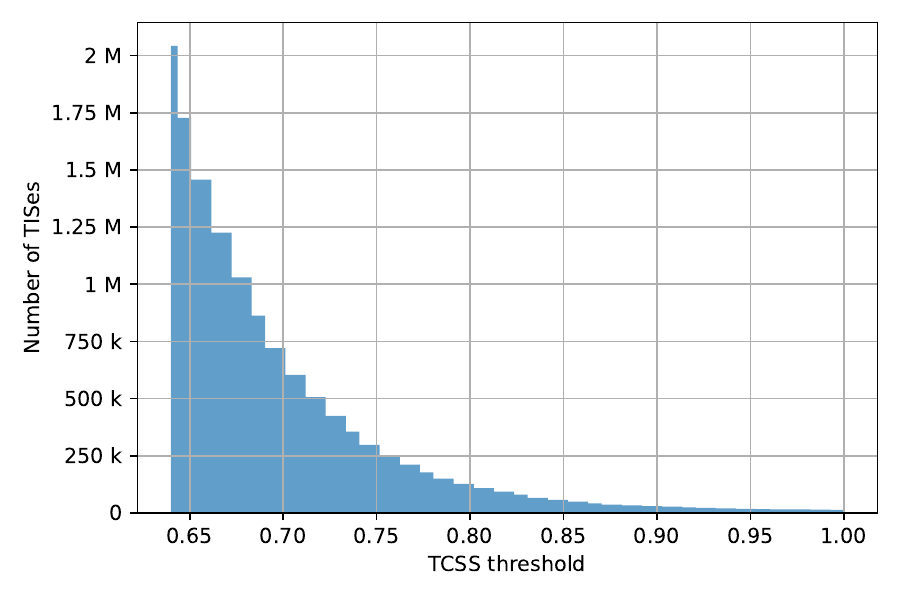}
    %\vspace*{-.5em}
    \caption{The number of TISes detected at different TCSS thresholds}
    \label{fig:tisvssim}
\end{figure}

Our analysis revealed two key findings. First, the similarity threshold directly influences both the number and relevance of TISes that exceed the threshold, as illustrated in Figure~\ref{fig:tisvssim}. This allows for automated promotion of TISes to TRSes by selecting thresholds that yield a desired number of promoted TRSes. Second, as shown in Table~\ref{tbl:annotations} (discussed in Section~\ref{sec:extending_trses}), the rate of correctly matched TISes typically exceeds 30\%, while matches for less general items range from 20\% to 30\%. Thus, a substantial number of TRSes can be added after just one feedback-loop iteration, given that there are 300 IPIP TRSes and 453 eTRSes.

Our analysis further indicates that different models yield markedly different similarity scores across TRSes. Although applying customized thresholds for each TRS would likely optimize performance by balancing precision and recall, implementing such an approach is neither straightforward nor practical. Consequently, we evaluated multiple models to identify one with both high performance and minimal threshold variance across TRSes. Among the three tested models, the paraphrase detection model demonstrated the most consistent relevance and quantity at a fixed similarity threshold. It also achieved the highest overall TCSS and favorable ratios of informative to noninformative errors. Therefore, we use only this paraphrase detection model to streamline implementation.

\section{Application}

In the previous sections, we described the implementation of all SIMPA stages, which provided insights into the quality of TRSes, the availability of TISes on Reddit, and the types of errors in the detection steps. However, these insights were not used at a target level for the individual personality assessment, which we explore next. In the following section, we bring all these insights into two potential applications of the SIMPA framework for personality assessment: (1) a supervised approach where we enhance a regression model for text-based personality prediction by incorporating features derived from relevance scores obtained from the utilization stage, and (2) an unsupervised approach in which we utilize percentile scores obtained from the utilization stage to estimate targets' Big Five scores. The first approach achieves state-of-the-art performance on the \pandora Reddit dataset. Finally, we present an experiment that demonstrates how detection and utilization steps can be merged into one using human-in-the-loop or LLMs. 

\subsection{Supervised personality assessment}

In the supervised setup, we extend the best-performing model on the \pandora dataset (see Section~\ref{sec:part_two_pandora_prediction}) by adding features based on TISes. The model uses Ridge regression with features consisting of Tf-Idf weighted unigrams, as well as predictions from MBTI classifiers. We consider any sentence with a TCSS of at least 0.6 when matched to the initial set of TRSes (including IPIP-NEO TRSes and eTRSes) to be a TIS. We also include TISes matched with a TCSS above that threshold to the expanded set of TRSes after one iteration of the feedback loop, as validated by a human expert.

To use the relevance scoring function as model features, we compute the sum of positively and negatively keyed TISes for all Big Five facets and domains. We then apply PCA (with 10 principal components) separately to the raw relevance scores matrix and a row-normalized score matrix for all targets. This approach allows us to obtain a fixed set of 20 dense features for each target, addressing the problem of TIS sparsity. We use the same cross-validation procedure and folds as in Section~\ref{sec:part_two_pandora_prediction} to allow direct comparison of the results.

\subsubsection{Results}
The correlation coefficients between the Big Five predicted and ground truth scores for all targets in the \pandora dataset ($n=1,608$), using the original model and our SIMPA-based extension, are presented in Table~\ref{tbl:predictions}. Our final model, which incorporates just 20 additional TIS-based features, achieves new state-of-the-art results for all five Big Five traits. In particular, a statistically significant improvement (p = 0.018, Steiger's test with two-tailed tails for dependent correlations \cite{steiger1980}) is realized for extraversion, with a correlation of .458. These results are promising for two reasons. First, they illustrate that TIS-based features complement word unigrams and features derived from other personality models. Second, given the numerous simplifications made during the detection and utilization stages, these results imply that there is substantial room for further improvements in predictive accuracy.
%\FloatBarrier
\begin{table}

  \centering
{\small
\setlength{\tabcolsep}{4.5pt}
\caption{\label{tbl:predictions}Pearson correlation coefficients between predicted Big Five scores and ground truth scores for \pandora best-performing prediction models with and without SIMPA TIS-based features.
Significant differences in correlations (p$<$.05) are shown in bold.}
}
\begin{tabular}{lcc}
\toprule 
Domains & \pandora-best & \pandora-best+SIMPA \\
\midrule 
Openness        &    .265 & .285\\
Conscientiousness &     .273 & .304\\  
Extraversion &   .387 &     \textbf{.458}\\ 
Agreeableness  &  .270 & .287\\ 
Neuroticism        &       .283 & .312\\ 
\bottomrule 
\end{tabular}
\end{table}
%\FloatBarrier

\subsection{Unsupervised personality assessment}

For the first application, we utilize percentile scores for each of the Big Five traits obtained as aggregate relevance scores (see Section~\ref{sec:utilization}). We aim to evaluate whether the estimations of Big Five traits, based on such a simplified implementation of the relevance scoring function, demonstrate convergent validity. To do so, we compare these estimations with the self-reported Big Five scores of targets in the \pandora dataset.

To simplify the implementation in the second application, we fixed certain parameters in the detection and utilization steps as described in Section~\ref{section:simpa_impl_detection} and Section~\ref{sec:utilization}. Specifically, we consider only TRSes that have at least one correct match among the top 10 most similar sentences, and we set a TCSS threshold of 0.6. Additionally, we analyze only targets with more than 10 detected TISes for at least one Big Five trait because of the high correlation of the number of detected TISes between different traits of the same target (r$>$.85). 

\subsubsection{Results}
\begin{figure}
    \centering
    %\vspace{-1em}
    %\hspace*{-.5cm}
    % \includegraphics[scale=.4]{mbti_kak.pdf}
    \includegraphics[scale=0.85]{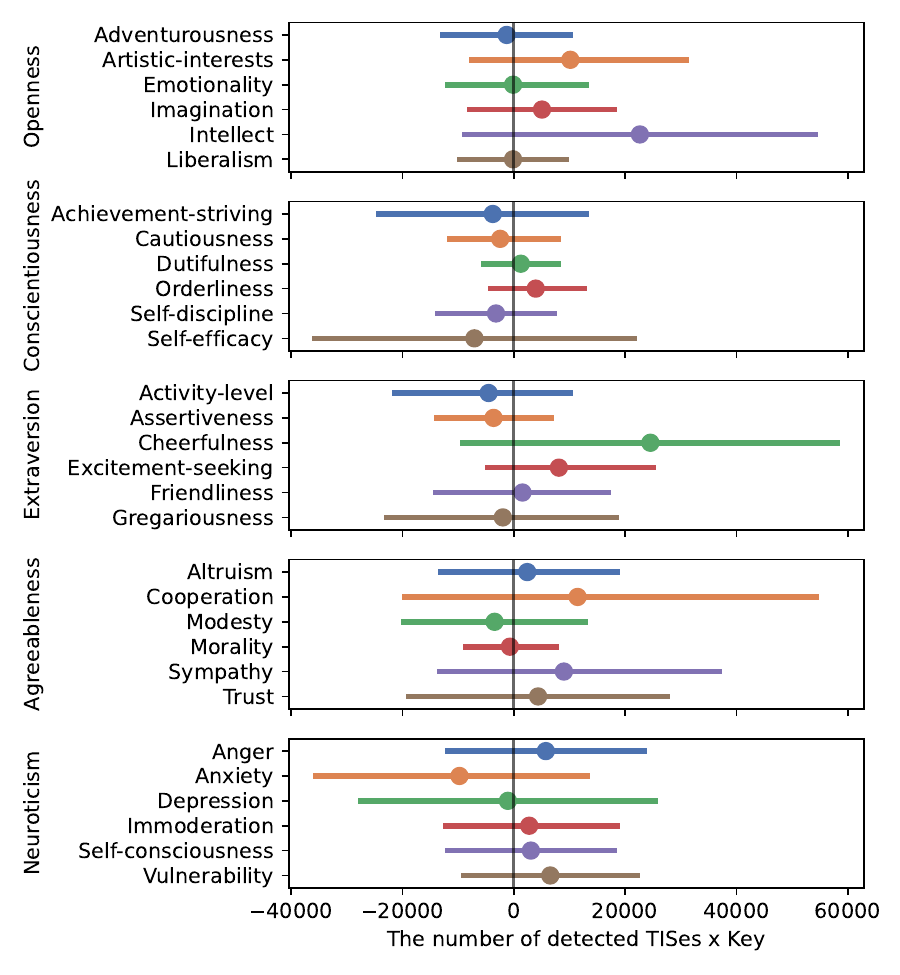}
    %\vspace*{-.5em}
    \caption{The number of detected TISes per domain and facet across all targets. The dots represent the difference between each facet's total number of positively and negatively keyed TISes. The lines show the range of the number of positively and negatively keyed TISes.}
    \label{fig:tisdist}
\end{figure}

Figure~\ref{fig:tisdist} displays the distribution of TISes detected for all targets, separated into positive and negative keys for all Big Five facets. The results show an imbalance between the numbers of positively and negatively keyed TISes for most facets and traits. For instance, the facet of cheerfulness has six times more positively keyed TISes than negatively keyed ones. The cause of this disparity may be the unequal availability of TISes for a specific key, an imbalance in the distribution of personality traits among the targets in the dataset, or inadequate coverage of TRSes for one of the two keys for certain facets.

Table~\ref{tbl:correlations} displays the Pearson correlation coefficients between the estimated and self-reported Big Five scores for TISes matched to TRSes based on IPIP-NEO items, eTRSes, and a combination of both. Due to the previously mentioned constraints, the number of targets is smaller than the initial 1,608 targets with self-reported Big Five scores, with 155 targets for IPIP-NEO TRSes, 280 for eTRSes, and 399 when both sets of TRSes are considered. Despite the many simplifications introduced in the implementation of the detection and utilization stages of the SIMPA framework (see~sections \ref{section:simpa_impl_detection} and \ref{sec:utilization}), the results indicate evidence for both convergent and discriminant validity. Specifically, using only eTRSes, we achieve significant positive correlations for all traits of interest (convergent validity), except for neuroticism, and no significant correlations between traits (discriminant validity). However, the effect sizes indicate ample room for improvement.

\begin{table}

  \centering
{\small
\setlength{\tabcolsep}{4.5pt}
\caption{\label{tbl:correlations}Pearson correlation coefficients between percentile estimates obtained using different sets of TRSes and ground truth scores for the Big Five domains (O -- openness; C -- conscientiousness; E -- extraversion; A -- agreeableness; N -- neuroticism). Significant correlations (p$<$.05) are shown in bold.}
}
\begin{tabular}{@{}ccrrrrr@{}}

\toprule
TRSes & Domains & O & C & E & A & N \\
\midrule
%IPIP NEO
&O & .099 &  .055 & $-$.020 &  .136 &  \textbf{.203} \\
&C &  \textbf{.231} &  .179 &  \textbf{.227} &  \textbf{.190} & \textbf{$-$.220 }\\
IPIP-NEO TRS &E &  .066 &  .113 &  \textbf{.285 }&  .075 &  .013 \\
&A &  .003 &  .010 &  .053 &  \textbf{.204 }& $-$.088 \\
&N & $-$.088 & \textbf{$-$.186} & $-$.146 & \textbf{$-$.253} &  \textbf{.175 }\\
\midrule 
%eTRS
&O & \textbf{.143} &  .045 &  .074 &  .084 & $-$.061 \\
&C & $-$.006 &  \textbf{.160} &  .018 &  .052 & $-$.104 \\
eTRS &E &  .040 &  .064 &  \textbf{.241} &  .034 & $-$.122 \\
&A &  .012 &  .054 &  .067 &  \textbf{.154} &  .073 \\
&N &  .036 & $-$.025 & $-$.003 & $-$.072 &  .121 \\
\midrule 
%Combined
&O & \textbf{.141 }&  .013 &  .067 &  \textbf{.165} &  .066 \\
&C &  \textbf{.126 }&  .102 &  \textbf{.102 }& $-$.003 & $-$.089 \\
Combined&E &  .070 &  \textbf{.075} &  \textbf{.226} &  \textbf{.138} & $-$.041 \\
&A &  .027 & $-$.021 &  \textbf{.106} &  \textbf{.185} &  .036 \\
&N & $-$.025 & $-$.064 & $-$.063 & $-$.086 &  \textbf{.106 }\\
\bottomrule

\end{tabular}
\end{table}

%\todo{move to experiments}

\subsection{Merging detection and utilization stages: human and LLM assisted assessment}

%\section{The comparison of LLM and human judgments}
Our next research question investigates how effectively human annotators and LLMs perform in a task centered on assessing a target's personality based on a list of their statements. This task successfully combines detection and utilization stages, resulting in a personality assessment at the target level. The experiment also illustrates how TCSS score computation can be accomplished in multiple steps. Initially, semantic similarity matching filters a TIS candidate set. The TIS candidates are then presented to a judge (either human or LLM) who conducts implicit detection, considering constraints outlined in the prompt while utilizing previously acquired expert knowledge (in their weights or brains).

In this task, we presented student annotators with statements related to the Big Five personality domains, compiled using the SIMPA pipeline implementation described in previous sections. Based on these statements, their task was to evaluate a person's personality traits, such as openness to experience. In the Alanno platform \cite{jukic-etal-2023-alanno}, annotators were given four labels to choose from for each set of statements: \emph{above average}, \emph{average}, \emph{below average}, and \emph{unable to decide}. To compile these statements, we employed the following method: selecting up to three statements for each facet of a personality domain based on their initial TCSS score and then aggregating these statements into a list. We intentionally reduced the number of statements per example to lessen the cognitive burden on annotators and the time spent on the task. In total, we created a set of 400 examples consisting of statements related to five domains from 80 authors.
The task example is as follows: 

\begin{quote}\textit{
"Based on the following statements, assess the \{domain\} of the person who wrote them. (in Croatian: Na temelju sljedećih izjava \{ime domene\} koja ih je napisala.)
------------
i used to like that place and i hate how it basically hardly ever changes., i want her to shock the world and try something new., i *really* liked the music, wow i love poetry, and i still like her music, i have mixed feelings about this., and I don't feel like drinking., i know that feeling so well ugh., I can't even picture that, i can see it., i can't even picture that, i just read over the wikipedia,, i like watching the same movies over and over too, I am so curious, ugh, well i mean, i like women's rights....., i don't necessarily think I'm right though, because I get it, I don't have a problem with religion in itself (though I definitely think fundamentalism/extremism is bad"
}\end{quote}
%\todo{move this example to some kind of figure/example}

We employ the OpenAI's Chat-GPT 3.5 Turbo 1106 model to perform a similar task. The model was given a modified prompt: \begin{quote}\textit{"You are a psychologist tasked with judging the personality trait of \{trait\} based on a set of statements. Respond with one of these grades: above average, average, below average, cannot decide. Statements: \{aggregated\_statements\}."}\end{quote}

\subsubsection{Results}
\begin{figure*}
    \centering
   % \vspace{-1em}
    %\hspace*{-0.5cm}
    \includegraphics[scale=.7]{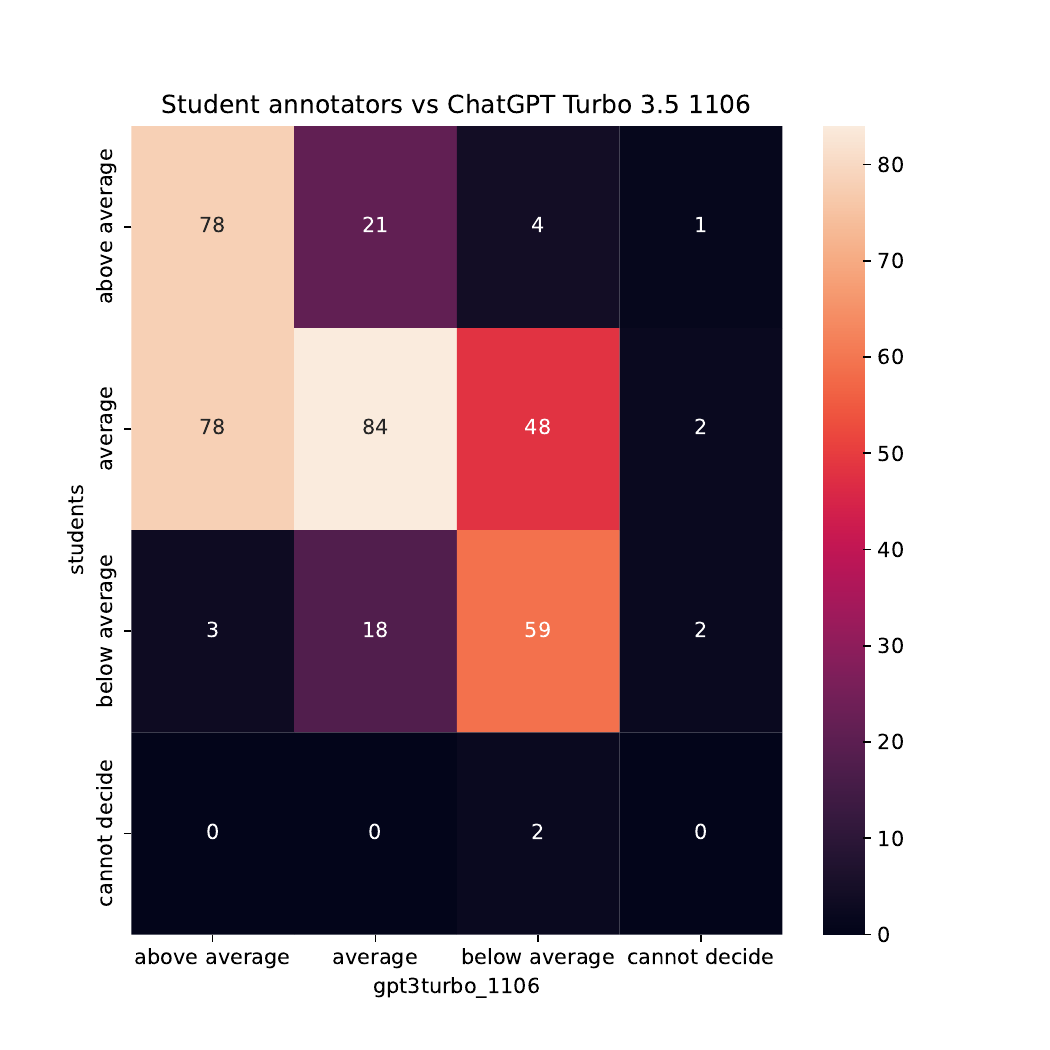}
    %\vspace*{-0.5em}
    \caption{Confusion matrix of assessments made by student annotators and ChatGPT 3.5 Turbo 1106}
    \label{fig:humans_vs_chatfpt}
\end{figure*}

Our findings showed that the Chat-GPT 3.5 model ranked 20th out of 49 annotators, performing better than average with a pairwise agreement score of 0.45 across all 49 annotators, where the mean pairwise agreement was 0.43 (see Section~\ref{sec:llm-trs}). 
The Krippendorff's alpha was 0.18 without the inclusion of ChatGPT and 0.19 with its inclusion, which is very low and indicates that the task of assessing personality traits on a limited number of statements is difficult and subjective, as there was a significant variability of judgments between human annotators.
The accuracy of the LLM was 55.25\% when taking into account the IPIP-NEO test scores of the authors. However, there were only seven opposite assessments, constituting 1.75\% of the cases. As Figure~\ref{fig:humans_vs_chatfpt} shows, one side (LLM or students) often opted for an average score in the event of below- or above-average results.

The results indicate that LLMs, like ChatGPT, have a comparable ability to human annotators in conducting personality assessments based on textual cues related to personality. These findings highlight the potential of LLMs in psychological evaluations and pave the way for future research.

\chapter{Conclusion}
\label{ch:conclusion}

The advantage of ATBPA lies in its ability to efficiently process large volumes of text data, identifying both explicit descriptions of behavior, thoughts, and emotions, as well as subtle, implicit personality traits. Unlike humans, who may struggle to detect nuanced linguistic cues such as language style, ATBPA consistently recognizes these patterns with a level of reliability that surpasses human performance. Moreover, its consistency helps mitigate common issues associated with traditional assessment methods, including respondent fatigue -- where lengthy surveys can compromise response quality -- and the tendency of individuals to provide socially desirable answers.

This thesis identified several shortcomings in existing research on text-based personality assessment. It emphasizes the impact of confounding variables at various levels, from the nuances of language generation to differences within the hierarchical taxonomy of personality traits, including their interplay. A significant focus is placed on demographics where existing research falls short, primarily due to available datasets' limited quantity, quality, and labeling accuracy.

Recognizing key gaps in existing personality research, we turned to a novel data source -- Reddit -- to extract richer personality cues and labels. Reddit’s unique structure, characterized by topically focused subreddits and user anonymity, encourages a broader range of discussions and enables more nuanced comparisons with other data sources. Leveraging these advantages, we developed two new datasets for personality analysis, MBTI9k, and PANDORA, as detailed in Part~\ref{part:two}. The \pandora dataset, in particular, is a large-scale collection comprising over 17 million comments from more than 10,000 authors, and it integrates multiple personality models, including MBTI and the Big Five, alongside demographic details such as age, gender, and location. This comprehensive resource has allowed us to address deficiencies in prior datasets and investigate research questions related to biases, confounders, and the interchangeability of personality model labels while validating existing findings in personality research.

One of the main obstacles in personality research based on text is that many datasets are not publicly available. We have made the \pandora dataset and the MBTI9k dataset accessible upon request and come with a User Agreement to ensure the privacy of Reddit users.

Through experimental work with the PANDORA dataset, we identified selection biases and confounders that impact model performance and validity. These findings have pinpointed areas for improvement in model interpretability, explainability, and validity. In response, we conceptualized a framework to guide the design of models with these qualities while being model-agnostic. The property of being model-agnostic allows for incorporating the latest developments in NLP and ML while maintaining fundamental concepts that communicate which parts of the process are delegated to NLP models.

The SIMPA (Statement-to-Item Matching Personality Assessment) framework integrates NLP and personality research concepts. Inspired by the Realistic Accuracy Model (RAM), SIMPA facilitates a sequential judgment process, offering insights into the vulnerabilities of this pipeline and highlighting that ML modeling is just one aspect of a broader challenge. Contrary to the prevailing trend in NLP of quickly deploying the latest models, SIMPA emphasizes the relevance of personality cues, which we term Trait Relevant Statements (TRSes), their availability in the chosen data source, and the iterative process of identifying and evaluating textual cues for personality assessment, which we term Trait Indicative Statements (TISes).

We demonstrated the utility of the SIMPA framework through its implementation on \pandora dataset. We showcased its ability to generate and evaluate TRSes, retrieve and utilize TISes, and make judgments comparable to human judges, with the added benefit of processing vast amounts of text efficiently.

\section{Limitations}

We acknowledge that the language used on social media may not always reflect an individual’s personality, as the specific subreddit, audience, or the nature of online presentation may influence it. However, on Reddit, where anonymity is encouraged, users may have less motivation to misrepresent themselves. Often, Redditors use their personality types to contextualize discussions or seek advice, which could lead to more honest representations of their personalities. Nevertheless, the language used on online forums can differ significantly from other forms of communication, limiting the generalizability of findings to different contexts.

\paragraph{MBTI9k.}
The MBTI9k dataset, with its large sample size, is a valuable source of personality-labeled data. However, we recognize its limitations, as it is based on the MBTI personality model, which, while popular in non-academic circles, is not extensively adopted in scientific research due to its lack of empirical validation (see Section~\ref{subsection:background_typetraitmodels}).

Another limitation of the MBTI9k dataset is the lack of demographic data. This omission raises concerns about the association between language use and personality dimensions, as confounding factors such as age, gender, and geographic location can significantly influence these associations, as demonstrated in Section~\ref{section:background_personality_demographics} and Section~\ref{section:dataset_descriptive_statistics}, and experimentally verified in Section~\ref{section:genderpred}. We acknowledge that the absence of demographic data may impact feature analysis in Chapter~\ref{ch:personality_features}. For example, features related to syntactic complexity or vocabulary richness in English texts may be more influenced by whether users wrote in their native language. As noted in Section~\ref{section:dataset_descriptive_statistics}, we identified users from countries where English is spoken as a second language.

Additionally, the dataset's data collection method is inherently biased, as it focuses solely on users who voluntarily disclosed their MBTI types through flairs. This restriction limits the dataset to individuals who are aware of their MBTI type, actively participate in related subreddits, and have chosen to share this information. Consequently, the distribution of personality types in the dataset may not accurately reflect the general population.

Another potential source of bias stems from the self-reported nature of MBTI types, as individuals may hold misconceptions about their personality traits or present themselves differently in online environments. Furthermore, variability in the MBTI tests taken by users introduces additional uncertainty in these labels.

Finally, the temporal scope of the dataset (2015–2018) presents another limitation, as it may exclude relevant authors from other periods. Additionally, contemporary trends and topics could influence language use, further impacting the dataset’s applicability to present-day analyses.

\paragraph{\pandora.}
In comparison, the \pandora dataset represents a significant advancement, as it adopts the widely accepted Big Five personality model and incorporates demographic variables to help control for confounding factors. However, it shares some limitations with the MBTI9k dataset, such as selection biases and the exclusive use of texts written in English, while also introducing new challenges. 

Specifically for \pandora, the Big Five test results underwent several processing steps that may have introduced selection bias, as detailed in Section~\ref{section:dataset_full_text_labels}. For example, only written reports of personality tests were processed, leaving out reports based on screenshots or images. Furthermore, some reports’ test types were uncertain, and we trained a classifier to identify the test based on its format. However, this classifier’s accuracy is imperfect, with a macro F1 score of 81.4\%, meaning that some tests may have been misclassified. There were likely errors in normalizing scores to percentiles for misclassified reports. Despite these limitations, we believe the final scores are as accurate as possible, given the ambiguity of the various web services offering tests, the tests themselves, and how Redditors report them in personality-related subreddits.

Another challenge is that the Big Five personality traits are often considered too general to correlate significantly with specific life outcomes. A more detailed approach using facets would be preferable. Unfortunately, facet scores are available only for some personality reports, which decreases the number of examples in experiments using self-reported facet scores.

Other limitations relate to demographic labels in the \pandora dataset. Gender labels are based on a single pattern (``/m/f/'') with the highest precision, meaning users who indicated their gender differently were not labeled. Additionally, we extracted gender labels from self-reports, such as ``I am (fe)male'', but users can indicate their gender in various ways. There are also limitations with location-based extractors, as only specific locations were captured, and sometimes, it was unclear which location the data referred to.

Finally, our experiments mainly used simple models, leaving room for performance improvement. In this regard, \pandora is well-suited for constructing more advanced deep learning models. However, creating an efficient user representation based on a potentially large number of comments remains a significant challenge.

As with the MBTI9k dataset and the MBTI classifier, we emphasize that predictive models for the Big Five have predictive validity comparable to previous research but remain unreliable at the individual level.

\paragraph{SIMPA.}
In our proof-of-concept implementation for the SIMPA framework (see Chapter~\ref{ch:implementation}, we made several simplifications to demonstrate its viability. 

For example, in the detection stage, we streamlined the preprocessing step and modeled TISes as complete sentences rather than, more accurately, as clauses. We also treated the calculation of the TCSS as a paraphrase detection task, thereby making the process of matching TISes to TRSes neutral with respect to trait constraints. Furthermore, the matching process was divided into two separate actions: embedding the sentences and calculating their cosine similarity, even though a combined approach might yield greater efficiency. 

In the application phase, we simplified the relevance scoring mechanism by equating the TCSS score with the relevance of TISes, only considering the highest TCSS score rather than leveraging the full range of similarity scores available for evaluation. Moreover, in the utilization stage we did not consider the expected base rates of TISes per TRS, TRSes per facet, and facets at the domain level, and we set all weights to 1. However, these simplifications were instrumental in showcasing the potential of the SIMPA framework and highlighting areas for enhancement. 

Incorporating LLMs tackled some of these improvement opportunities. We implicitly introduced trait constraints in the context provided to LLMs through carefully crafted prompts. We demonstrated that using LLM models following the SIMPA paradigm can approach human-level performance in generating and evaluating TRSes or making judgments. However, we utilized previous-generation LLMs and employed simple prompts, which establishes their performance at the lower bounds of what is achievable with state-of-the-art NLP. 

More effort should be directed toward developing a more effective relevance scoring function by considering extralinguistic contextual information. A more robust TCSS implementation should address discourse-level phenomena, such as co-references and ambiguities that extend beyond the sentence level, by providing LLMs with extended context. Besides using LLMs, another promising approach to addressing these challenges would be to train a TCSS model using metric learning techniques \cite{reimers-2019-sentence-bert} on a dataset annotated by experts. 

There are currently several apparent challenges and opportunities for future research based on the known limitations. One major challenge is that, unlike self-report measures where participants respond to every item, our approach may not capture all TISes for all TRSes for each participant. This can challenge content validity, which refers to the degree to which the cues are construct-relevant and adequately cover the construct (e.g., TISes should be relevant for a facet and cover all nuances of the facet). Achieving full coverage in our framework may be difficult. However, the framework does provide evidence based on content validity, as it provides information on the match between cues and valid items.

\section{Implications}

The experiments described in Part~\ref{part:two} demonstrate the potential for recognizing and predicting Redditor personalities based on simple features, such as word n-grams. These experiments also show face validity, with most features aligning with expected associations at the dimension and type levels of personality. A key finding is the unique personality distribution on Reddit, which differs from the general population, suggesting a higher prevalence of introverted, thinking, and intuitive types among Redditors. This implies a trend toward higher openness to experience, lower agreeableness, and increased introversion compared to the general population’s Big Five personality trait distribution.

The experiment on personality bias in gender classification in Section~\ref{section:genderpred} highlighted the importance of considering both demographics and personality for personality-relevant tasks. We also showed how personality influences tasks like gender prediction. This has implications for various tasks, primarily since Reddit data is often used to train conversational models. Our confirmatory study on the propensity for philosophy demonstrated that the \pandora dataset could provide additional insights into previous findings from other data sources or theories. Moreover, our study on using abundant MBTI data to improve Big Five predictions has practical implications for developing more effective and accurate systems.

However, personality prediction systems are unlikely to achieve the same level of performance as those used for predicting demographic variables. Our analysis in Chapter~\ref{ch:part_two_experiments} reveals a clear performance disparity, with accuracy exceeding 90\% for gender prediction, a Pearson correlation coefficient of approximately 0.7 for age, and a lower range of 0.2 to 0.4 for personality traits. These differences are primarily attributable to the nature of the construct itself, as personality models are designed to be parsimonious representations of the latent structure of personality. More specific personality traits are expected to yield higher correlations (see Chapter~\ref{ch:background_personality}), suggesting that focusing on personality facets rather than broad traits may enhance predictive performance.

Additionally, the availability and quality of personality cues play a crucial role. Some traits are more readily expressed on Reddit than others, as demonstrated in our prediction metrics, where extraversion is easier to detect than other traits. Future research should prioritize identifying instances rich in personality-relevant cues to enable more reliable predictions while ensuring face validity.

Following that conclusion, in Part~\ref{part:three}, we proposed the SIMPA framework, which relies on TISes by matching natural language with TRSes. However, applying SIMPA to the noisy data of social media presents novel and challenging NLP tasks. Our preliminary investigations indicate that TRSes crafted by experts outperform those derived from existing IPIP items, and we have demonstrated that LLMs can generate valid TRSes tailored to the natural discourse on platforms like Reddit. Although this LLM-based approach does not eliminate the need for a feedback loop, it significantly accelerates the convergence of the SIMPA process. Furthermore, the rapidly evolving ecosystem around Retrieval Augmented Generation \cite{lewis2020retrieval} offers a promising foundation for preselecting TIS candidates while addressing computational inefficiencies observed in our proof-of-concept implementation.

An alternative strategy leverages the increased context sizes of modern LLMs, enabling them to process comprehensive contextual information, including linguistic nuances, TISes, and relevant demographic details, for final analysis and personality assessment. This approach is orthogonal to another promising avenue for future work: developing agentic workflows that allow SIMPA to dynamically adapt and proactively address subtasks, such as utilizing cues with lower availability.

Future implementations of the SIMPA pipeline should leverage advances in LLM interpretability research. In practical applications, where transparency depends on interpretability and explainability, LLMs should not only provide assessments but also articulate their reasoning processes in an accessible manner, detailing the logical and inferential steps that lead to conclusions about personality traits.

Finally, while our current focus is on employing SIMPA for personality assessment, the framework is highly versatile. It can be applied to any construct of interest, regardless of its relation to psychology. SIMPA is particularly well-suited for tasks involving a limited number of labels, complex label taxonomies, and layered cues with varying degrees of relevance. Additionally, it is adaptable to large datasets and diverse data sources, making it applicable to a wide range of research and practical questions.

%%%%%%%%%%%%%%%%%%%%%%%%%%%%%%%%%%%%%%%%%%%%%%%%%%%%%%%%%%%%%%%%%%%%%%%%%%%
\backmatter

%%%%%%%%%%%%%%%%%%%%%%% LITERATURA / BIBLIOGRAPHY %%%%%%%%%%%%%%%%%%%%%%%%%
% bibliography style file is modified IEEEtran.bst file,
% changed to suit FER's literature style
\addcontentsline{toc}{chapter}{References}
\bibliographystyle{IEEEtranFER} 
\bibliography{eg_biblio}

%%%%%%%%%%%%%%%%%%%%%%% POPIS OZNAKA / NOMENCLATURE %%%%%%%%%%%%%%%%%%%%%%%
% notation and list of symbols if needed
%\printnomenclature

%%%%%%%%%%%%%%%%%%%%%%%%%%% KAZALO POJMOVA / INDEX %%%%%%%%%%%%%%%%%%%%%%%%
% optional index
%\printindex

%%%%%%%%%%%%%%%%%%%%%%%%%%%%%%%%% LOF %%%%%%%%%%%%%%%%%%%%%%%%%%%%%%%%%%%%%
% insert optional list of figures
\listoffigures
%\cleardoublepage % start new page
%%%%%%%%%%%%%%%%%%%%%%%%%%%%%%%%% LOT %%%%%%%%%%%%%%%%%%%%%%%%%%%%%%%%%%%%%
% insert optional list of tables
\listoftables
%\cleardoublepage % start new page

%%%%%%%%%%%%%%%%%%%%%%%%% ŽIVOTOPIS / BIOGRAPHY %%%%%%%%%%%%%%%%%%%%%%%%%%%
\renewcommand{\leftmark}{Biography}
\chapter*{Biography}
\addcontentsline{toc}{chapter}{Biography}

Matej Gjurković was born on May 23, 1985, in Zagreb. He is the co-founder and director of Semadot d.o.o. (since 2023), where he leads the development and commercialization of technologies based on natural language processing. He is also a co-founder of Ulpian AI d.o.o. (since 2024), developing tools to support legal professionals.

He graduated in electrical engineering, telecommunications and informatics from the Faculty of Electrical Engineering and Computing (FER), University of Zagreb, under the mentorship of Prof. Mario Kušek. Since 2018, he has been pursuing doctoral studies at FER under the mentorship of Prof. Jan Šnajder. Until April 2024, he worked as a research associate at the Department of Electronics, Microelectronics, Computer and Intelligent Systems (ZEMRIS), TakeLab.

Matej has extensive industry experience. At FG Microtec (2009–2012), he developed the first VoIP solution for BlackBerry devices. At British Telecom (2012), he worked as a consultant on the BT One Voice Anywhere solution. At Mavenir Systems (2012–2017), he led the development of mobile applications for VoWiFi, RCS, and T-Mobile Digits.

He actively participated in several research projects, including the Croatian Science Foundation (HRZZ) project "Computational Models for Personality Prediction from Text" (2021–), the Adria Digital Media Observatory (ADMO), the ESF project "The Establishment of an Integrated System for Managing Official Documentation of the Republic of Croatia" (2018–2023), and the HRZZ COVID-19 project SOCRES: Resilience of Croatian Society during and after the Pandemic (2020–2022).

Matej Gjurković (born 1985, Zagreb) is co-founder and director of Semadot d.o.o. (since 2023), focusing on NLP technologies, and co-founder of Ulpian AI d.o.o. (since 2024), developing tools for legal professionals. He graduated from University of Zagreb Faculty of Electrical Engineering and Computing, and since 2018 has been pursuing a PhD at FER under Prof. Jan Šnajder. He worked at ZEMRIS TakeLab until 2024. His industry experience includes VoIP development at FG Microtec, consultancy at British Telecom, and mobile app leadership at Mavenir Systems (2012–2017). Matej participated in HRZZ and ESF research projects on personality prediction, disinformation in digital media, semantic search engine development, and social resilience during COVID-19.

\section*{List of published works}
\subsection*{Journal articles}
\begin{itemize}
    \item Gjurković, Matej; Vukojević, Iva; Šnajder, Jan. 
    \emph{SIMPA: Statement-to-Item Matching Personality Assessment from text.} 
    Future Generation Computer Systems, 130 (2022), 114--127. 
    doi:10.1016/j.future.2021.12.014
    \item Bilić, P.; Dukić, D.; Arambašić, L.; Gjurković, M.; Šnajder, J.; Furman, I. (2023). Digital news media as a social resilience proxy: A computational political economy perspective. New Media \& Society, 0(0). https://doi.org/10.1177/14614448231214149,  (međunarodna recenzija, članak, znanstveni)
\end{itemize}

\subsection*{Conference proceedings}
\begin{itemize}
    \item Gjurković, Matej; Karan, Mladen; Vukojević, Iva; Bošnjak, Mihaela; Šnajder, Jan. 
    \emph{PANDORA Talks: Personality and Demographics on Reddit.} 
    In: Proceedings of the Ninth International Workshop on Natural Language Processing for Social Media. Association for Computational Linguistics (ACL), 2021, pp. 138--152. 
    doi:10.18653/v1/2021.socialnlp-1.12

    \item Gjurković, Matej; Šnajder, Jan. 
    \emph{Reddit: A Gold Mine for Personality Prediction.} 
    In: Proceedings of the Second Workshop on Computational Modeling of People’s Opinions, Personality, and Emotions in Social Media. Association for Computational Linguistics (ACL), 2018, pp. 87--97. 
    doi:10.18653/v1/w18-1112 
    (awarded Best paper award at the PEOPLES Workshop)

    \item Sekulić, Ivan; Gjurković, Matej; Šnajder, Jan. 
    \emph{Not Just Depressed: Bipolar Disorder Prediction on Reddit.} 
    In: Proceedings of the 9th Workshop on Computational Approaches to Subjectivity, Sentiment and Social Media Analysis. Association for Computational Linguistics (ACL), 2018, pp. 72--78.
\end{itemize}

\renewcommand{\leftmark}{Životopis}
\chapter*{Životopis}
\addcontentsline{toc}{chapter}{Životopis}

Matej Gjurković rođen je 23. svibnja 1985. u Zagrebu. Suosnivač je i direktor tvrtke Semadot d.o.o. (od 2023.), gdje vodi razvoj i komercijalizaciju tehnologija temeljenih na obradi prirodnog jezika, te suosnivač Ulpian AI d.o.o. (od 2024.), usmjerene na razvoj alata za pravne stručnjake. Diplomirao je elektrotehniku, telekomunikacije i informatiku na Fakultetu elektrotehnike i računarstva (FER) Sveučilišta u Zagrebu s diplomskim radom ``Usluge s lokacijski označenim sadržajem'' pod mentorstvom prof. dr. sc. Marija Kušeka. Doktorski studij na FER-u upisao je 2018. godine pod mentorstvom prof. dr. sc. Jana Šnajdera, gdje je do 2024. bio suradnik na Zavodu za elektroniku, mikroelektroniku, računalne i inteligentne sustave (ZEMRIS), TakeLab.

Ima bogato industrijsko iskustvo: u FG Microtecu (2009.–2012.) razvio je prvo VoIP rješenje za BlackBerry, u British Telecomu (2012.) radio je kao konzultant, a u Mavenir Systemsu (2012.–2017.) vodio je razvoj mobilnih aplikacija za VoWiFi, RCS i T-Mobile Digits. Sudjelovao je u znanstvenim projektima poput HRZZ-ovog ``Računalni modeli za predviđanje i analizu ličnosti na temelju teksta'', Adria Digital Media Observatory (ADMO), te projektu Uspostava integralnog sustava za upravljanje službenom dokumentacijom Republike Hrvatske.

\end{document}